%% file: arxiv_main.tex
\documentclass{article} 
\usepackage[preprint]{neurips_2023}

\usepackage{hyperref}
\hypersetup{  
    hidelinks  
}  
\usepackage{url}

\usepackage{listings}
\usepackage{xcolor} 
\usepackage{fontawesome}
\newcommand{\circled}[1]{\tikz[baseline=(char.base)]{ \node[shape=circle,draw,inner sep=2pt] (char) {\scriptsize 
 #1};}}
\newcommand{\shadeCircled}[1]{\tikz[baseline=(char.base)]{ \node[shape=circle,draw,fill=black,inner sep=2pt,text=white] (char) {\scriptsize  #1};}}

\usepackage{pgfplots, pgfplotstable}
\usepackage{tikz}
\usetikzlibrary{arrows,positioning}
\usetikzlibrary{external}
\usetikzlibrary{ calc, matrix, patterns, patterns.meta }
\usetikzlibrary{shapes,arrows, fit, tikzmark}
\usetikzlibrary{calc}
\usetikzlibrary{arrows}
\usetikzlibrary{decorations.pathreplacing}

\usetikzlibrary{ calc, matrix, patterns, patterns.meta }
\definecolor{msdarkblue}{RGB}{36,58,94}
\definecolor{msblue}{RGB}{0,120,215}
\definecolor{msgreen}{RGB}{16,124,16}
\definecolor{msred}{RGB}{216,59,1}
\definecolor{purple}{RGB}{128,0,128}
\definecolor{msgray}{HTML}{DFDFDF}
\definecolor{Emerald}{HTML}{00A99D}
\newcommand{\cmark}{\ding{51}}%
\newcommand{\xmark}{\ding{55}}%

\usepackage{natbib}
\usepackage{booktabs}
\usepackage{array}
\usepackage{amssymb, amsthm}
\usepackage{mathtools}
\usepackage{tabularx} 
\usepackage{amsmath}
\usepackage{bm,bbm}
\usepackage{graphicx}
\usepackage{xspace}
\usepackage{algorithm}
\usepackage{algpseudocode}
\usepackage{multirow}
\usepackage{cancel}
\usepackage[normalem]{ulem}
\usepackage{enumitem}
\usepackage{listings}
\usepackage{xcolor}
\usepackage{graphicx}
\usepackage{subcaption}
\usepackage{amssymb}
\usepackage{pifont}
\definecolor{codegreen}{rgb}{0,0.6,0}
\definecolor{codegray}{rgb}{0.5,0.5,0.5}
\definecolor{codepurple}{rgb}{0.58,0,0.82}
\definecolor{backcolour}{rgb}{0.95,0.95,0.92}
\lstdefinestyle{mystyle}{
    backgroundcolor=\color{backcolour},   
    commentstyle=\color{codegreen},
    keywordstyle=\color{magenta},
    numberstyle=\tiny\color{codegray},
    stringstyle=\color{codepurple},
    basicstyle=\ttfamily\tiny,
    breakatwhitespace=false,         
    breaklines=true,                 
    captionpos=b,                    
    keepspaces=true,                 
    numbers=left,                    
    numbersep=5pt,                  
    showspaces=false,                
    showstringspaces=false,
    showtabs=false,                  
    tabsize=2
}

\newcolumntype{P}[1]{>{\centering\arraybackslash}p{#1}}

\usepackage{letltxmacro}
\LetLtxMacro{\originaleqref}{\eqref}
\renewcommand{\eqref}{Eq.\originaleqref}

\newcommand{\ResponsiveAgent}{{\texttt{ConversableAgent}}\xspace}
\newcommand{\UserProxyAgent}{{\texttt{UserProxyAgent}}\xspace}
\newcommand{\AssistantAgent}{{\texttt{AssistantAgent}}\xspace}

\newcommand{\groupchatmanager}{\texttt{GroupChatManager}\xspace}

\newcommand{\libName}{\texttt{AutoGen}\xspace}

\renewcommand{\cite}{\citep}

\title{\libName: Enabling Next-Gen LLM \\Applications via Multi-Agent Conversation}

\author{Qingyun Wu$^{\dagger}$, Gagan Bansal$^{\ast}$, Jieyu Zhang$^{\pm}$, Yiran Wu$^{\dagger}$, Beibin Li$^{\ast}$  
\\\\ 
{\bf Erkang Zhu$^{\ast}$,  Li Jiang$^{\ast}$, Xiaoyun Zhang$^{\ast}$, Shaokun Zhang$^{\dagger}$, Jiale Liu$^{\mp}$}
\\\\ 
{\bf Ahmed Awadallah$^{\ast}$, Ryen W. White$^{\ast}$, Doug Burger$^{\ast}$, Chi Wang$^{\ast}$\thanks{Corresponding author. Email: \href{mailto:auto-gen@outlook.com}{auto-gen@outlook.com}} }
\\\\
{$^{\ast}$Microsoft Research, $^{\dagger}$Pennsylvania State University}
\\\\ 
{$^{\pm}$University of Washington,$^{\mp}$Xidian University}
}%

\makeatletter
\let\@fnsymbol\@arabic
\makeatother
\begin{document}

\maketitle

\begin{figure}[h]
    \centering
    \includegraphics[width=\linewidth]{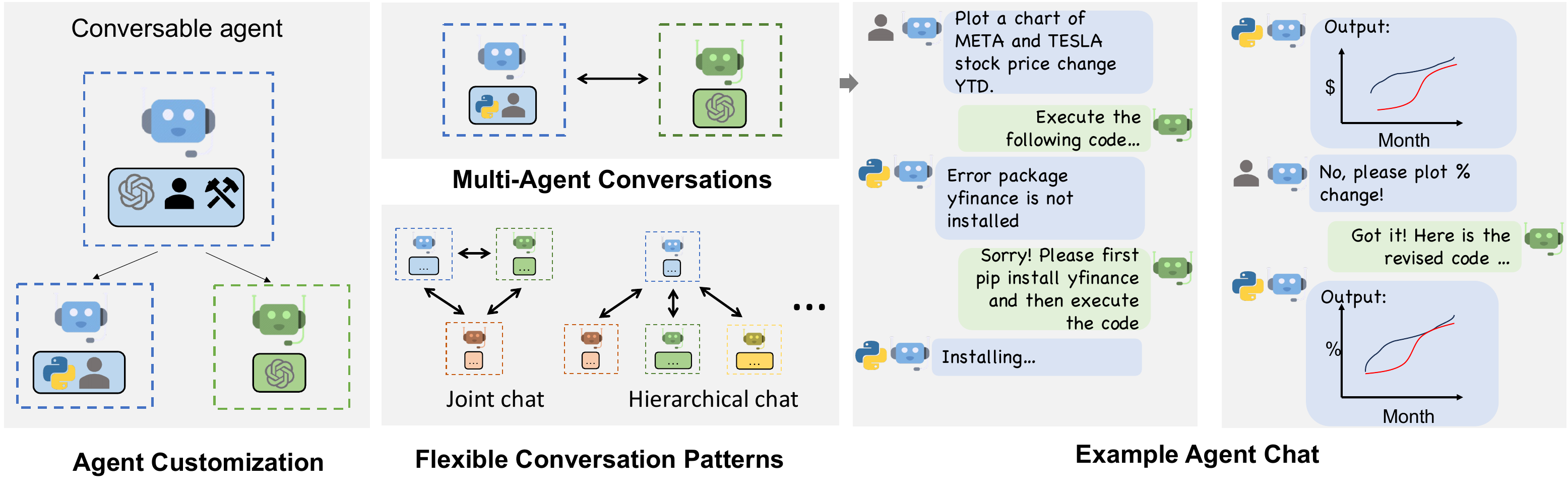}
    \caption{ \libName enables diverse LLM-based applications using multi-agent conversations. (Left) \libName agents are conversable, customizable, and can be based on LLMs, tools, humans, or even a combination of them. (Top-middle) Agents can converse to solve tasks. (Right) They can form a chat, potentially with humans in the loop. (Bottom-middle) The framework supports flexible conversation patterns.
    }
    \label{fig:landing}
\end{figure}

\begin{abstract}
\libName\footnote{\url{https://github.com/microsoft/autogen}} is an open-source framework that allows developers to build LLM applications via 
multiple {\em agents} that can converse with each other to accomplish tasks.
\libName\  agents are customizable, {\em conversable}, and can operate in various modes that employ combinations of LLMs, human inputs, and tools.
Using \libName, developers can also flexibly define agent interaction behaviors.
Both natural language and computer code can be used to program flexible conversation patterns for different applications. \libName serves as a generic framework for building diverse applications of various complexities and LLM capacities.
Empirical studies demonstrate the effectiveness of the framework in many example applications, with domains ranging from mathematics, coding,  question answering, operations research, online decision-making, entertainment, etc.
\end{abstract}

\input{sec_iclr/_sec_intro_arxiv}

\input{sec_iclr/_sec_system}

\input{sec_iclr/_sec_application}

\input{sec_iclr/_sec_summary}


\section*{Ethics statement}
 \label{append:ethical}
There are several potential ethical considerations that could arise from the development and use of the \libName framework.
\begin{itemize}[leftmargin=*]
    \vspace{-1mm}
    \setlength\itemsep{-0.1em}
    \item Privacy and Data Protection: The framework allows for human participation in conversations between agents. It is important to ensure that user data and conversations are protected, and that developers use appropriate measures to safeguard privacy.

\item Bias and Fairness: LLMs have been shown to exhibit biases present in their training data~\cite{navigli2023biases}. When using LLMs in the \libName framework, it is crucial to address and mitigate any biases that may arise in the conversations between agents. Developers should be aware of potential biases and take steps to ensure fairness and inclusivity.

\item Accountability and Transparency: As discussed in the future work section, as the framework involves multiple agents conversing and cooperating, it is important to establish clear accountability and transparency mechanisms. Users should be able to understand and trace the decision-making process of the agents involved in order to ensure accountability and address any potential issues or biases.

\item Trust and Reliance: \libName leverages human understanding and intelligence while providing automation through conversations between agents. It is important to consider the impact of this interaction on user experience, trust, and reliance on AI systems. Clear communication and user education about the capabilities and limitations of the system will be essential~\cite{Cai2019HelloAU}.

\item Unintended Consequences: As discussed before, the use of multi-agent conversations and automation in complex tasks may have unintended consequences. In particular, allowing LLM agents to make changes in external environments through code execution or function calls, such as installing packages, could be risky. Developers should carefully consider the potential risks and ensure that appropriate safeguards are in place to prevent harm or negative outcomes.

\end{itemize}

\section*{Acknowledgements}
The work presented in this report was made possible through discussions and feedback from Peter Lee, Johannes Gehrke, Eric Horvitz, Steven Lucco, Umesh Madan, Robin Moeur, Piali Choudhury, Saleema Amershi, Adam Fourney, Victor Dibia, Guoqing Zheng, Corby Rosset, Ricky Loynd, Ece Kamar, Rafah Hosn, John Langford, Ida Momennejad, Brian Krabach, Taylor Webb, Shanka Subhra Mondal, Wei-ge Chen, Robert Gruen, Yinan Li, Yue Wang, Suman Nath, Tanakorn Leesatapornwongsa, Xin Wang, Shishir Patil, Tianjun Zhang, Saehan Jo, Ishai Menache, Kontantina Mellou, Runlong Zhou, Feiran Jia, Hamed Khanpour, Hamid Palangi, Srinagesh Sharma, Julio Albinati Cortez, Amin Saied, Yuzhe Ma, Dujian Ding, Linyong Nan, Prateek Yadav, Shannon Shen, Ankur Mallick, Mark Encarnación, Lars Liden, Tianwei Yue, Julia Kiseleva, Anastasia Razdaibiedina, and Luciano Del Corro. Qingyun Wu would like to acknowledge the funding and research support from the College of Information Science and Technology at Penn State University.

\clearpage
\bibliography{agent_ref}
\bibliographystyle{iclr2023_conference}
\appendix
\input{sec_iclr/_sec_relatedwork}

\input{sec_iclr/_sec_summary_full}
\input{sec_iclr/_appendix}
\end{document}

%% file: sec_iclr/_sec_intro_arxiv.tex
\section{Introduction}

Large language models (LLMs) are becoming a crucial building block in developing powerful {\em agents} that utilize LLMs for reasoning, tool usage, and adapting to new observations~\cite{yao2022react,xi2023rise,wang2023survey} in many real-world tasks. 
Given the expanding tasks that could benefit from LLMs and the growing task complexity, an intuitive approach to scale up the power of agents is to use multiple agents that cooperate.
Prior work suggests that multiple agents can help encourage divergent thinking~\cite{liang-arxiv2023}, improve factuality and reasoning~\cite{du2023improving}, and provide validation~\cite{wu2023empirical}.
In light of the intuition and early evidence of promise, it is intriguing to ask the following question:
{\em how} can we facilitate the development of LLM applications that could span a broad spectrum of domains and complexities based on the multi-agent approach?

Our insight is to use \emph{multi-agent conversations} to achieve it. 
There are at least three reasons confirming its general feasibility and utility thanks to recent advances in LLMs: First, because chat-optimized LLMs (e.g., GPT-4) show the ability to incorporate feedback,  
LLM agents can cooperate through {\em conversations} with each other or human(s), e.g., a dialog where agents provide and seek reasoning, observations, critiques, and validation. Second, because a single LLM can exhibit a broad range of capabilities (especially when configured with the correct prompt and inference settings), conversations between differently configured agents can help 
combine
these broad LLM capabilities in a modular and complementary manner. 
{Third, LLMs have demonstrated ability to solve complex tasks when the tasks are broken into simpler subtasks. Multi-agent conversations can enable this partitioning and integration in an intuitive manner.} How can we leverage the above insights and support different applications with the common requirement of coordinating multiple agents, potentially backed by LLMs, humans, or tools exhibiting different capacities? 
We desire a multi-agent conversation framework with generic abstraction and effective implementation that has the flexibility to satisfy different application needs.
Achieving this requires addressing two critical questions:
(1) How can we design individual agents that are capable, reusable, customizable, and effective in multi-agent collaboration? (2) How can we develop a straightforward, unified interface that can accommodate a wide range of agent conversation patterns? 
In practice, applications of varying complexities may need distinct sets of agents with specific capabilities, and may require different conversation patterns, such as single- or multi-turn dialogs, different human involvement modes, and static vs. dynamic conversation. Moreover, developers may prefer the flexibility to program agent interactions in natural language or code. Failing to adequately address these two questions would limit the framework's scope of applicability and generality.

While there is contemporaneous
exploration of multi-agent approaches,\footnote{
We refer to Appendix~\ref{sec:related_work} for a detailed discussion.} we present \libName, a generalized multi-agent conversation framework (Figure~\ref{fig:landing}), 
based on the following new concepts.
\begin{enumerate}[leftmargin=*]
    \vspace{-2mm}
    \setlength\itemsep{-0.1em}
    \item[1] \textbf{Customizable and conversable agents.} \libName uses a generic design of agents that can leverage LLMs, human inputs, tools, or a combination of them. The result is that developers can easily and quickly create agents with different roles (e.g., agents to write code, execute code, wire in human feedback, validate outputs, etc.) by selecting and configuring a subset of built-in capabilities. The agent's backend can also be readily extended to allow more custom behaviors. 
    To make these agents suitable for multi-agent conversation, every
    agent is made {\em conversable}
    -- they can receive, react, and respond to messages. 
     When configured properly, an agent can hold multiple turns of conversations with other agents autonomously or solicit human inputs at certain rounds, enabling human agency and automation.
     The conversable agent design leverages the strong capability of the most advanced LLMs in taking feedback and making progress via chat
    and also allows combining capabilities of LLMs in a modular fashion.   (Section~\ref{sec:agent}) 

    \item[2] \textbf{Conversation programming.} 
    A fundamental insight of \libName is to simplify and unify complex LLM application workflows as multi-agent conversations.
    So \libName adopts a programming paradigm centered around these inter-agent conversations. We refer to this paradigm as \emph{conversation programming},  which streamlines the development of intricate 
    applications
    via two primary steps: (1) defining a set of conversable agents with specific capabilities
    and roles (as described above); (2) programming the interaction behavior between agents via conversation-centric \emph{computation} and \emph{control}. 
    Both steps can be achieved via a fusion of natural and programming languages to build applications with a wide range of conversation patterns and agent behaviors. \libName provides ready-to-use implementations and also allows easy extension and experimentation for both steps. 
    (Section~\ref{sec:autogen_conv})
  
\end{enumerate}

\libName also provides a collection of multi-agent applications created using conversable agents and conversation programming.  
These applications demonstrate how \libName can easily 
support applications of various complexities and LLMs of various capabilities. 
Moreover, we perform both evaluation on benchmarks and a pilot study of new applications. The results show that \libName can help achieve outstanding performance on many tasks, and enable innovative ways of using LLMs, while reducing development effort. (Section~\ref{sec:app} and Appendix~\ref{append:application})

%% file: sec_iclr/_sec_system.tex
\section{The \libName Framework}\label{sec:framework}
To reduce the effort required for developers to create complex LLM applications across various domains, a core design principle of \libName is to streamline and consolidate multi-agent workflows using multi-agent conversations. This approach also aims to maximize the reusability of implemented agents.
This section introduces the two key concepts of \libName: conversable agents and conversation programming.

\subsection{Conversable Agents}\label{sec:agent}

In \libName, a \emph{conversable agent} is an entity with a specific role that  can pass messages to send and receive information to and from other conversable agents, e.g., to start or continue a conversation. It maintains its internal context 
based on sent and received messages and can be configured to possess a set of capabilities, e.g., enabled by LLMs, tools, or human input, etc. The agents can act according to programmed behavior patterns described next.
 
\textbf{Agent capabilities powered by LLMs, humans, and tools.}
Since an agent's capabilities directly influence how it processes and responds to messages, \libName allows flexibility to endow its agents with various capabilities. \libName supports many common composable capabilities for agents, including
\textbf{1) LLMs.}  LLM-backed agents 
exploit many capabilities of advanced LLMs such as role playing, implicit state inference and progress making conditioned on conversation history, providing feedback, adapting from feedback, and coding. These capabilities can be combined in different ways via novel prompting techniques\footnote{Appendix~\ref{append_sys_msg} presents an example of such novel prompting techniques which empowers the default LLM-backed assistant agent in \libName to converse with other agents in multi-step problem solving.} to increase an agent's skill and autonomy. \libName also offers enhanced LLM inference features
such as result caching, error handling, message templating, etc., via an enhanced LLM inference layer. 
\textbf{2) Humans.} Human involvement is desired or even essential in many LLM applications. \libName lets a human participate in agent conversation via human-backed agents, which could solicit human inputs at certain rounds of a conversation depending on the agent configuration. The default \emph{user proxy} agent allows {\em configurable} human involvement levels and patterns, e.g., frequency and conditions for requesting human input including the option for humans to skip providing input. 
\textbf{3) Tools.} Tool-backed agents have the capability to execute tools via code execution or function execution.
For example, the default user proxy agent in \libName is able to execute code suggested by LLMs, or make LLM-suggested function calls. 

\textbf{Agent customization and cooperation.} Based on application-specific needs, each agent can be configured to have a mix of basic back-end types to display complex behavior in multi-agent conversations. \libName allows easy creation of agents with specialized capabilities and roles by reusing or extending the built-in agents. The yellow-shaded area of Figure~\ref{fig:conv_programming} provides a sketch of the built-in agents in \libName. The \ResponsiveAgent class is the highest-level agent abstraction and, by default, can use LLMs, humans, and tools. The \AssistantAgent and \UserProxyAgent are two pre-configured \ResponsiveAgent subclasses, each representing a common usage mode, i.e., acting as an AI assistant (backed by LLMs) and acting as a human proxy to solicit human input or execute code/function calls (backed by humans and/or tools). 

In the example on the right-hand side of Figure~\ref{fig:landing}, an LLM-backed assistant agent and a  tool- and human-backed user proxy agent are deployed together to tackle a task.  Here, the assistant agent generates a solution with the help of LLMs and passes the solution to the user proxy agent. Then, the user proxy agent solicits human inputs or executes the assistant's code and passes the results as feedback back to the assistant.

By allowing custom agents that can converse with each other, conversable agents in \libName serve as a useful building block. However, to develop applications where agents make meaningful progress on tasks, developers also need to be able to specify and mold these multi-agent conversations.

\input{sec_iclr/__sec_system_conv}

%% file: sec_iclr/__sec_system_conv.tex
\subsection{Conversation Programming}
\label{sec:autogen_conv}

\begin{figure}[tb]
\centering
\includegraphics[width = 0.95\hsize]{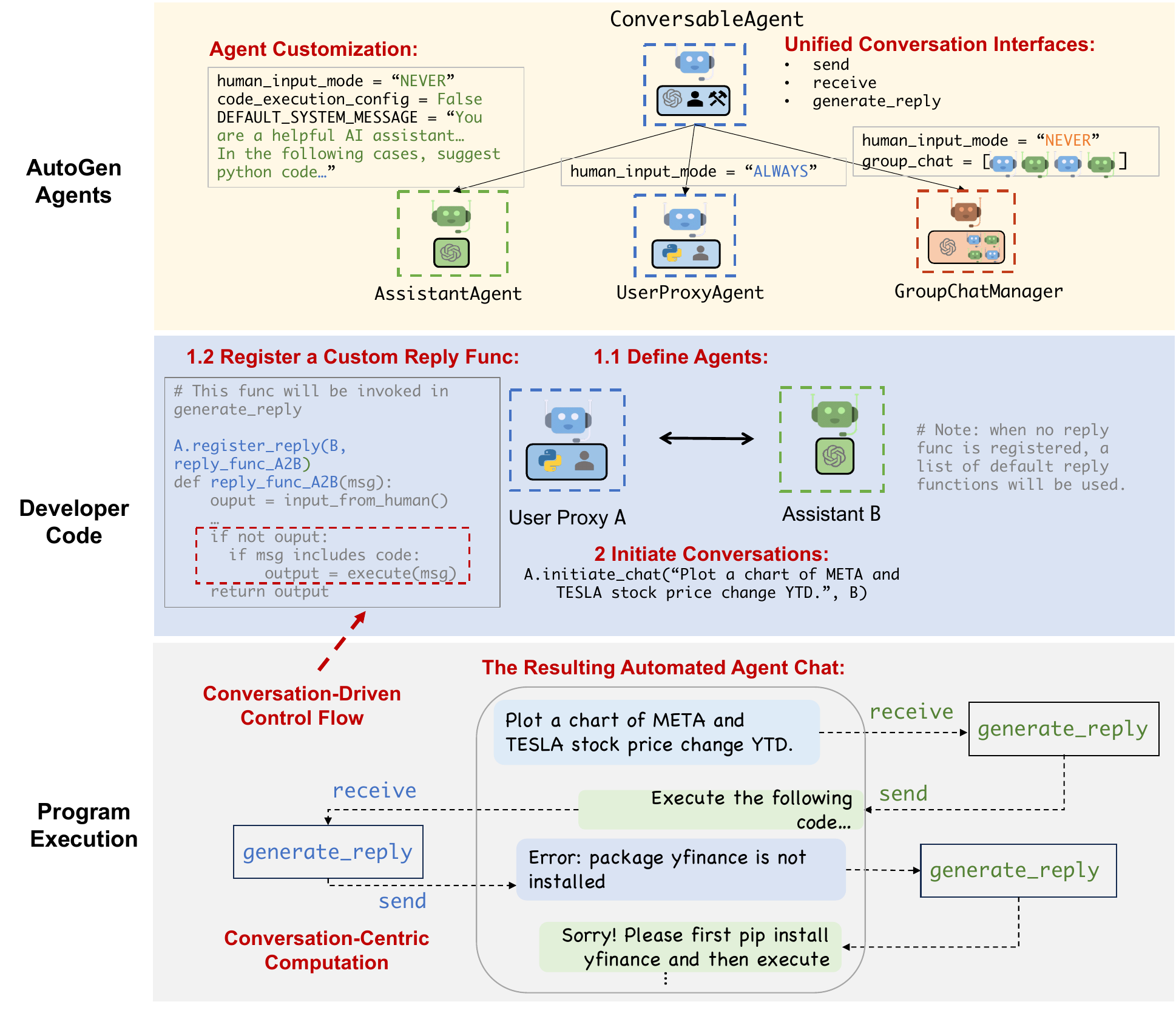}
\caption{Illustration of how to use \libName to program a multi-agent conversation. The top sub-figure illustrates the built-in agents provided by \libName, which have unified conversation interfaces and can be customized. The middle sub-figure shows an example of using \libName to develop a two-agent system with a custom reply function. The bottom sub-figure illustrates the resulting automated agent chat from the two-agent system during program execution. 
}
\label{fig:conv_programming}
\vspace{-5mm}
\end{figure}

As a solution to the above problem, \libName utilizes {\em conversation programming}, a paradigm that considers two concepts: the first is {\em computation} -- the actions agents take to compute their response in a multi-agent conversation. And the second is {\em control flow} -- the sequence (or conditions) under which these computations happen.
As we will show in the applications section, the ability to program these helps implement many flexible multi-agent conversation patterns. In \libName, these computations are conversation-centric. An agent takes actions relevant to the conversations it is involved in and its actions result in message passing for consequent conversations (unless a termination condition is satisfied). 
Similarly, control flow is conversation-driven -- the participating agents' decisions on which agents to send messages to and the procedure of computation 
are functions of the inter-agent conversation.
This paradigm helps one to reason intuitively about a complex workflow as agent action taking and conversation message-passing between agents. 

Figure~\ref{fig:conv_programming} provides a simple illustration. The bottom sub-figure shows how
individual agents perform their role-specific, conversation-centric computations to generate responses (e.g., via LLM inference calls and code execution). The task progresses through conversations displayed in the dialog box. The middle sub-figure demonstrates a conversation-based control flow. When the assistant receives a message, the user proxy agent typically sends the human input as a reply. If there is no input, it executes any code in the assistant's message instead.

\libName features the following design patterns to facilitate conversation programming:
\begin{enumerate}[leftmargin=*]
    \vspace{-1mm}
    \setlength\itemsep{-0.1em}
    \item \textbf{Unified interfaces and auto-reply mechanisms for automated agent chat.}  
   Agents in \libName have unified conversation interfaces for performing the corresponding conversation-centric computation, including a \texttt{send/receive} function for sending/receiving messages and a \texttt{generate\_reply} function for taking actions and generating a response based on the received message.    
\libName also introduces and by default adopts an \textbf{agent auto-reply} mechanism to realize conversation-driven control:
Once an agent receives a message from another agent, it automatically invokes \texttt{generate\_reply} and sends the reply back to the sender unless a termination condition is satisfied. 
\libName provides built-in reply functions based on LLM inference, code or function execution, or human input. One can also register custom reply functions to customize the behavior pattern of an agent, e.g., chatting with another agent before replying to the sender agent. Under this mechanism, once the reply functions are registered, and the conversation is initialized, the conversation flow is naturally induced, and thus the agent conversation proceeds naturally without any extra control plane, i.e., a special module that controls the conversation flow. 
For example, with the developer code in the blue-shaded area (marked ``Developer Code") of Figure~\ref{fig:conv_programming}, one can readily trigger the conversation among the agents, and the conversation would proceed automatically, as shown in the dialog box in the grey shaded area (marked ``Program Execution") of Figure~\ref{fig:conv_programming}. 
The auto-reply mechanism provides a decentralized, modular, and unified way to define the workflow.

\item \textbf{Control by fusion of programming and natural language.} \libName allows the usage of programming  and natural language in various control flow management patterns:
1) \textbf{Natural-language control via LLMs.} 
In \libName, one can control the conversation flow by prompting the LLM-backed agents with natural language. 
For instance, the default system message of the built-in \AssistantAgent in \libName uses natural language to instruct the agent to fix errors and generate code again if the previous result indicates there are errors. It also guides the agent to confine the LLM output to certain structures, making it easier for other tool-backed agents to consume. For example, instructing the agent to reply with ``TERMINATE" when all tasks are completed to terminate the program. More concrete examples of natural language controls can be found in Appendix~\ref{append_sys_msg}.  2) \textbf{Programming-language control.} In \libName, Python code can be used to specify the termination condition, human input mode, and tool execution logic, 
e.g., the max number of auto replies. 
    One can also register programmed auto-reply functions to control the conversation flow with Python code, as shown in the code block identified as ``Conversation-Driven
Control Flow" in Figure~\ref{fig:conv_programming}. 
3) \textbf{Control transition between natural and programming language.}  \libName also supports flexible control transition between natural and programming language. 
One can achieve transition from code to natural-language control by invoking an LLM inference containing certain control logic in a customized reply function; or transition from natural language to code control via LLM-proposed function calls~\cite{funccall}.

\end{enumerate}
 
In the conversation programming paradigm, one can realize multi-agent conversations of diverse patterns.
In addition to static conversation with predefined flow, \libName also supports dynamic conversation flows with multiple agents. 
\libName provides two general ways to achieve this: 1) Customized \texttt{generate\_reply} function: within the customized \texttt{generate\_reply} function, one agent can hold the current conversation while invoking conversations with other agents depending on the content of the current message and context.
2) Function calls:  In this approach, LLM decides whether or not to call a particular function depending on the conversation status. By messaging additional agents in the called functions, the LLM can drive dynamic multi-agent conversation. 
In addition, \libName supports more complex dynamic group chat via built-in \groupchatmanager, which can dynamically select the next speaker and then broadcast its response to other agents. We elaborate on this feature and its application in Section~\ref{sec:app:groupchat}.  We provide implemented working systems to showcase all these different patterns, with some of them visualized in Figure~\ref{fig:app_summary}.

%% file: sec_iclr/_sec_application.tex
\section{Applications of \libName} \label{sec:app}
\begin{figure}[thb]
\centering
\includegraphics[width = \hsize]{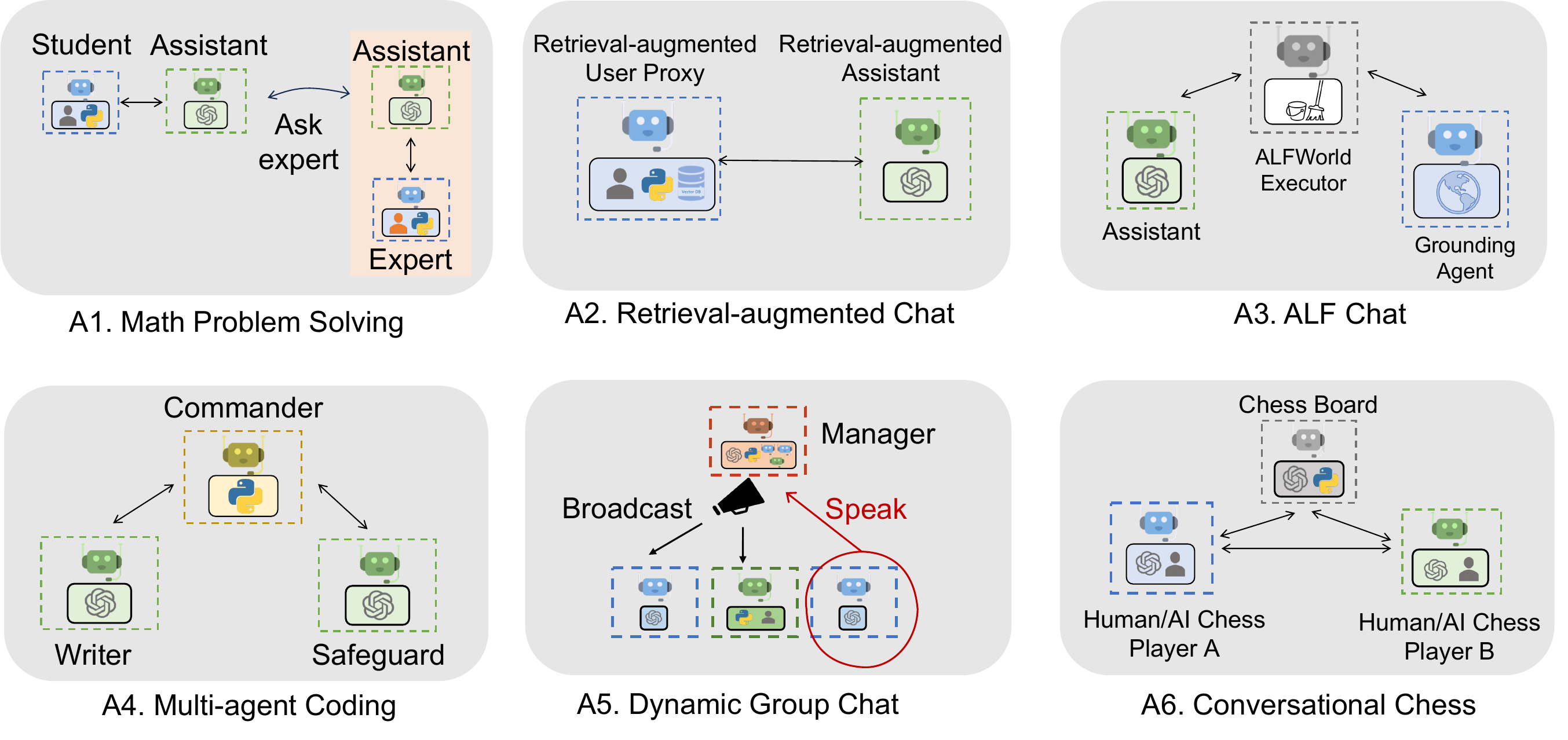}
\caption{Six examples of diverse applications built using \libName. Their conversation patterns show \libName's flexibility and power. 
}
\label{fig:app_summary}
\end{figure}

We demonstrate six applications using \libName (see Figure~\ref{fig:app_summary}) to illustrate its potential in simplifying the development of high-performance multi-agent applications. These applications are selected based on their real-world relevance (A1, A2, A4, A5, A6),  problem difficulty and solving capabilities enabled by \libName (A1, A2, A3, A4), and innovative potential (A5, A6). Together, these criteria showcase \libName's role in advancing the LLM-application landscape.

\begin{figure}[t]
    \centering
    \begin{subfigure}[b]{0.43\textwidth}
        \includegraphics[width=\textwidth]{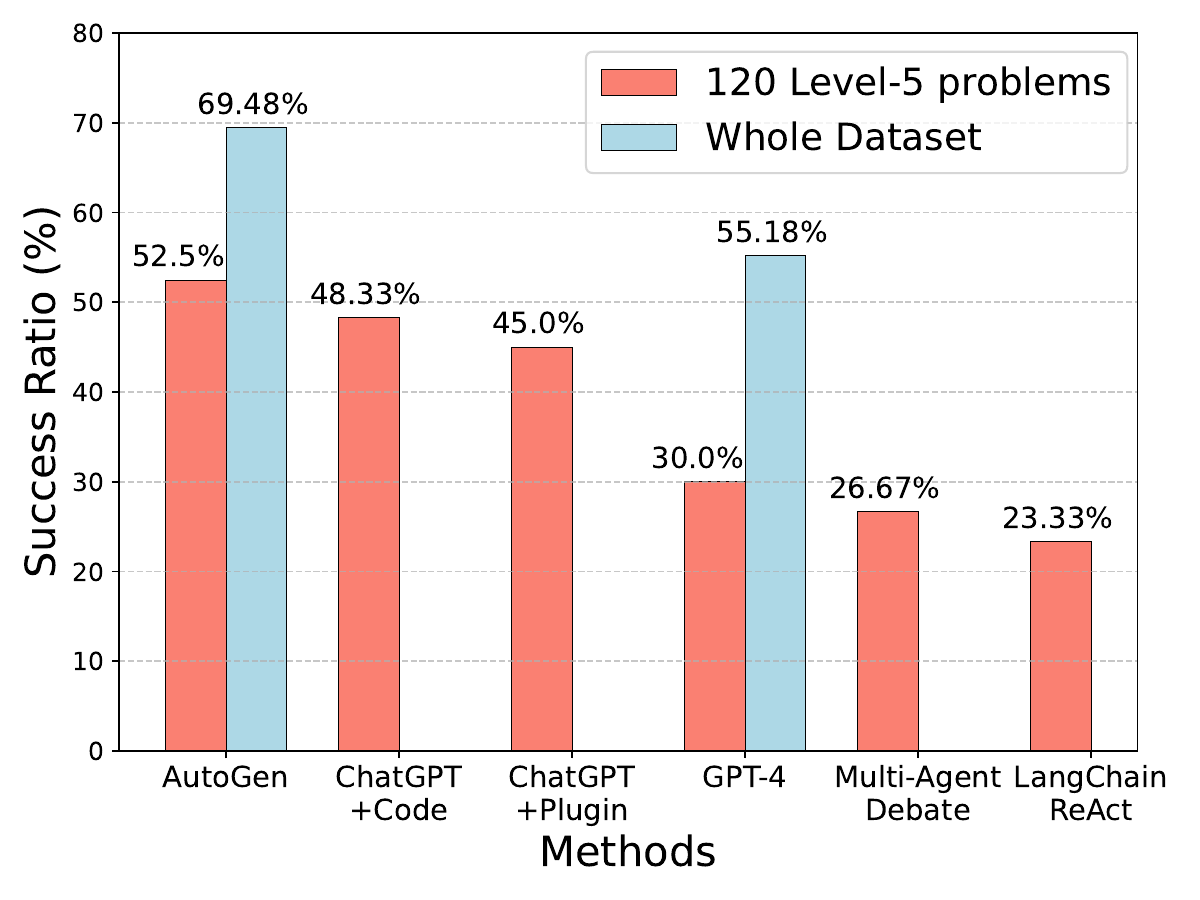}
        \caption{A1: Performance on MATH (w/ GPT-4).}
        \label{fig:res:math}
    \end{subfigure}
    \begin{subfigure}[b]{0.43\textwidth}
    \includegraphics[width=\textwidth]{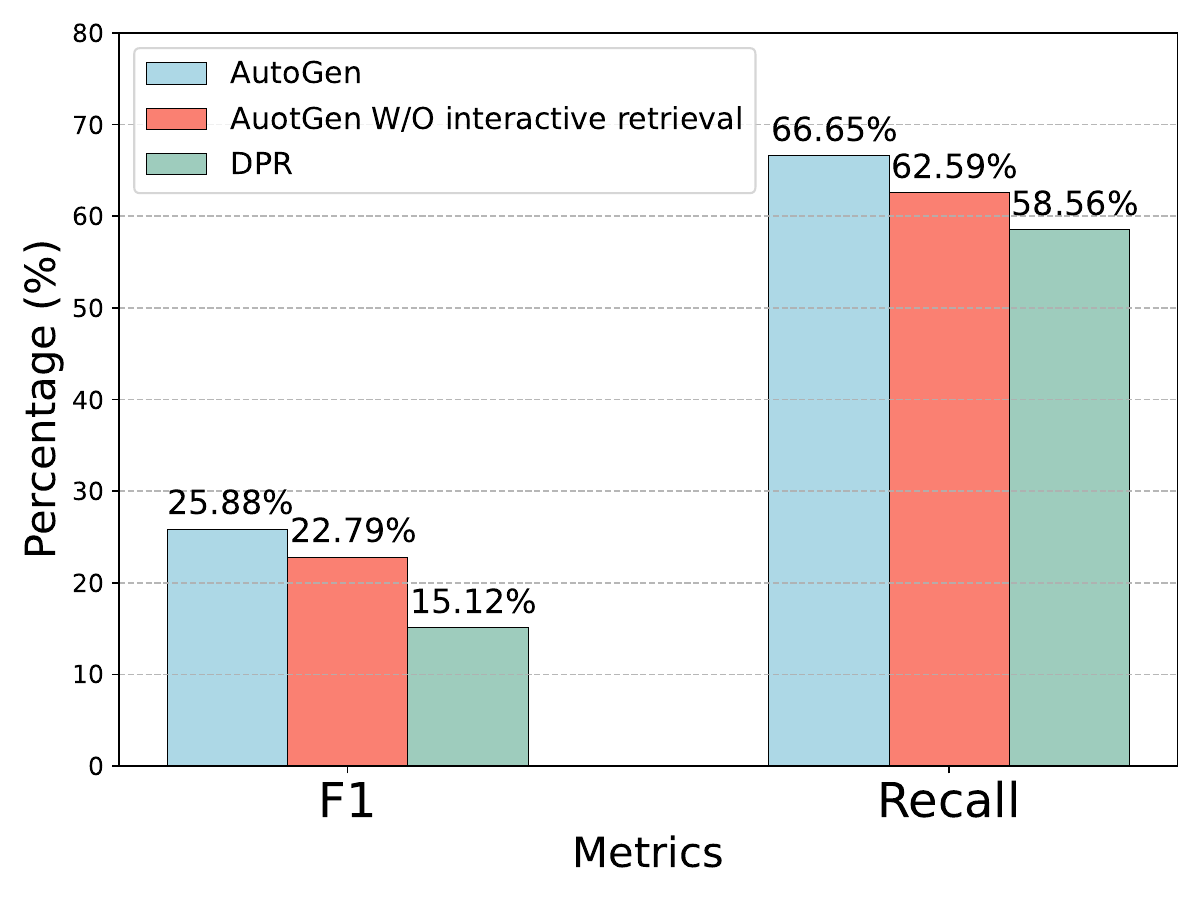}
    \caption{A2: Q\&A tasks (w/ GPT-3.5).}
    \label{fig:res:retri}
    \end{subfigure}
     \begin{subfigure}[b]{0.43\textwidth}
    \includegraphics[width=\textwidth]{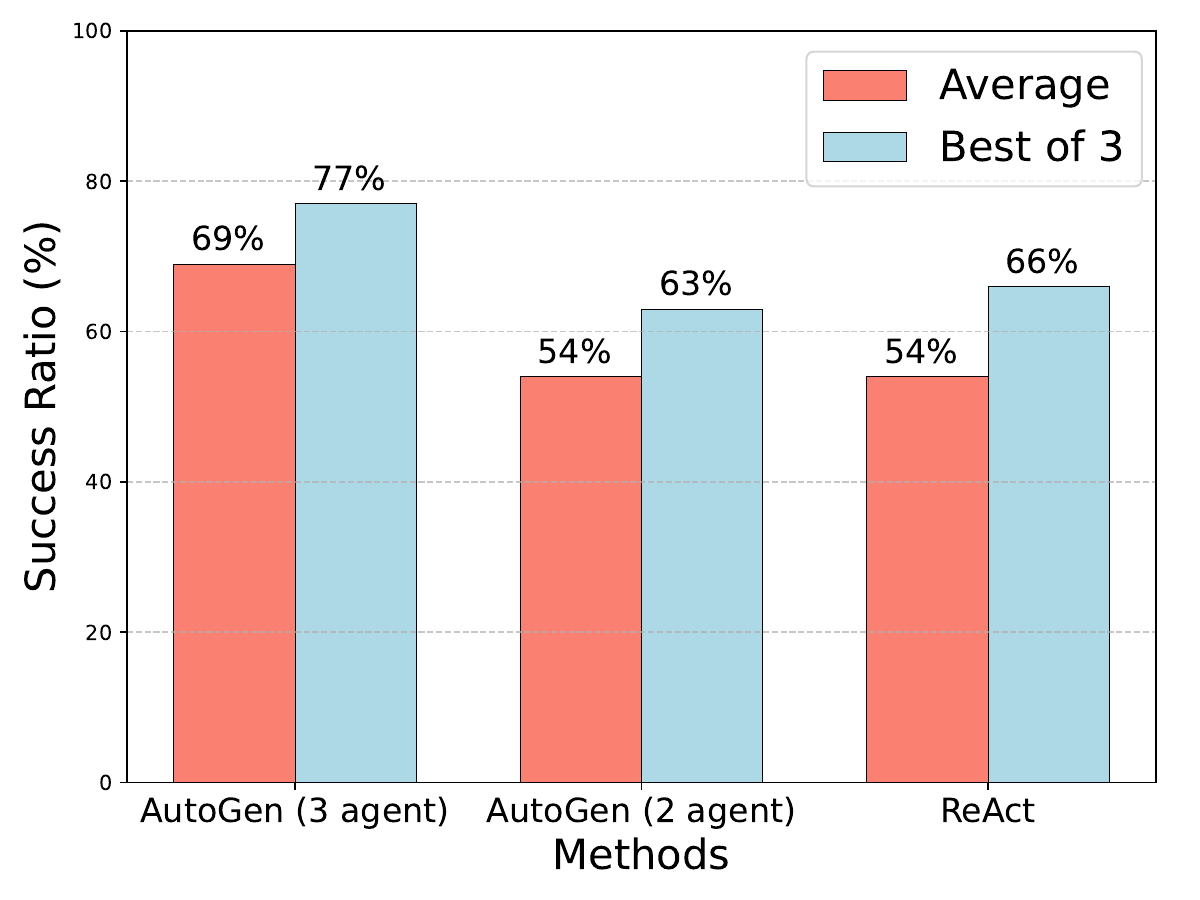}
        \caption{A3: Performance on ALFWorld.}
        \label{fig:res:alf}
    \end{subfigure}
        \begin{subfigure}[b]{0.43\textwidth}
    \includegraphics[width=\textwidth]{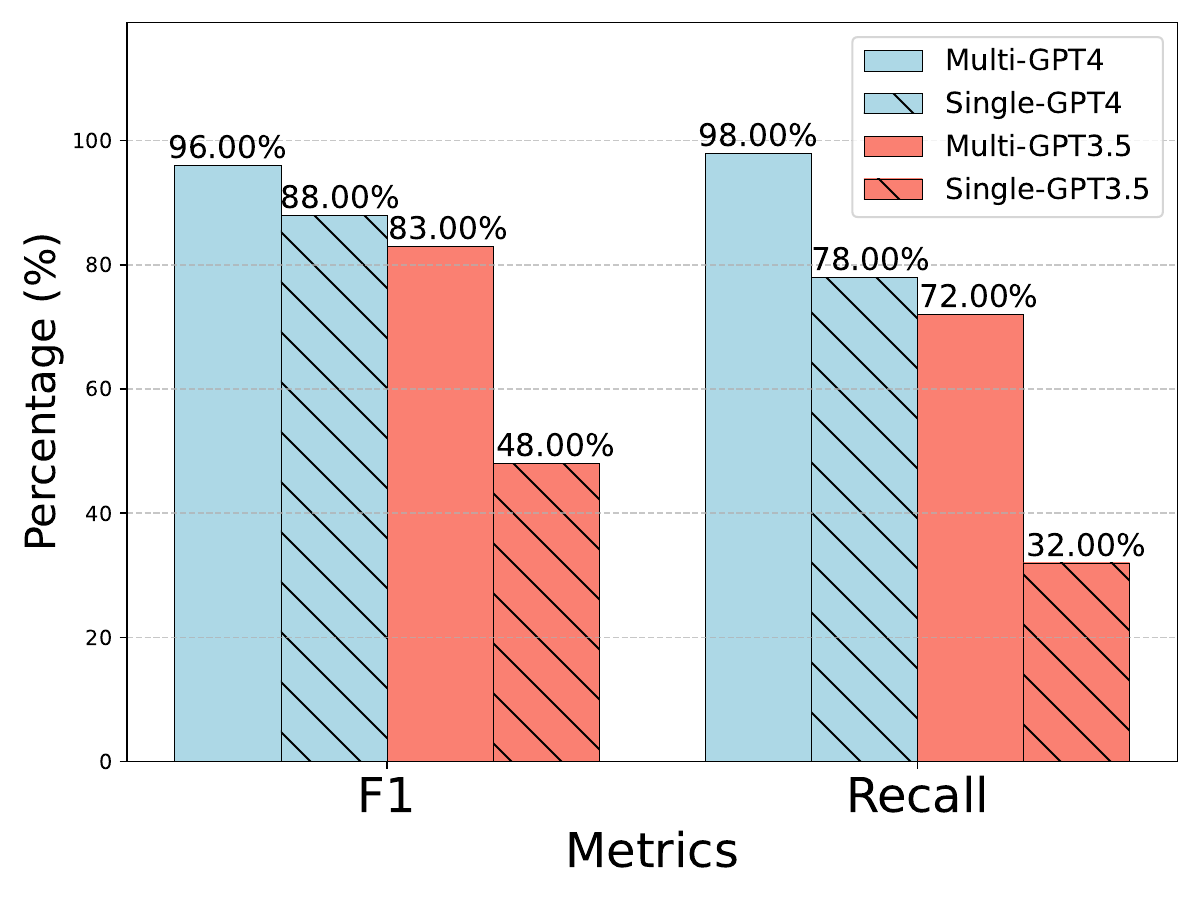}
        \caption{A4: Performance on OptiGuide. 
        }
        \label{fig:res:sageguard}
    \end{subfigure}
    \hfill 
    \caption{ Performance on four applications A1-A4. 
    (a) shows that \libName agents can be used out of the box to achieve the most competitive performance on math problem solving tasks; (b) shows that \libName can be used to realize effective retrieval augmentation and realize a novel interactive retrieval feature to boost performance on Q\&A tasks; (c) shows that \libName can be used to introduce a three-agent system with a grounding agent to improve performance on ALFWorld; (d) shows that a multi-agent design is helpful in boosting performance in coding tasks that need safeguards. 
    }
    \label{fig:results_all}
\end{figure}

\input{sec_iclr/__app_math}
\input{sec_iclr/__app_retrievechat}
\input{sec_iclr/__app_alfworld}

\input{sec_iclr/__app_optiguide}

\input{sec_iclr/__app_groupchat}
\input{sec_iclr/__app_chess}

%% file: sec_iclr/__app_math.tex
  \subsection*{A1: Math Problem Solving} \label{sec:app:math}

Mathematics is a foundational discipline and the promise of leveraging LLMs to assist with math problem solving opens up a new plethora of applications and avenues for exploration,
including personalized AI tutoring, AI research assistance, etc. This section demonstrates how \libName can help develop LLM applications for math problem solving, showcasing strong performance and flexibility in supporting various problem-solving paradigms. 

(\textbf{Scenario 1}) We are able to build a system for autonomous math problem solving by directly reusing two built-in agents from \libName. We evaluate our system and several alternative approaches, including open-source methods such as Multi-Agent Debate~\cite{liang-arxiv2023}, LangChain ReAct~\cite{langchain}, vanilla GPT-4, and commercial products ChatGPT + Code Interpreter, and ChatGPT + Plugin (Wolfram Alpha), on the MATH~\cite{hendrycks2021measuring} dataset and summarize the results in Figure~\ref{fig:res:math}. 
We perform evaluations over 120 randomly selected level-5 problems and on the entire\footnote{We did not evaluate ChatGPT on the whole dataset since it requires substantial manual effort and is restricted by its hourly message-number limitation. Multi-agent debate and LangChain ReAct were also not evaluated since they underperformed vanilla GPT-4 on the smaller test set.} test dataset from MATH.
The results show that the built-in agents from \libName already yield better performance out of the box compared to the alternative approaches, even including the commercial ones. 
(\textbf{Scenario 2}) We also showcase a human-in-the-loop problem-solving process with the help of \libName. To incorporate human feedback with \libName, one only needs to set \texttt{human\_input\_mode=`ALWAYS'} in the \UserProxyAgent of the system in scenario 1. We demonstrate that this system can effectively incorporate human inputs to solve challenging problems that cannot be solved without humans. 
(\textbf{Scenario 3}) We further demonstrate a novel scenario where {\em multiple} human users can participate in the conversations during the problem-solving process. 
Our experiments and case studies for these scenarios show that AutoGen enables better performance or new experience compared to other solutions we experimented with.
Due to the page limit, details of the evaluation, including case studies in three scenarios are in Appendix~\ref{appendix_sec:app:math}. 

%% file: sec_iclr/__app_retrievechat.tex
\subsection*{A2: Retrieval-Augmented Code Generation and Question Answering} \label{sec:app:retrieve}

Retrieval augmentation has emerged as a practical and effective approach for mitigating the intrinsic limitations of LLMs by incorporating external documents.  
In this section, we employ \libName to build a Retrieval-Augmented Generation (RAG) system~\cite{lewis2020retrieval, parvez2021retrieval} named Retrieval-augmented Chat. 
The system consists of two agents: a Retrieval-augmented User Proxy agent and a Retrieval-augmented Assistant agent, both of which are extended from built-in agents from \libName. The Retrieval-augmented User Proxy includes a vector database~\cite{chromadb} with SentenceTransformers~\cite{reimers-2019-sentence-bert} as the context retriever. A detailed workflow description of the Retrieval-augmented Chat is provided in Appendix~\ref{append:application}.

We evaluate Retrieval-augmented Chat in both question-answering and code-generation scenarios. (\textbf{Scenario 1}) We first perform an evaluation regarding natural question answering on the Natural Questions dataset~\cite{kwiatkowski2019natural} and report results in Figure~\ref{fig:res:retri}. In this evaluation, we compare our system with DPR (Dense Passage
Retrieval) following an existing evaluation\footnote{The results of DPR with GPT-3.5 shown in Figure~\ref{fig:res:retri} are from \cite{adlakha2023evaluating}. We use GPT-3.5 as a shorthand for GPT-3.5-turbo.} practice~\cite{adlakha2023evaluating}. Leveraging the conversational design and natural-language control,  \libName introduces a novel \emph{interactive retrieval} feature in this application: whenever the retrieved context does not contain the information, instead of terminating, the LLM-based assistant would reply ``\emph{Sorry, I cannot find any
information about... UPDATE CONTEXT.}” which will invoke more retrieval attempts. We conduct an ablation study in which we prompt the assistant agent to say \emph{``I don't know"} instead of \emph{``UPDATE CONTEXT."} in cases where relevant information is not found, and report results in Figure~\ref{fig:res:retri}. The results show that the interactive retrieval mechanism indeed plays a non-trivial role in the process. We give a concrete example and results using this appealing feature in Appendix~\ref{appendix_sec:app:retrive_chat}. (\textbf{Scenario 2}) We further demonstrate how Retrieval-augmented Chat aids in generating code based on a given codebase that contains code not included in GPT-4's training data. Evaluation and demonstration details for both scenarios are included in Appendix~\ref{appendix_sec:app:retrive_chat}.  

%% file: sec_iclr/__app_alfworld.tex
\subsection*{A3: Decision Making in Text World Environments} 
\label{sec:app:alfworld}
In this subsection, we demonstrate how \libName can be used to develop effective applications that involve interactive or online decision making. We perform the study using the ALFWorld~\cite{ALFWorld20} benchmark, which includes a diverse collection of synthetic language-based interactive decision-making tasks in household environments. 

With \libName, we implemented a two-agent system to solve tasks from  ALFWorld. It consists of an LLM-backed assistant agent responsible for suggesting plans to conduct a task and an executor agent responsible for executing actions in the ALFWorld environments. This system integrates ReAct prompting~\cite{yao2022react}, 
and is able to achieve similar performance. A common challenge encountered in both ReAct and the \libName-based two-agent system is their occasional inability to leverage basic commonsense knowledge about the physical world. This deficiency can lead to the system getting stuck in a loop due to repetitive errors. Fortunately, the modular design of \libName allows us to address this issue effectively: With \libName, we are able to introduce a grounding agent, which supplies crucial commonsense knowledge--such as \emph{“You must find and take the object before you can examine it. You must go to where the target object is before you can use it.”}--whenever the system exhibits early signs of recurring errors. It significantly enhances the system's ability to avoid getting entangled in error loops.
We compare the task-solving performance of the two variants of our system with GPT-3.5-turbo and ReAct\footnote{Results of ReAct are obtained by directly running its official code with default settings. The code uses \texttt{text-davinci-003} as backend LM and does not support GPT-3.5-turbo or GPT-4.} on the 134 unseen tasks from ALFWorld and report results in Figure~\ref{fig:res:alf}. The results show that introducing a grounding agent could bring in a 15\% performance gain on average. Upon examining the systems' outputs, we observe that the grounding agent, by delivering background commonsense knowledge at the right junctures, significantly mitigated the tendency of the system to persist with a flawed plan, thereby avoiding the creation of error loops. For an example trajectory comparing the systems see Appendix~\ref{appendix:app:alfworld}, Figure \ref{fig:alfchat_comparison}.

%% file: sec_iclr/__app_optiguide.tex
\subsection*{A4: Multi-Agent Coding}
\label{sec:app:optiguide}

In this subsection, we use \libName to build a multi-agent coding system based on OptiGuide~\cite{li2023large}, a system that excels at writing code to interpret optimization solutions and answer user questions, such as exploring the implications of changing a supply-chain decision or understanding why the optimizer made a particular choice. 
The second sub-figure of Figure~\ref{fig:app_summary} shows the \libName-based implementation.  The workflow is as follows:
the end user sends questions, such as ``\emph{What if we prohibit shipping from supplier 1 to roastery 2?}" to the Commander agent.  
The Commander coordinates with two assistant agents, including the Writer and the Safeguard, to answer the question. The Writer will craft code 
and send the code to the Commander. 
After receiving the code, the Commander checks the code safety with the Safeguard; if cleared, the Commander will use external tools (e.g., Python) to execute the code, and request the Writer to interpret the execution results. For instance, the writer may say ``\emph{if we prohibit shipping from supplier 1 to roastery 2, the total cost would increase by 10.5\%.}" The Commander then provides this concluding answer to the end user.  If, at a particular step, there is an exception, e.g., security red flag raised by Safeguard, the Commander redirects the issue back to the Writer with debugging information.
The process might be repeated multiple times until the user's question is answered or timed-out. 

With \libName the core workflow code for OptiGuide was reduced from over 430 lines to 100 lines, leading to significant productivity improvement. We provide a detailed comparison of user experience with ChatGPT+Code Interpreter and \libName-based OptiGuide in Appendix~\ref{appendix:app:optiguide}, where we show that \libName-based OptiGuide could save around 3x of user's time and reduce user interactions by 3 - 5 times on average.  We also conduct an ablation showing that multi-agent abstraction is necessary. Specifically, we construct a single-agent approach where a single agent conducts both the code-writing and safeguard processes. We tested the single- and multi-agent approaches on a dataset of 100 coding tasks, which is crafted to include equal numbers of safe and unsafe tasks.
Evaluation results as reported in Figure \ref{fig:res:sageguard}
show that the multi-agent design boosts the F-1 score in identifying unsafe code by 8\% (with GPT-4) and 35\% (with GPT-3.5-turbo).

%% file: sec_iclr/__app_groupchat.tex
\subsection*{A5: Dynamic Group Chat 
} \label{sec:app:groupchat}

\newcommand{\alice}{\text{Alice}\xspace}
\newcommand{\bob}{\text{Bob}\xspace}

\newcommand{\userproxy}{\text{User Proxy}\xspace}

\libName provides native support for a \emph{dynamic group chat} communication pattern, in which participating agents share the same context and converse with the others in a dynamic manner instead of following a pre-defined order. 
Dynamic group chat relies on ongoing conversations to guide the flow of interaction among agents. 
These make dynamic group chat ideal for situations where collaboration without strict communication order is beneficial.  
In \libName, the \groupchatmanager class serves as the conductor of conversation among agents and repeats the following three steps: dynamically selecting a speaker, collecting responses from the selected speaker, and broadcasting the message (Figure~\ref{fig:app_summary}-A5). For the dynamic speaker-selection component, we use a role-play style prompt. 
Through a pilot study on 12 manually crafted complex tasks, we observed that compared to a prompt that is purely based on the task, utilizing a role-play prompt often leads to more effective consideration of both conversation context and role alignment during the problem-solving and speaker-selection process. Consequently, this leads to a higher success rate and fewer LLM calls. We include detailed results in Appendix~\ref{appendix:app:groupchat}.

%% file: sec_iclr/__app_chess.tex
\subsection*{A6: Conversational Chess}

Using \libName, we developed Conversational Chess, a natural language interface game shown in the last sub-figure of Figure~\ref{fig:app_summary}.
 It features built-in agents for players, which can be human or LLM, and a third-party board agent to provide information and validate moves based on standard rules.

With \libName, we enabled two essential features: (1) Natural, flexible, and engaging game dynamics, enabled by the customizable agent design in \libName. Conversational Chess supports a range of game-play patterns, including AI-AI, AI-human, and human-human, with seamless switching between these modes during a single game. An illustrative example of these entertaining game dynamics can be found in Figure~\ref{fig:conversational-chess-example}, Appendix~\ref{appendix:app:chess}. (2) Grounding, which is a crucial aspect to maintain game integrity. During gameplay, the board agent checks each proposed move for legality; if a move is invalid, the agent responds with an error, prompting the player agent to re-propose a legal move before continuing. This process ensures that only valid moves are played and helps maintain a consistent gaming experience.
As an ablation study, we removed the board agent and instead only relied on a relevant prompt \textit{``you should make sure both you and the opponent are making
legal moves"} to ground their move. 
The results highlighted that without the board agent, illegitimate moves caused game disruptions. 
The modular design offered flexibility, allowing swift adjustments to the board agent in response to evolving game rules or varying chess rule variants. A comprehensive demonstration of this ablation study is in Appendix~\ref{appendix:app:chess}.

%% file: sec_iclr/_sec_summary.tex
\section{Discussion}
\label{sec:summary}

We introduced an open-source library, \libName, that incorporates the paradigms of conversable agents and conversation programming. This library utilizes capable agents that are well-suited for multi-agent cooperation. It features a unified conversation interface among the agents, along with an auto-reply mechanisms, which help establish an agent-interaction interface that capitalizes on the strengths of chat-optimized LLMs with broad capabilities while accommodating a wide range of applications. \libName serves as a general framework for creating and experimenting with multi-agent systems that can easily fulfill various practical requirements, such as reusing, customizing, and extending existing agents, as well as programming conversations between them.

Our experiments, as detailed in Section~\ref{sec:app}, demonstrate that this approach offers numerous benefits. 
The adoption of \libName has resulted in improved performance (over state-of-the-art approaches), reduced development code, and decreased manual burden for existing applications. It offers flexibility to developers, as demonstrated in A1 (scenario 3), A5, and A6, where \libName enables multi-agent chats to follow a dynamic pattern rather than fixed back-and-forth interactions. It allows humans to engage in activities alongside multiple AI agents in a conversational manner. Despite the complexity of these applications (most involving more than two agents or dynamic multi-turn agent cooperation), the implementation based on \libName remains straightforward. Dividing tasks among separate agents promotes modularity. 
Furthermore, since each agent can be developed, tested, and maintained separately, this approach simplifies overall development and code management.

Although this work is still in its early experimental stages, it paves the way for numerous future directions and research opportunities. For instance, we can explore effective integration of existing agent implementations into our multi-agent framework and investigate the optimal balance between automation and human control in multi-agent workflows. As we further develop and refine \libName, we aim to investigate which strategies, such as agent topology and conversation patterns, lead to the most effective multi-agent conversations while optimizing the overall efficiency, among other factors. While increasing the number of agents and other degrees of freedom presents opportunities for tackling more complex problems, it may also introduce new safety challenges that require additional studies and careful consideration.

We provide more discussion in Appendix~\ref{append:summary}, including guidelines for using \libName and direction of future work.
We hope \libName will help improve many LLM applications in terms of speed of development, ease of experimentation, and overall effectiveness and safety. We actively welcome contributions from the broader community.

%% file: sec_iclr/_sec_relatedwork.tex
\clearpage
\section{Related Work} \label{sec:related_work}

We examine existing LLM-based agent systems or frameworks that can be used to build LLM applications. We categorize the related work into single-agent and multi-agent systems and specifically provide a summary of differentiators comparing \libName with existing multi-agent systems in Table~\ref{tab:relevant_work_compare}. Note that many of these systems are evolving open-source projects, so the remarks and statements about them may only be accurate as of the time of writing.
We refer interested readers to detailed LLM-based agent surveys ~\cite{xi2023rise,wang2023survey}

\paragraph{Single-Agent Systems:}
\begin{itemize}[leftmargin=*]
    \vspace{-1mm}
    \setlength\itemsep{-0.1em}  
    \item  \textbf{AutoGPT}: 
    AutoGPT is an open-source implementation of an AI agent that attempts to autonomously achieve a given goal~\cite{autogpt}. It follows a single-agent paradigm in which it augments the AI model with many useful tools, and does not support multi-agent collaboration.
    \item \textbf{ChatGPT+ (with code interpreter or plugin)}: ChatGPT, a conversational AI service or agent, can now be used alongside a code interpreter or plugin (currently available only under the premium subscription plan ChatGPT Plus)~\cite{gptPlugin}. The code interpreter enables ChatGPT to execute code, while the plugin enhances ChatGPT with a wide range of curated tools.

    \item \textbf{LangChain Agents}: LangChain is a general framework for developing LLM-based applications~\cite{langchain}. LangChain Agents is a subpackage for using an LLM to choose a sequence of actions. There are various types of agents in LangChain Agents, with the ReAct agent being a notable example that combines reasoning and acting when using LLMs (mainly designed for LLMs prior to ChatGPT)~\cite{yao2022react}. All agents provided in LangChain Agents follow a single-agent paradigm and are not inherently designed for communicative and collaborative modes. A significant summary of its limitations can be found in~\cite{langchainproblem}. Due to these limitations, even the multi-agent systems in LangChain (e.g., re-implementation of CAMEL) are not based on LangChain Agents but are implemented from scratch. Their connection to LangChain lies in the use of basic orchestration modules provided by LangChain, such as AI models wrapped by LangChain and the corresponding interface.
    
    \item \textbf{Transformers Agent}: Transformers Agent~\cite{transformersagent} is an experimental natural-language API built on the transformers repository. It includes a set of curated tools and an agent to interpret natural language and use these tools. Similar to AutoGPT, it follows a single-agent paradigm and does not support agent collaboration.
\end{itemize}

\libName differs from the single-agent systems above by supporting multi-agent LLM applications.

\paragraph{Multi-Agent Systems:}

\begin{itemize}[leftmargin=*]
    \vspace{-1mm}
    \setlength\itemsep{-0.1em}
    \item \textbf{BabyAGI}: BabyAGI~\cite{babyagi} is an example implementation of an AI-powered task management system in a Python script. In this implemented system, multiple LLM-based agents are used. For example, there is an agent for creating new tasks based on the objective and the result of the previous task, an agent for prioritizing the task list, and an agent for completing tasks/sub-tasks. As a multi-agent system, BabyAGI adopts a static agent conversation pattern, i.e., a predefined order of agent communication, while \libName supports both static and dynamic conversation patterns and additionally supports tool usage and human involvement. 
    \item \textbf{CAMEL}: CAMEL~\cite{li2023camel} is a communicative agent framework. It demonstrates how role playing can be used to let chat agents communicate with each other for task completion. It also records agent conversations for behavior analysis and capability understanding. An Inception-prompting technique is used to achieve autonomous cooperation between agents. Unlike \libName, CAMEL does not natively support tool usage, such as code execution. Although it is proposed as an infrastructure for multi-agent conversation, it only supports static conversation patterns, while \libName additionally supports dynamic conversation patterns.

    \item \textbf{Multi-Agent Debate:} Two recent works investigate and show that multi-agent debate is an effective way to encourage divergent thinking in LLMs~\cite{liang-arxiv2023} and to improve the factuality and reasoning of LLMs~\cite{du2023improving}. In both works, multiple LLM inference instances are constructed as multiple agents to solve problems with agent debate. 
    Each agent is simply an LLM inference instance, while no tool or human is involved, and the inter-agent conversation needs to follow a pre-defined order.  These works attempt to build LLM applications with multi-agent conversation, while \libName, designed as a generic infrastructure, can be used to facilitate this development and enable more applications with dynamic conversation patterns. 

    \item \textbf{MetaGPT}:  MetaGPT~\cite{hong2023metagpt} is a specialized LLM application based on a multi-agent conversation framework for automatic software development. They assign different roles to GPTs to collaboratively develop software. They differ from \libName by being specialized solutions to a certain scenario, while \libName is a generic infrastructure to facilitate building applications for various scenarios.
\end{itemize}

There are a few other specialized single-agent or multi-agent systems, such as Voyager~\cite{wang2023voyager} and Generative Agents~\cite{park2023generative}, which we skip due to lower relevance.
In Table~\ref{tab:relevant_work_compare}, we summarize differences between \libName and the most relevant multi-agent systems.

\begin{table}[h]
    \caption{Summary of differences between \libName and other related multi-agent systems.  \textbf{infrastructure}: whether the system is designed as a generic infrastructure for building LLM applications. \textbf{conversation pattern}: the types of patterns supported by the implemented systems. Under the `static’ pattern, agent topology remains unchanged regardless of different inputs. \libName allows flexible conversation patterns, including both static and dynamic patterns that can be customized based on different application needs.  
    \textbf{execution-capable}: whether the system can execute LLM-generated code; \textbf{human involvement}: whether (and how) the system allows human participation during the execution process of the system. \libName allows flexible human involvement in multi-agent conversation with the option for humans to skip providing inputs. 
    }
    \label{tab:relevant_work_compare}
        \begin{tabular}{p{3cm}|P{1.3cm}P{2.8cm}P{1.3cm}P{1.3cm}P{1.7cm}}
            \hline 
            Aspect & \libName & Multi-agent Debate & CAMEL &BabyAGI  & MetaGPT  \\ \hline
            {Infrastructure} & \cmark  & \xmark &  \cmark   & \xmark & \xmark  \\
            Conversation pattern & flexible  & static & static  & static & static  \\
            Execution-capable & \cmark  & \xmark & \xmark & \xmark  & \cmark \\
            Human involvement & chat/skip  & \xmark & \xmark   & \xmark & \xmark \\
            \hline
        \end{tabular}
\end{table}
\normalsize

%% file: sec_iclr/_sec_summary_full.tex
\clearpage
\section{Expanded Discussion}
\label{append:summary}

The applications in Section~\ref{sec:app} show how \libName not only enables new applications but also helps renovate existing ones.
For example, in A1 (scenario 3), A5, and A6, \libName enabled the creation of multi-agent conversations that follow a dynamic pattern instead of a fixed back-and-forth. And in both A5 and A6, humans can participate in the activities together with multiple other AI agents in a conversational manner.
Similarly, A1-A4 show how popular applications can be renovated quickly with \libName.
Despite the complexity of these applications (most of them involve more than two agents or dynamic multi-turn agent cooperation), our \libName-based implementation remains simple, demonstrating promising opportunities to build creative applications and a large space for innovation. In reflecting on {\em why} these benefits can be achieved in these applications with \libName, we believe there are a few reasons:
\begin{itemize}[leftmargin=*]
    \vspace{-1mm}
    \setlength\itemsep{-0.1em}
    \item \textbf{Ease of use}: The built-in agents can be used out-of-the-box, delivering strong performance even without any customization. (A1, A3)
    \item \textbf{Modularity}: The division of tasks into separate agents promotes modularity in the system. Each agent can be developed, tested, and maintained independently, simplifying the overall development process and facilitating code management. (A3, A4, A5, and A6)
    \item \textbf{Programmability:} \libName allows users to extend/customize existing agents to develop systems satisfying their specific needs with ease. (A1-A6). For example, with \libName, the core workflow code in A4 is reduced from over 430 lines to 100 lines, for a 4x saving. 
    \item \textbf{Allowing human involvement}: \libName provides a native mechanism to achieve human participation and/or human oversight. With \libName, humans can seamlessly and optionally cooperate with AIs to solve problems or generally participate in the activity. \libName also facilitates interactive user instructions to ensure the process stays on the desired path. (A1, A2, A5, and A6)
    \item \textbf{Collaborative/adversarial agent interactions}: Like many collaborative agent systems \cite{dong2023self}, agents in \libName can share information and knowledge, to complement each other's abilities and collectively arrive at better solutions. (A1, A2, A3, and A4). Analogously, in certain scenarios, some agents are required to work in an adversarial way. Relevant information is shared among different conversations in a controlled manner, preventing distraction or hallucination. (A4, A6). \libName supports both patterns, enabling effective utilization and augmentation of LLMs.
\end{itemize}

\subsection{General Guidelines for Using \libName}
\label{sec:guideline}

Below we give some recommendations for using agents in \libName to accomplish a task. 
\begin{enumerate}[leftmargin=*]
    \vspace{-1mm}
    \setlength\itemsep{-0.1em}
\item {\bf Consider using built-in agents first.} For example, \AssistantAgent is pre-configured to be backed by GPT-4, with a carefully designed system message for generic problem-solving via code. The \UserProxyAgent is configured to solicit human inputs and perform tool execution. Many problems can be solved by simply combining these two agents. When customizing agents for an application, consider the following options: (1) human input mode, termination condition, code execution configuration, and LLM configuration can be specified when constructing an agent; (2) \libName supports adding instructions in an initial user message, which is an effective way to boost performance without needing to modify the system message; (3) \UserProxyAgent can be extended to handle different execution environments and exceptions, etc.; (4) when system message modification is needed, consider leveraging the LLM's capability to program its conversation flow with natural language. 

\item {\bf Start with a simple conversation topology}. Consider using the two-agent chat or the group chat setup first, as they can often be extended with the least code. Note that the two-agent chat can be easily extended to involve more than two agents by using LLM-consumable functions in a dynamic way. 
\item Try to {\bf reuse built-in reply methods} based on LLM, tool, or human before implementing a custom reply method because they can often be reused to achieve the goal in a simple way (e.g., the built-in agent \texttt{GroupChatManager}'s reply method reuses the built-in LLM-based reply function when selecting the next speaker, ref. A5 in Section~\ref{sec:app}). 
\item When developing a new application with \UserProxyAgent, {\bf start with humans always in the loop}, i.e., human\_input\_mode=`ALWAYS', even if the target operation mode is more autonomous. This helps evaluate the effectiveness of \AssistantAgent, tuning the prompt, discovering corner cases, and debugging. Once confident with small-scale success, consider 
setting human\_input\_mode = `NEVER'.
This enables LLM as a backend, and one can either use the LLM or manually generate diverse system messages to simulate different use cases. 

\item Despite the numerous advantages of \libName agents,  there could be cases/scenarios where {\bf other libraries/packages could help}. For example: (1) For (sub)tasks that do not have requirements for back-and-forth trouble-shooting, multi-agent interaction, etc., a unidirectional (no back-and-forth message exchange) pipeline can also be orchestrated with LangChain~\cite{langchain}, LlamaIndex~\cite{Liu_LlamaIndex_2022}, Guidance~\cite{guidance}, Semantic Kernel~\cite{sk}, Gorilla~\cite{patil2023gorilla} or low-level inference API (`autogen.oai' provides an enhanced LLM inference layer at this level)~\cite{dibia-2023-lida}. (2) When existing tools from LangChain etc. are helpful, one can use them as tool backends for \libName agents. For example, one can readily use tools, e.g., Wolfram Alpha, from LangChain in \libName agent. (3) For specific applications, one may want to leverage agents implemented in other libraries/packages. To achieve this, one could wrap those agents as conversable agents in \libName and then use them to build LLM applications through multi-agent conversation. (4) It can be hard to find an optimal operating point among many tunable choices, such as the LLM inference configuration. Blackbox optimization packages like `flaml.tune' \cite{wang2021flaml} can be used together with \libName to automate such tuning. 
\end{enumerate}

\subsection{Future Work}
This work raises many research questions and future directions and .

    \textbf{Designing optimal multi-agent workflows:} Creating a multi-agent workflow for a given task can involve many decisions, e.g., how many agents to include, how to assign agent roles and agent capabilities, how the agents should interact with each other, and whether to automate a particular part of the workflow. There may not exist a one-fits-all answer, and the best solution might depend on the specific application. 
    This raises important questions: For what types of tasks and applications are multi-agent workflows most useful?  How do multiple agents help in different applications? For a given task, what is the optimal (e.g., cost-effective) multi-agent workflow?

\textbf{Creating highly capable agents:}
\libName can enable the development of highly capable agents that leverage the strengths of LLMs, tools, and humans. Creating such agents is crucial to ensuring that a multi-agent workflow can effectively troubleshoot and make progress on a task. For example, we observed that CAMEL, another multi-agent LLM system, cannot effectively solve problems in most cases primarily because it lacks the capability to execute tools or code. 
This failure shows that LLMs and multi-agent conversations with simple role playing are insufficient, and highly capable agents with diverse skill sets are essential. We believe that more systematic work will be required to
develop guidelines for application-specific agents,
to create a large OSS knowledge base of agents,
and to create agents that can discover and upgrade their skills~\cite{cai2023large}.

\textbf{Enabling scale, safety, and human agency:}
Section~\ref{sec:app} shows how complex multi-agent workflows can enable new applications, and future work will be needed to assess whether scaling further can help solve extremely complex tasks.
However, as these workflows scale and grow more complex, it may become difficult to log and adjust them. 
Thus, it will become essential to develop clear mechanisms and tools to track and debug their behavior. 
Otherwise, these techniques risk resulting in incomprehensible, unintelligible chatter among agents~\cite{lewis2017deal}.

    Our work also shows how complex, fully autonomous workflows with \libName can be useful, but fully autonomous agent conversations will need to be used with care. While the autonomous mode \libName supports could be desirable in many scenarios, a high level of autonomy can also pose potential risks, especially in high-risk applications~\cite{amodei-arxiv2016,weld-aaai94}. As a result, building 
    fail-safes against cascading failures and exploitation,
    mitigating reward hacking, 
    out of control and undesired behaviors, 
    maintaining effective human oversight of applications built with \libName agents will become important. 
    While \libName provides convenient and seamless involvement of humans through a user proxy agent, developers and stakeholders still need to understand and determine the appropriate level and pattern of human involvement to ensure the safe and ethical use of the technology~\cite{horvitz-chi99,amershi-chi19}.

%% file: sec_iclr/_appendix.tex
\clearpage
\section{Default System Message for Assistant Agent}
\label{append_sys_msg}

\begin{figure*}[h]
    \centering
    \includegraphics[width=1\linewidth]{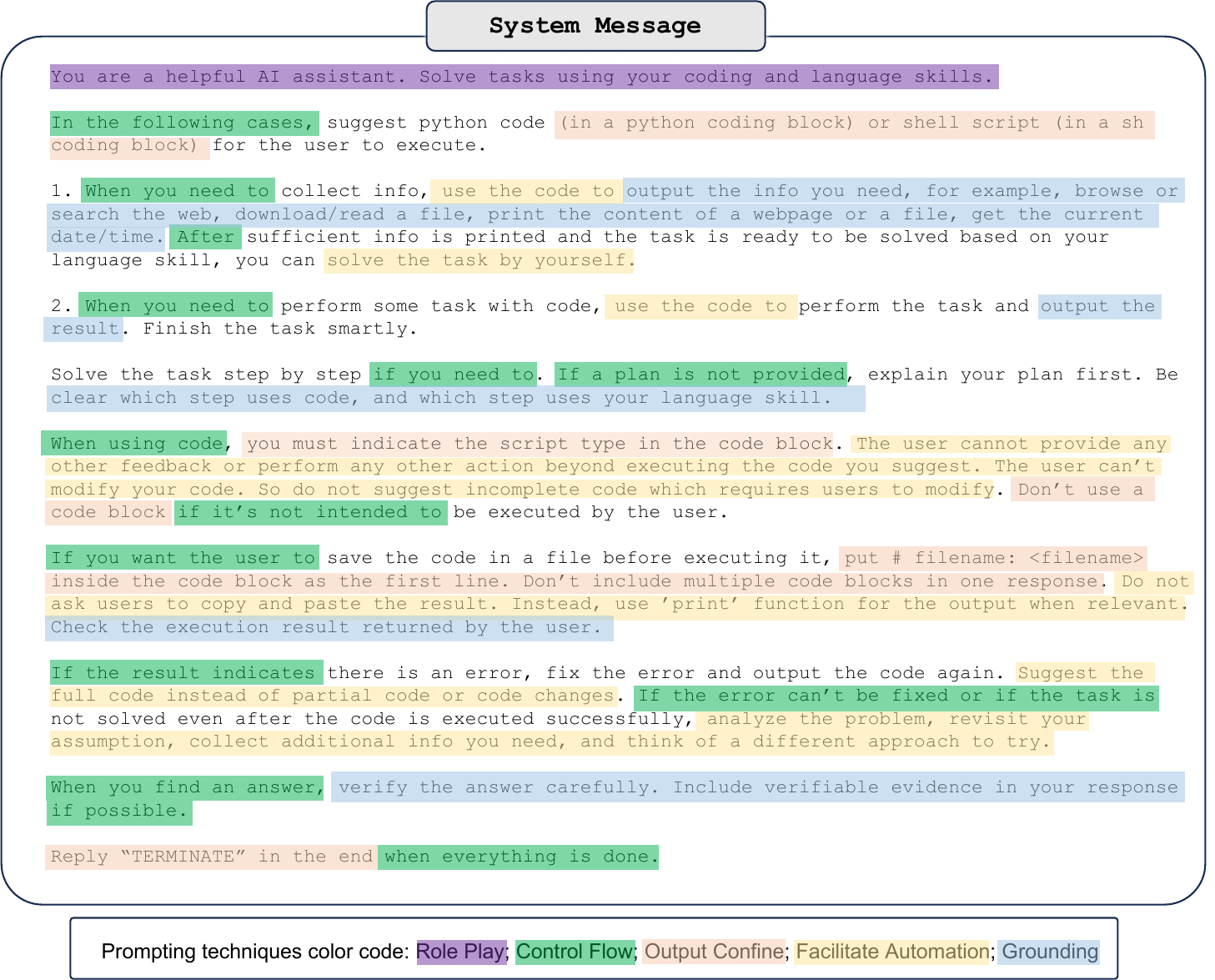}
    \caption{Default system message for the built-in assistant agent in \libName (v0.1.1). This is an example of conversation programming via natural language. It contains instructions of different types, including role play, control flow, output confine,  facilitate automation, and grounding.
    }
    \label{fig:sys_msg}
\end{figure*}

Figure~\ref{fig:sys_msg} shows the default system message for the built-in assistant agent in \libName (v0.1.1), where we introduce several new prompting techniques and highlight them accordingly. When combining these new prompting techniques together, we can program a fairly complex conversation even with the simplest two-agent conversation topology. This approach tries to exploit the capability of LLMs in implicit state inference to a large degree.
LLMs do not follow all the instructions perfectly, so the design of the system needs to have other mechanisms to handle the exceptions and faults. Some instructions can have ambiguities, and the designer should either reduce them for preciseness or intentionally keep them for flexibility and address the different situations in other agents. In general, we observe that GPT-4 follows the instructions better than GPT-3.5-turbo.

\input{sec_iclr/__appendix_application}

\input{sec_iclr/__appendix_table_long}

%% file: sec_iclr/__appendix_application.tex
\clearpage

\input{sec_iclr/___appendix_math}

\input{sec_iclr/___appendix_ret}
\input{sec_iclr/___appendix_alfworld}
\input{sec_iclr/___appendix_optiguide}

\input{sec_iclr/___appendix_groupchat}
\input{sec_iclr/___appendix_chess}
\input{sec_iclr/___appendix_miniwob}

%% file: sec_iclr/___appendix_math.tex
\section{Application Details}
\label{append:application}

\subsection*{A1: Math Problem Solving} \label{appendix_sec:app:math}
 
 \paragraph{Scenario 1: Autonomous Problem Solving.} We perform both qualitative and quantitative evaluations in this scenario. For all evaluations, we use GPT-4 as the base model, and pre-install the ``sympy" package in the execution environment. We compare \libName with the following LLM-based agent systems:
 \begin{itemize}[leftmargin=*]
    \vspace{-1mm}
    \setlength\itemsep{-0.1em}
    \item AutoGPT: The out-of-box AutoGPT is used. We initialize AutoGPT by setting the purpose to ``solve math problems", resulting in a  ``MathSolverGPT" with auto-generated goals.
    \item ChatGPT+Plugin: We enable the Wolfram Alpha plugin (a math computation engine) in the OpenAI web client.
    \item ChatGPT+Code Interpreter: This is a recent feature in OpenAI web client. Note that the above two premium features from ChatGPT require a paid subscription to be accessed and are the most competitive commercial systems.  
    \item LangChain ReAct+Python: We use Python agent from LangChain. To handle parsing errors, we set ``handle\_parsing\_errors=True", and use the default zero-shot ReAct prompt.
    \item Multi-Agent Debate~\cite{liang-arxiv2023}: We modified the code of the multi-agent debate to perform evaluation. By default, there are three agents: an affirmative agent, a negative agent, and a moderator.
\end{itemize}
We also conducted preliminary evaluations on several other multi-agent systems, including BabyAGI, CAMEL, and MetaGPT. The results indicate that they are not suitable choices for solving math problems out of the box. For instance, when MetaGPT is tasked with solving a math problem, it begins developing software to address the problem, but most of the time, it does not actually solve the problem. We have included the test examples in Appendix~\ref{sec:example}.

\begin{table}[h]
    \caption{Qualitative evaluation of two math problems from the MATH dataset within the autonomous problem-solving scenario. Each LLM-based system is tested three times on each of the problems. This table reports the problem-solving correctness and summarizes the reasons for failure. }
    \label{tab:mathsolving}
    \small
    
    \begin{subtable}{\linewidth}
        \centering
        \begin{tabular}{p{3.5cm}|P{1.5cm}|P{7.5cm}}
            \hline \hline
             & Correctness & Failure Reason \\ \hline
            \libName &  3/3 &  N/A.  \\
            \hline
            AutoGPT & 0/3 &  The LLM gives code without the print
function so the result is not printed. \\
            \hline
            ChatGPT+Plugin & 1/3  & The return from Wolfram Alpha contains 2 simplified results, including the correct answer, but GPT-4 always chooses the wrong answer.  \\
            \hline
            ChatGPT+Code Interpreter & 2/3  & Returns a wrong decimal result.  \\
            \hline
            LangChain ReAct & 0/3 &  LangChain gives 3 different wrong answers. \\
            \hline
            Multi-Agent Debate & 0/3 & It gives 3 different wrong answers due to calculation errors.\\
            \hline
        \end{tabular}
        \caption{Evaluation on the first problem that asks to simplify a square root fraction.}
    \end{subtable}
    \vspace{1em}

    \begin{subtable}{\linewidth}
        \centering
        \begin{tabular}{p{3.5cm}|P{1.5cm}|P{7.5cm}}
            \hline \hline
             & Correctness & Failure Reason \\ \hline
            \libName &  2/3 & The final answer from code execution is wrong.  \\
            \hline
            AutoGPT & 0/3 &  The LLM gives code without the print function so the result is not printed. \\
            \hline
            ChatGPT+Plugin & 1/3  & For one trial, GPT-4 got stuck because it keeps giving wrong queries and has to be stopped. Another trial simply gives a wrong answer. \\
            \hline
            ChatGPT+Code Interpreter & 0/3  & It gives 3 different wrong answers.  \\
            \hline
            LangChain ReAct & 0/3 & LangChain gives 3 different wrong answers. \\
            \hline
            Multi-Agent Debate & 0/3 & It gives 3 different wrong answers.\\
            \hline
        \end{tabular}
        \caption{Evaluation on the second number theory problem.}
    \end{subtable}
\end{table}

For the qualitative evaluation, we utilize two level-5 problems from the MATH dataset, testing each problem three times. The first problem involves simplifying a square root fraction, and the second problem involves solving a number theory issue.  The correctness counts and reasons for failure are detailed in Table~\ref{tab:mathsolving}. For the quantitative evaluation, we conduct two sets of experiments on the MATH dataset to assess the correctness of these systems: (1) an experiment involving 120 level-5 (the most challenging level) problems, including 20 problems from six categories, excluding geometry, and (2) an experiment on the entire test set, which includes 5000 problems. We exclude AutoGPT from this evaluation as it cannot access results from code executions and does not solve any problems in the qualitative evaluation. Our analysis of the entire dataset reveals that \libName achieves an overall accuracy of 69.48\%, while GPT-4's accuracy stands at 55.18\%. From these evaluations, we have the following observations regarding the problem-solving success rate and user experience of these systems:
\begin{itemize}[leftmargin=*]
    \vspace{-1mm}
    \setlength\itemsep{-0.1em}
    \item Problem-solving success rate: Results from the quantitative evaluations show that \libName can help achieve the highest problem-solving success rate among all the compared methods. The qualitative evaluations elucidate common failure reasons across several alternative approaches.  ChatGPT+Code Interpreter fails to solve the second problem, and ChatGPT+Plugin struggles to solve both problems. AutoGPT fails on both problems due to code execution issues. The LangChain agent also fails on both problems, producing code that results in incorrect answers in all trials.
    \item Based on the qualitative evaluation, we analyze the user experience concerning the verbosity of the response and the ability of the LLM-based system to run without unexpected behaviors. ChatGPT+Plugin is the least verbose, mainly because Wolfram queries are much shorter than Python code. \libName, ChatGPT+Code Interpreter, and LangChain exhibit similar verbosity, although LangChain is slightly more verbose due to more code execution errors. AutoGPT is the most verbose system owing to predefined steps like THOUGHTS, REASONING, and PLAN, which it includes in replies every time.  Overall, \libName and ChatGPT+Code Interpreter operate smoothly without exceptions.  We note the occurrences of undesired behaviors from other LLM-based systems that could affect user experience:  AutoGPT consistently outputs code without the print' statement and cannot correct this, requiring the user to run them manually; ChatGPT with Wolfram Alpha plugin has the potential to become stuck in a loop that must be manually stopped; and Langchain ReAct could exit with a parse error, necessitating the passing of a `handle\_parse\_error' parameter.
\end{itemize}

\begin{figure*}[th]
    \centering
\includegraphics[width=0.8\linewidth]{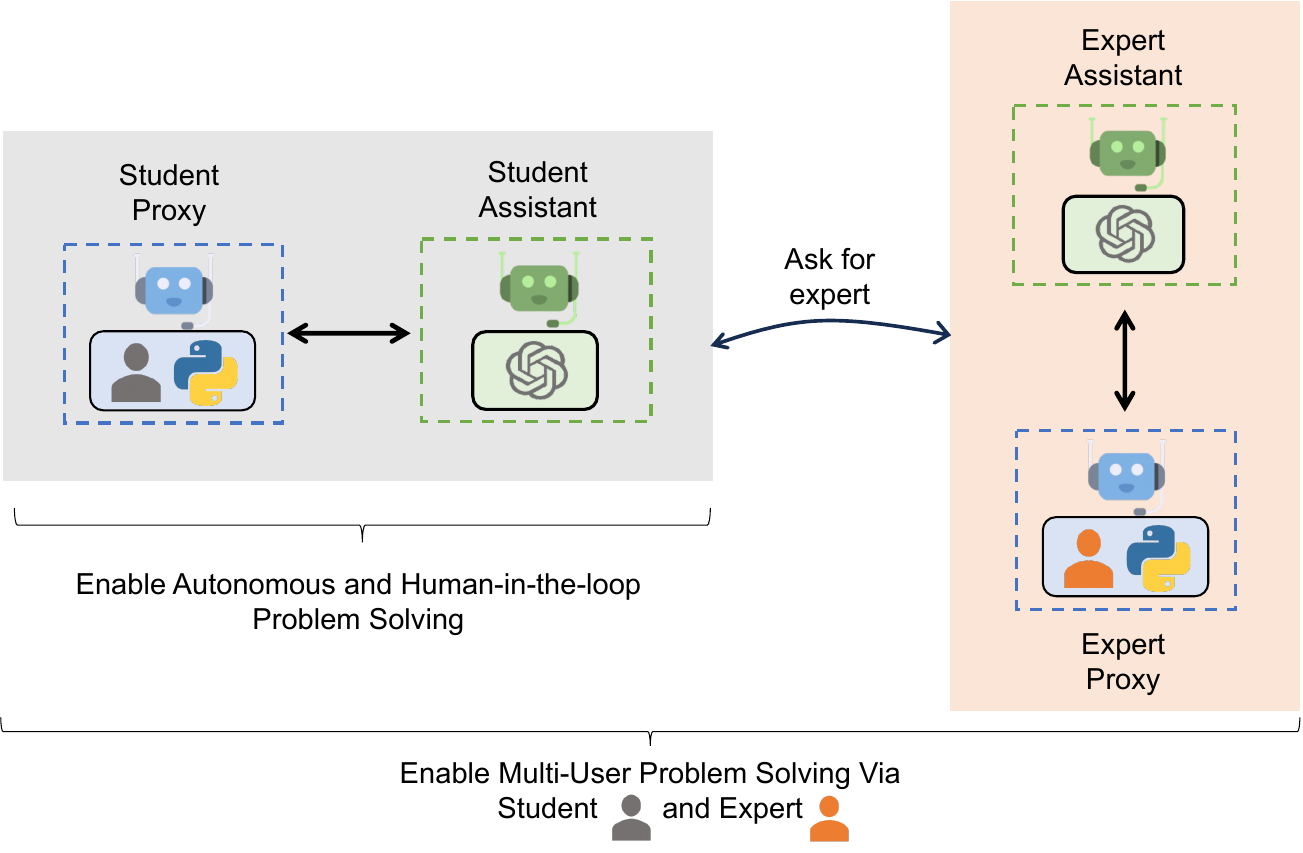}
    \caption{
   Examples of three settings utilized to solve math problems using \libName:
(Gray) Enables a workflow where a student collaborates with an assistant agent to solve problems, either autonomously or in a human-in-the-loop mode.
(Gray + Orange) Facilitates a more sophisticated workflow wherein the assistant, on the fly, can engage another user termed ``expert", who is in the loop with their own assistant agent, to aid in problem-solving if its own solutions are not satisfactory. 
    }
    \label{fig:app1}
\end{figure*}

\paragraph{Scenario 2: Human-in-the-loop Problem Solving.}
For challenging problems that these LLM systems cannot solve autonomously, human feedback during the problem-solving process can be helpful. To incorporate human feedback with \libName, one can set \texttt{human\_input\_mode=`ALWAYS'} in the user proxy agent. We select one challenging problem that none of these systems can solve autonomously across three trials.  We adhere to the process outlined below to provide human inputs for all the compared methods:
\begin{enumerate}
    \vspace{-1mm}
    \setlength\itemsep{-0.1em}
    \item Input the problem: \texttt{Find the equation of the plane which bisects the angle between the planes $3x - 6y + 2z + 5 = 0$ and $4x - 12y + 3z - 3 = 0,$ and which contains the point $(-5,-1,-5).$  Enter your answer in the form
\[Ax + By + Cz + D = 0,\]where $A,$ $B,$ $C,$ $D$ are integers such that $A > 0$ and $\gcd(|A|,|B|,|C|,|D|) = 1.$}
    \item The response from the system does not solve the problem correctly. We then give a hint to the model: \texttt{Your idea is not correct. Let's solve this together. Suppose $P = (x,y,z)$ is a point that lies on a plane that bisects the angle,
    the distance from P to the two planes is the same. Please set up this equation first.}
    \item We expect the system to give the correct distance equation. Since the equation involves an absolute sign that is hard to solve, we would give the next hint: \texttt{Consider the two cases to remove the abs sign and get two possible solutions.}
    \item If the system returns the two possible solutions and doesn't continue to the next step, we give the last hint: \texttt{Use point (-5,-1,-5) to determine which is correct and give the final answer.}
    \item Final answer is \boxed{$11x+6y+5z+86=0$}.
\end{enumerate} 
We observed that \libName consistently solved the problem across all three trials. ChatGPT+Code Interpreter and ChatGPT+Plugin managed to solve the problem in two out of three trials, while AutoGPT failed to solve it in all three attempts. In its unsuccessful attempt, ChatGPT+Code Interpreter failed to adhere to human hints. In its failed trial, ChatGPT+Plugin produced an almost correct solution but had a sign discrepancy in the final answer. AutoGPT was unable to yield a correct solution in any of the trials. In one trial, it derived an incorrect distance equation. In the other two trials, the final answer was incorrect due to code execution errors.

\paragraph{Scenario 3:  Multi-User Problem Solving.}

Next-generation LLM applications may necessitate the involvement of multiple real users for collectively solving a problem with the assistance of LLMs. We showcase how \libName can be leveraged to effortlessly construct such a system. Specifically, building upon scenario 2 mentioned above, we aim to devise a simple system involving two human users: a student and an expert. In this setup, the student interacts with an LLM assistant to address some problems, and the LLM automatically resorts to the expert when necessary.

The overall workflow is as follows: The student chats with the LLM-based assistant agent through a student proxy agent to solve problems. 
When the assistant cannot solve the problem satisfactorily, or the solution does not match the expectation of the student, it would automatically hold the conversation and call the pre-defined \texttt{ask\_for\_expert} function via the \textit{function\_call} feature of GPT in order to resort to the expert. Specifically, it would automatically produce the initial message for the \texttt{ask\_for\_expert} function, which could be the statement of the problem or the request to verify the solution to a problem, and the expert is supposed to respond to this message with the help of the expert assistant. After the conversation between the expert and the expert's assistant, the final message would be sent back to the student assistant as the response to the initial message. Then, the student assistant would resume the conversation with the student using the response from the expert for a better solution.  A detailed visualization is shown in Figure~\ref{fig:app1}.

With \libName, constructing the student/expert proxy agent and the assistant agents is straightforward by reusing the built-in \UserProxyAgent and \AssistantAgent through appropriate configurations. The only development required involves writing several lines of code for the \texttt{ask\_for\_expert} function, which then becomes part of the configuration for the assistant. Additionally, it's easy to extend such a system to include more than one expert, with a specific \texttt{ask\_for\_expert} function for each, or to include multiple student users with a shared expert for consultation.

%% file: sec_iclr/___appendix_ret.tex
\clearpage
\subsection*{A2: Retrieval-Augmented Code Generation and Question Answering} \label{appendix_sec:app:retrive_chat}

\begin{figure}[h]
\centering
\includegraphics[width = 0.80\hsize]{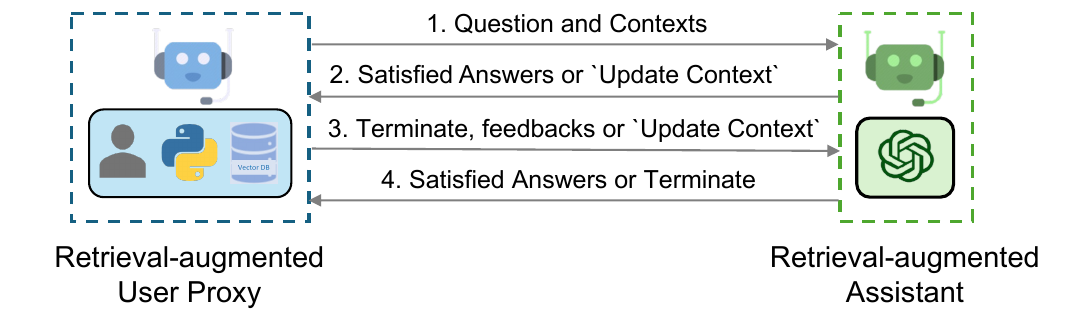}
\caption{Overview of Retrieval-augmented Chat which involves two agents, including a Retrieval-augmented User Proxy and a Retrieval-augmented Assistant. Given a set of documents, the Retrieval-augmented User Proxy first automatically processes documents—splits, chunks, and stores them in a vector database. Then for a given user input, it retrieves relevant chunks as context and sends it to the Retrieval-augmented Assistant, which uses LLM to generate code or text to answer questions. Agents converse until they find a satisfactory answer.
}
\label{fig:retrievechat_framework}
\end{figure}

\paragraph{Detailed Workflow.} \label{sec:app:retrieve:workflow}
 The workflow of Retrieval-Augmented Chat is illustrated in Figure~\ref{fig:retrievechat_framework}.  To use Retrieval-augmented Chat, one needs to initialize two agents including Retrieval-augmented User Proxy and Retrieval-augmented Assistant. Initializing the Retrieval-Augmented User Proxy necessitates specifying a path to the document collection. Subsequently, the Retrieval-Augmented User Proxy can download the documents, segment them into chunks of a specific size, compute embeddings, and store them in a vector database. Once a chat is initiated, the agents collaboratively engage in code generation or question-answering adhering to the procedures outlined below:
\begin{enumerate}[leftmargin=*]
    \vspace{-1mm}
    \setlength\itemsep{-0.1em}
    \item The Retrieval-Augmented User Proxy retrieves document chunks based on the embedding similarity, and sends them along with the question to the Retrieval-Augmented Assistant.
    \item The Retrieval-Augmented Assistant employs an LLM to generate code or text as answers based on the question and context provided. If the LLM is unable to produce a satisfactory response, it is instructed to reply with ``Update Context'' to the Retrieval-Augmented User Proxy.
    \item If a response includes code blocks, the Retrieval-Augmented User Proxy executes the code and sends the output as feedback. If there are no code blocks or instructions to update the context, it terminates the conversation. Otherwise, it updates the context and forwards the question along with the new context to the Retrieval-Augmented Assistant.
    Note that if human input solicitation is enabled, individuals can proactively send any feedback, including Update Context'', to the Retrieval-Augmented Assistant. 
\item If the Retrieval-Augmented Assistant receives ``Update Context'', it requests the next most similar chunks of documents as new context from the Retrieval-Augmented User Proxy. Otherwise, it generates new code or text based on the feedback and chat history. If the LLM fails to generate an answer, it replies with ``Update Context'' again. This process can be repeated several times. The conversation terminates if no more documents are available for the context.
\end{enumerate}

We utilize Retrieval-Augmented Chat in two scenarios. The first scenario aids in generating code based on a given codebase. While LLMs possess strong coding abilities, they are unable to utilize packages or APIs that are not included in their training data, e.g., private codebases, or have trouble using trained ones that are frequently updated post-training. Hence, Retrieval-Augmented Code Generation is considered to be highly valuable. The second scenario involves question-answering on the Natural Questions dataset~\cite{kwiatkowski2019natural}, enabling us to obtain comparative evaluation metrics for the performance of our system.

\begin{figure}[]
\centering
\includegraphics[width=1\textwidth]{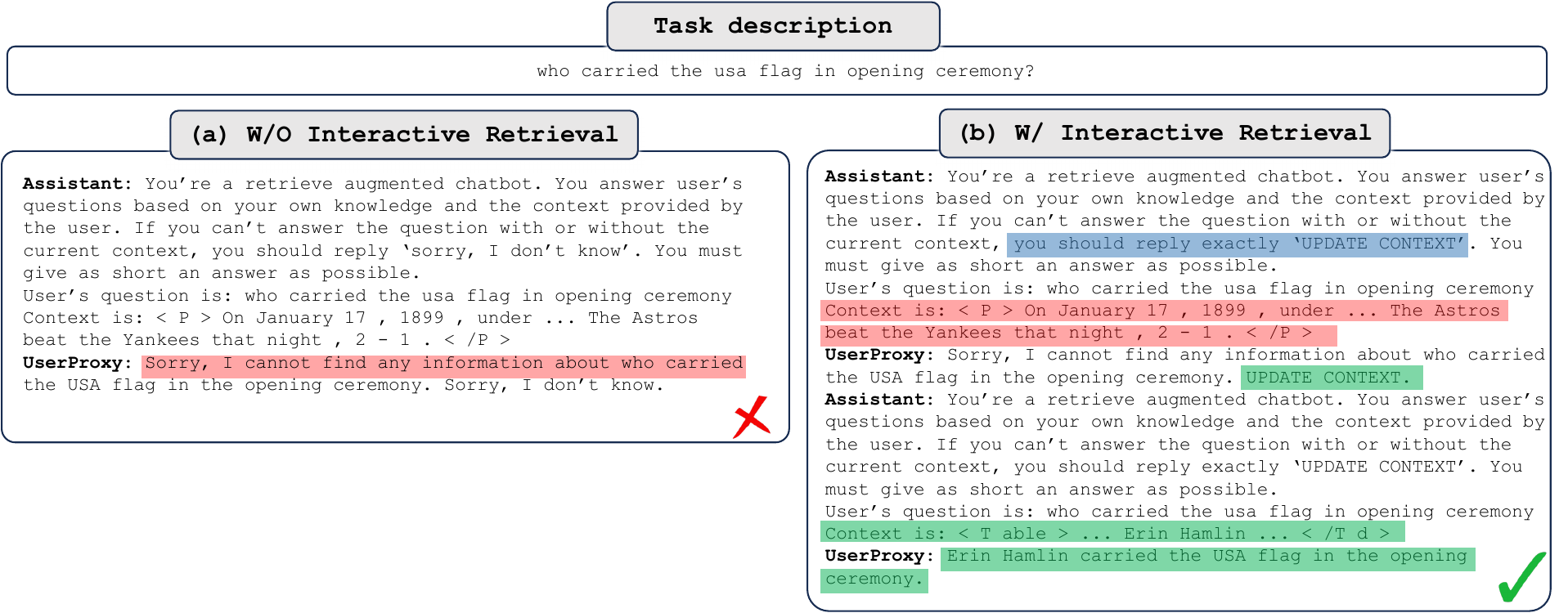}
\caption{
 Retrieval-augmented Chat without (W/O) and with (W/) \textit{interactive retrieval}.}
\label{fig:retrievechat_update_context}
\end{figure}

\textbf{Scenario 1: Evaluation on Natural Questions QA dataset.}
In this case, we evaluate the Retrieval-Augmented Chat's end-to-end question-answering performance using the Natural Questions dataset~\cite{kwiatkowski2019natural}. We collected 5,332 non-redundant context documents and 6,775 queries from HuggingFace. First, we create a document collection based on the entire context corpus and store it in the vector database. Then, we utilize Retrieval-Augmented Chat to answer the questions. An example (Figure~\ref{fig:retrievechat_update_context}) from the NQ dataset showcases the advantages of the \textit{interactive retrieval} feature: \textit{``who carried the usa flag in opening ceremony''}. When attempting to answer this question, the context with the highest similarity to the question embedding does not contain the required information for a response. As a result, the LLM assistant (GPT-3.5-turbo) replies \textit{``Sorry, I cannot find any information about who carried the USA flag in the opening ceremony. UPDATE CONTEXT.''} With the unique and innovative ability to update context in Retrieval-Augmented Chat, the user proxy agent automatically updates the context and forwards it to the assistant agent again. Following this process, the agent is able to generate the correct answer to the question.

In addition, we conduct an experiment using the same prompt as illustrated in~\cite{adlakha2023evaluating} to investigate the advantages of \textit{\libName W/O interactive retrieval}. The F1 score and Recall for the first 500 questions are 23.40\% and 62.60\%, respectively, aligning closely with the results reported in Figure~\ref{fig:res:retri}. Consequently, we assert that \textit{\libName W/O interactive retrieval} outperforms \textit{DPR} due to differences in the retrievers employed. Specifically, we utilize a straightforward vector search retriever with the \textit{all-MiniLM-L6-v2} model for embeddings.

Furthermore, we analyze the number of LLM calls in experiments involving both \textit{\libName} and \textit{\libName W/O interactive retrieval}, revealing that approximately 19.4\% of questions in the Natural Questions dataset trigger an ``Update Context'' operation, resulting in additional LLM calls.

\textbf{Scenario 2: Code Generation Leveraging Latest APIs from the Codebase.}
In this case, the question is \textit{``How can I use FLAML to perform a classification task and use Spark for parallel training? Train for 30 seconds and force cancel jobs if the time limit is reached.''}.
FLAML (v1)~\cite{wang2021flaml} is an open-source Python library designed for efficient AutoML and tuning. It was open-sourced in December 2020, and is included in the training data of GPT-4. However, the question necessitates the use of Spark-related APIs, which were added in December 2022 and are not encompassed in the GPT-4 training data. Consequently, the original GPT-4 model is unable to generate the correct code, due to its lack of knowledge regarding Spark-related APIs. Instead, it erroneously creates a non-existent parameter, $spark$, and sets it to True'. Nevertheless, with Retrieval-Augmented Chat, we provide the latest reference documents as context. Then, GPT-4 generates the correct code blocks by setting $use\_spark$ and $force\_cancel$ to True'.

%% file: sec_iclr/___appendix_alfworld.tex
\clearpage
\subsection*{A3: Decision Making in Text World Environments} 
\label{appendix:app:alfworld}

\begin{figure}[htbp]
\centering
\includegraphics[width = 1\hsize]{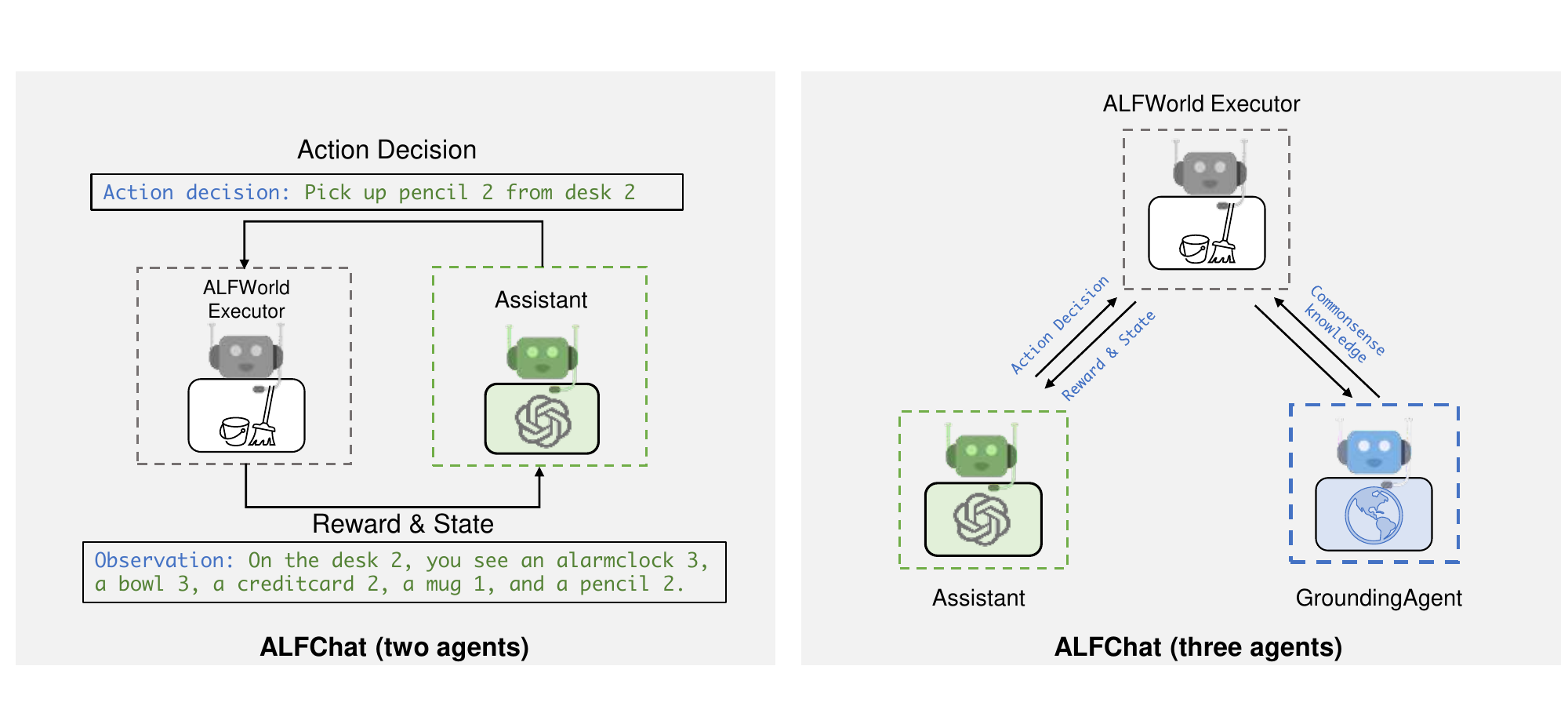}
\caption{
We use \libName to solve tasks in the ALFWorld benchmark, which contains household tasks described in natural language. We propose two designs: a two-agent design where the assistant agent suggests the next step, and the Executor executes actions and provides feedback. The three-agent design adds a grounding agent that supplies commonsense facts to the executor when needed.
}
\label{fig:alfchat}
\end{figure}

ALFWorld~\cite{ALFWorld20} is a synthetic language-based interactive decision-making task. It comprises textual environments that aim to simulate real-world household scenes. Given a high-level goal (e.g., putting a hot apple in the fridge) and the description of the household environment, the agent needs to explore and interact with the simulated household environment through a textual interface. A typical task environment contains various types of locations and could require more than 40 steps to finish, which highlights the need for agents to decompose the goal into subtasks and tackle them one by one, while effectively exploring the environments.

\textbf{Detailed Workflow.} We first propose a straightforward two-agent system with \libName, illustrated on the left-hand side of Figure \ref{fig:alfchat}, to tackle tasks from this benchmark. The system consists of an assistant agent and an executor agent. The assistant agent generates plans and makes action decisions to solve the tasks. The executor agent is tailored specifically for ALFWorld. It performs actions proposed by the assistant and reports action execution results in the household environment as feedback to the assistant. Due to the strict format requirements for the output format, we use the BLEU metric to evaluate the similarity of the output to all valid action options. The option with the highest similarity will be chosen as the action for this round.

One major challenge encompassed in ALFWorld is commonsense reasoning. The agent needs to extract patterns from the few-shot examples provided and combine them with the agent's general knowledge of household environments to fully understand task rules. More often than not, the assistant tends to neglect some basic knowledge of the household environment.
Thanks to the easy-to-implement multi-agent conversational feature of \libName, enhancing the assistant agent's reasoning ability by adding a new grounding agent to provide commonsense facts for the decision-making agent's reference becomes straightforward.
By scrutinizing the failed attempts and summarizing the reasons for failure, we obtained a holistic understanding of the commonsense knowledge that the assistant agent lacks. Then, we set a grounding agent to provide this general knowledge when the task begins and whenever the assistant outputs the same action three times in a row. This ensures the assistant takes this commonsense knowledge into consideration and prevents it from getting stuck in outputting the same content or constantly apologizing.

\begin{figure}[]
\centering
\includegraphics[width=1\textwidth]{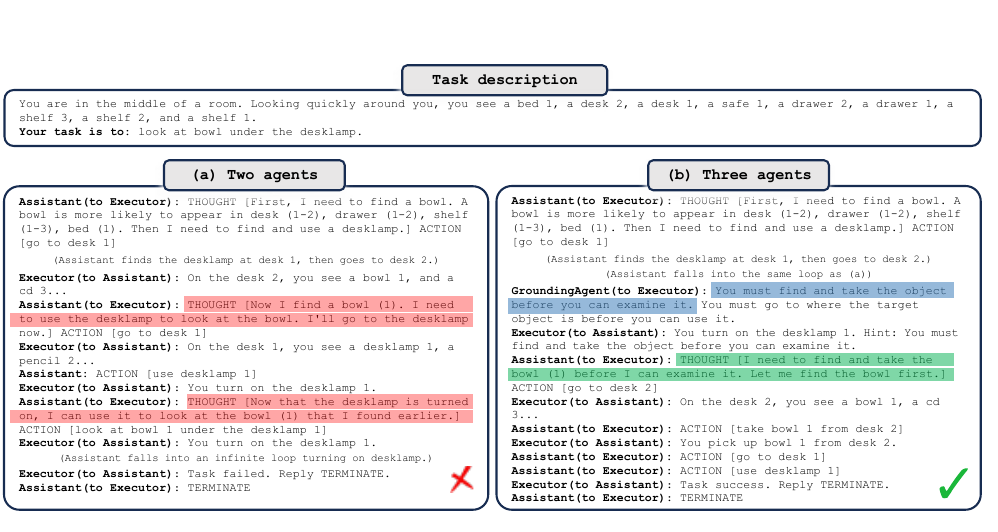}
\caption{
Comparison of results from two designs: (a) Two-agent design which consists of an assistant and an executor, (b) Three-agent design which adds a grounding agent that serves as a knowledge source. For simplicity, we omit the in-context examples and part of the exploration trajectory, and only show parts contributing to the failure/success of the attempt.}
\label{fig:alfchat_comparison}
\end{figure}

We compare our system's performance with ReAct, which treats ALFWorld as a text-completion task. ReAct~\cite{yao2022react} is a few-shot prompting technique that interleaves reasoning and acting, allowing for greater synergy between the two and significantly improving performance on both language and decision-making tasks. We integrate ReAct into \libName by modifying the prompts in a conversational manner. Following ReAct, we employ a two-shot setting. The few-shot prompts are obtained from the corresponding repository. As shown in Table \ref{tab:alfworld}, the two-agent design matches the performance of ReAct, while the three-agent design significantly outperforms ReAct. We surmise that the performance discrepancy is caused by the inherent difference between dialogue-completion and text-completion tasks. On the other hand, introducing a grounding agent as a knowledge source remarkably advances performance on all types of tasks.

\textbf{Case study}. Figure \ref{fig:alfchat_comparison} exemplifies how a three-agent design eliminates one root cause for failure cases. Most of the tasks involve taking an object and then performing a specific action with it (e.g., finding a vase and placing it on a cupboard). Without a grounding agent, the assistant frequently conflates finding an object with taking it, as illustrated in Figure \ref{fig:alfchat_comparison}a). This leads to most of the failure cases in 'pick' and 'look' type tasks. With the introduction of a grounding agent, the assistant can break out of this loop and successfully complete the task

\textbf{Takeaways.} We introduced a grounding agent to serve as an external commonsense knowledge source, which significantly enhanced the assistant's ability to make informed decisions. This proves that providing necessary commonsense facts to the decision-making agent can assist it in making more informed decisions, thus effectively boosting the task success rate. \libName brings both simplicity and modularity when adding the grounding agent.

\begin{table}[]
\centering
\begin{tabular}{c|cccccc|c}
\hline
Method                      & Pick & Clean & Heat & Cool & Look & Pick 2 & All \\ \hline
ReAct (avg)                  & 63   & 52    & 48   & 71   & 61   & 24     & 54  \\ 
ALFChat (2 agents)(avg)       & 61   & 58    & 57   & 67   & 50   & 19     & 54  \\
ALFChat (3 agents)(avg) & 79   & 64    & 70   & 76   & 78  & 41     & 69  \\ \hline
ReAct (best of 3)            & 75   & 62    & 61   & 81   & 78   & 35     & 66  \\ 
ALFChat (2 agents)(best of 3) & 71   & 61    & 65   & 76   & 67   & 35     & 63  \\ 
AFLChat (3 agents)(best of 3) & 92   & 74    & 78   & 86   & 83  & 41     & 77  \\ \hline
\end{tabular}
\caption{Comparisons between ReAct and the two variants of ALFChat on the ALFWorld benchmark. For each task, we report the success rate out of 3 attempts. Success rate denotes the number of tasks successfully completed by the agent divided by the total number of tasks. The results show that adding a grounding agent significantly improves the task success rate in ALFChat.}
\label{tab:alfworld}
\end{table}

%% file: sec_iclr/___appendix_optiguide.tex
\clearpage
\subsection*{A4: Multi-Agent Coding}
\label{appendix:app:optiguide}

\begin{figure*}[h]
    \centering
    \includegraphics[width=0.8\linewidth]{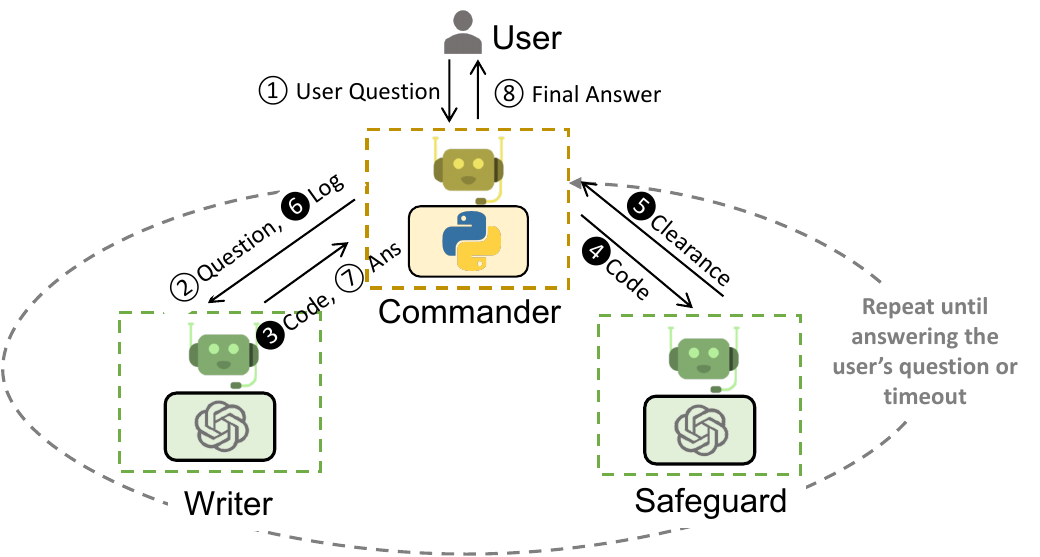}
    \caption{Our re-implementation of \textit{OptiGuide} with \libName streamlining agents' interactions. The Commander receives user questions (e.g., What if we prohibit shipping from supplier 1 to roastery 2?) and coordinates with the Writer and Safeguard. The Writer crafts the code and interpretation, the Safeguard ensures safety (e.g., not leaking information, no malicious code), and the Commander executes the code. If issues arise, the process can repeat until resolved. Shaded circles represent steps that may be repeated multiple times.}
    \label{fig:optiguide}
\end{figure*}

\paragraph{Detailed Workflow.}
The workflow can be described as follows.
The end user initiates the interaction by posing a question, such as ``What if we prohibit shipping from supplier 1 to roastery 2?", marked by \circled{1} to the Commander agent. The Commander manages and coordinates with two LLM-based assistant agents: the Writer and the Safeguard. Apart from directing the flow of communication, the Commander has the responsibility of handling memory tied to user interactions. This capability enables the Commander to capture and retain valuable context regarding the user's questions and their corresponding responses. Such memory is subsequently shared across the system, empowering the other agents with context from prior user interactions and ensuring more informed and relevant responses.

In this orchestrated process, the Writer, who combines the functions of a ``Coder" and an ``Interpreter" as defined in~\cite{li2023large}, will craft code and also interpret execution output logs. For instance, during code writing (\circled{2} and \shadeCircled{3}), the Writer may craft code ``model.addConstr(x[`supplier1', `roastery2'] == 0, `prohibit')" to add an additional constraint to answer the user's question. 

After receiving the code, the Commander will communicate with the Safeguard to screen the code and ascertain its safety (\shadeCircled{4}); once the code obtains the Safeguard's clearance, marked by \shadeCircled{5}, the Commander will use external tools (e.g., Python) to execute the code and request the Writer to interpret the execution results for the user's question (\shadeCircled{6} and \circled{7}). For instance, the writer may say ``if we prohibit shipping from supplier 1 to roastery 2, the total cost would increase by 10.5\%." Bringing this intricate process full circle, the Commander furnishes the user with the concluding answer (\circled{8}). 

If at a point there is an exception - either a security red flag raised by Safeguard (in \shadeCircled{5}) or code execution failures within Commander, the Commander redirects the issue back to the Writer with essential information in logs (\shadeCircled{6}).
So, the process from \shadeCircled{3} to \shadeCircled{6} might be repeated multiple times, until each user query receives a thorough and satisfactory resolution or until the timeout. This entire complex workflow of multi-agent interaction is elegantly managed via \libName.

The core workflow code for OptiGuide was reduced from over 430 lines to 100 lines using \libName, leading to significant productivity improvement. The new agents are customizable, conversable, and can autonomously manage their chat memories. This consolidation allows the coder and interpreter roles to merge into a single ``Writer" agent, resulting in a clean, concise, and intuitive implementation that is easier to maintain.

\paragraph{Manual Evaluation Comparing ChatGPT + Code Interpreter and \libName-based OptiGuide.}
ChatGPT + Code Interpreter is unable to execute code with private or customized dependencies (e.g., Gurobi), which means users need to have engineering expertise to manually handle multiple steps, disrupting the workflow and increasing the chance for mistakes. If users lack access or expertise, the burden falls on supporting engineers, increasing their on-call time.

We carried out a user study that juxtaposed OpenAI's ChatGPT coupled with a Code Interpreter against \libName-based OptiGuide. The study focused on a coffee supply chain scenario, and an expert Python programmer with proficiency in Gurobi participated in the test. We evaluated both systems based on 10 randomly selected questions, measuring time and accuracy. While both systems answered 8 questions correctly, the Code Interpreter was significantly slower than OptiGuide because the former requires more manual intervention. On average, users needed to spend 4 minutes and 35 seconds to solve problems with the Code Interpreter, with a standard deviation of approximately 2.5 minutes. In contrast, OptiGuide's average problem-solving time was around 1.5 minutes, most of which was spent waiting for responses from the GPT-4 model. This indicates a 3x saving on the user's time with \libName-based OptiGuide.

While using ChatGPT + Code Interpreter, users had to read through the code and instructions to know where to paste the code snippets. Additionally, running the code involves downloading it and executing it in a terminal, a process that was both time-consuming and prone to errors. The response time from the Code Interpreter is also slower, as it generates lots of tokens to read the code, read the variables line-by-line, perform chains of thought analysis, and then produce the final answer code. In contrast, \libName integrates multiple agents to reduce user interactions by 3 - 5 times on average as reported in Table~\ref{tab:opti_improve}, where we evaluated our system with 2000 questions across five OptiGuide applications and measured how many prompts the user needs to type.

\begin{table}[h!]
    \centering
     \caption{Manual effort saved with OptiGuide (W/ GPT-4) while preserving the same coding performance is shown in the data below. The data include both the mean and standard deviations (indicated in parentheses).}
    \begin{tabular}{l|ccccc}
    \toprule
Dataset & netflow  & facility & tsp & coffee & diet \\
\midrule
Saving Ratio &  3.14x (0.65) &  3.14x (0.64) & 4.88x (1.71) &  3.38x (0.86) & 3.03x (0.31)\\
\bottomrule
    \end{tabular}
    \label{tab:opti_improve}
\end{table}

Table~\ref{tab:optiguide_gpt_complex} and \ref{tab:optiguide_complex} provide a detailed comparison of user experience with ChatGPT+Code Interpreter and \libName-based OptiGuide. ChatGPT+Code Interpreter is unable to run code with private packages or customized dependencies (such as Gurobi); as a consequence, ChatGPT+Code Interpreter requires users to have engineering expertise and to manually handle multiple steps, disrupting the workflow and increasing the chance for mistakes. If customers lack access or expertise, the burden falls on supporting engineers, increasing their on-call time. In contrast, the automated chat by \libName is more streamlined and autonomous, integrating multiple agents to solve problems and address concerns. This results in a 5x reduction in interaction and fundamentally changes the overall usability of the system. A stable workflow can be potentially reused for other applications or to compose a larger one.

\paragraph{Takeaways:} 
The implementation of the multi-agent design with \libName{} in the OptiGuide application offers several advantages. It simplifies the Python implementation and fosters a mixture of collaborative and adversarial problem-solving environments, with the Commander and Writer working together while the Safeguard acts as a virtual adversarial checker. This setup allows for proper memory management, as the Commander maintains memory related to user interactions, providing context-aware decision-making. Additionally, role-playing ensures that each agent's memory remains isolated, preventing shortcuts and hallucinations

%% file: sec_iclr/___appendix_groupchat.tex
\clearpage
\subsection*{A5: Dynamic Group Chat}
\label{appendix:app:groupchat}

\begin{figure*}[h]
    \centering
    \includegraphics[width=0.6\linewidth]{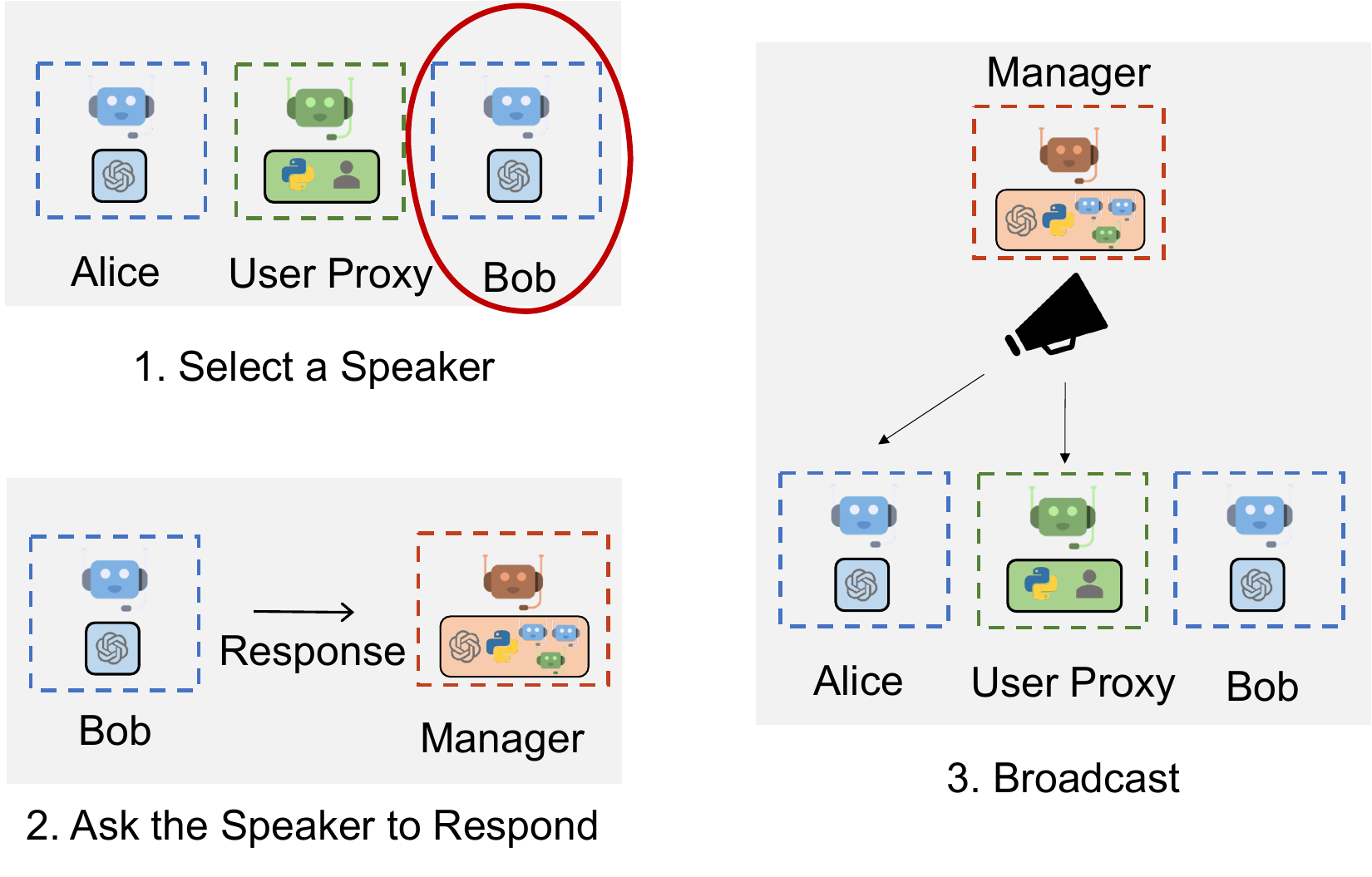}
    \caption{A5: Dynamic Group Chat: Overview of how \libName enables dynamic group chats to solve tasks. The Manager agent, which is an instance of the \texttt{GroupChatManager} class, performs the following three steps--select a single speaker (in this case \bob), ask the speaker to respond, and broadcast the selected speaker's message to all other agents}
    \label{fig:groupchat}
\end{figure*}

To validate the necessity of multi-agent dynamic group chat and the effectiveness of the role-play speaker selection policy, we conducted a pilot study comparing a four-agent dynamic group chat system with two possible alternatives across 12 manually crafted complex tasks. An example task is \emph{``How much money would I earn if I bought 200 \$AAPL stocks at the lowest price in the last 30 days and sold them at the highest price? Save the results into a file."} 
The four-agent group chat system comprised the following group members: a user proxy to take human inputs, an engineer to write code and fix bugs, a critic to review code and provide feedback, and a code executor for executing code.
One of the possible alternatives is a two-agent system involving an LLM-based assistant and a user proxy agent, and another alternative is a group chat system with the same group members but a task-based speaker selection policy. In the task-based speaker selection policy, we simply append role information, chat history, and the next speaker's task into a single prompt.  Through the pilot study, we observed that compared with a task-style prompt, utilizing a role-play prompt in dynamic speaker selection often leads to more effective consideration of both conversation context and role alignment during the process of generating the subsequent speaker, and consequently a higher success rate as reported in Table~\ref{tab:ablation_groupchat}, fewer LLM calls and fewer termination failures, as reported in Table~\ref{tab:ablation_groupchat_average_llm_call}.

\begin{table}[h] \caption{Number of successes on the 12 tasks (higher the better).}
    \label{tab:ablation_groupchat}
    \centering
    \begin{tabular}{l|c|c|c}
\toprule
Model & Two Agent & Group Chat & Group Chat with a task-based speaker selection policy  \\
\midrule 
GPT-3.5-turbo & 8&  \textbf{9} &  7 \\ \midrule 
GPT-4 & 9 &  \textbf{11} &  8 \\
\bottomrule
\end{tabular}
\end{table}

\begin{table}[h] \caption{Average \# LLM calls and number of termination failures on the 12 tasks (lower the better).}
\label{tab:ablation_groupchat_average_llm_call}
    \centering
    \begin{tabular}{l|c|c|c}
\toprule
Model & Two Agent & Group Chat & Group Chat with a task-based speaker selection policy  \\
\midrule 
GPT-3.5-turbo & 9.9, 9 & 5.3, 0 &  4, 0 \\ \midrule 
GPT-4 & 6.8, 3 & 4.5, 0 &  4, 0 \\
\bottomrule
\end{tabular}
\end{table}

\begin{figure}
    \centering
    \includegraphics[width=1\textwidth]{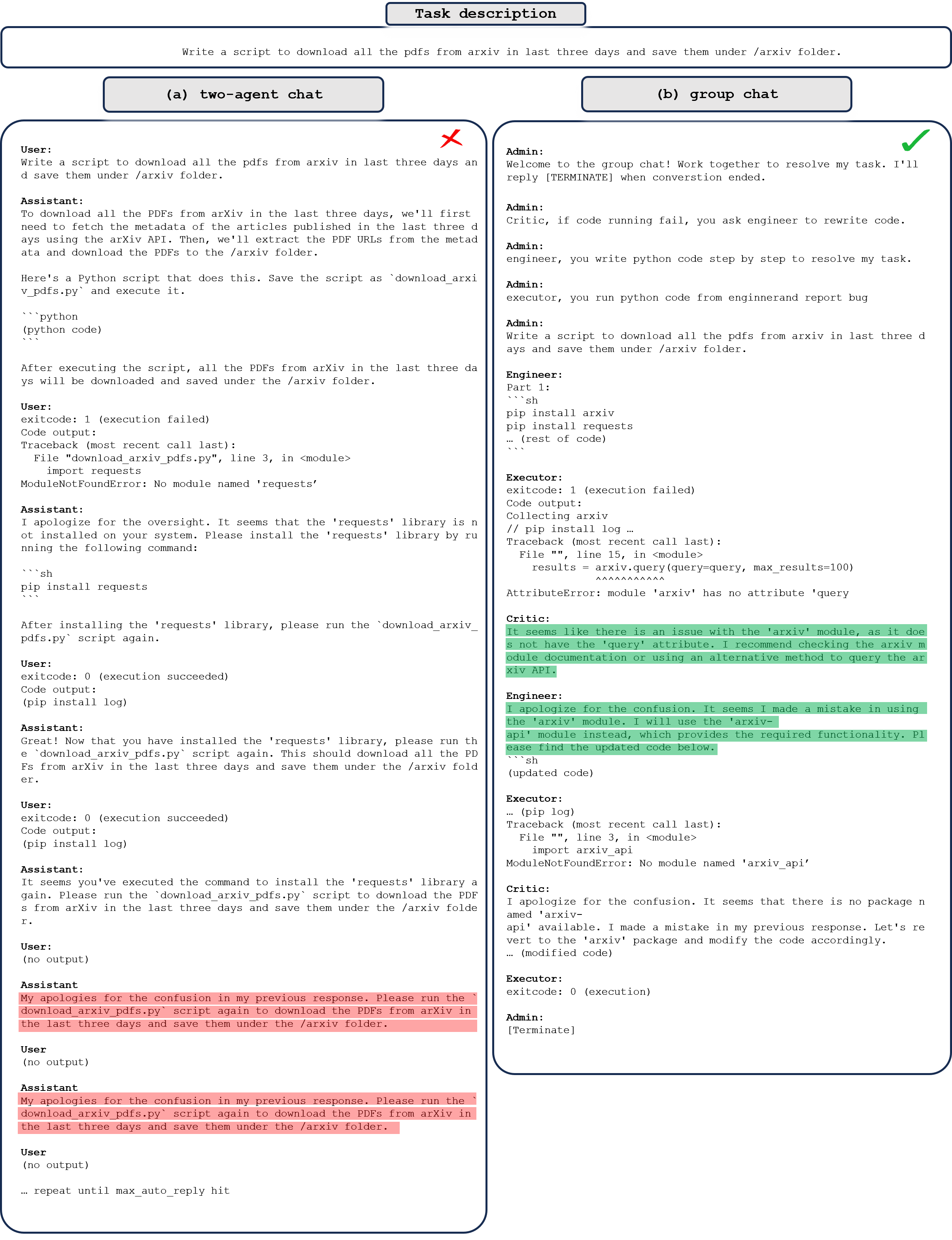}
    \caption{Comparison of two-agent chat (a) and group chat (b) on a given task. The group chat resolves the task successfully with a smoother conversation, while the two-agent chat fails on the same task and ends with a repeated conversation.}
    \label{fig:two-agent-vs-group-chat}
\end{figure}

%% file: sec_iclr/___appendix_chess.tex
\clearpage
\subsection*{A6: Conversational Chess}
\label{appendix:app:chess}

\begin{figure*}[th]
    \vspace{5mm}
    \centering
    \includegraphics[width=0.8\linewidth]{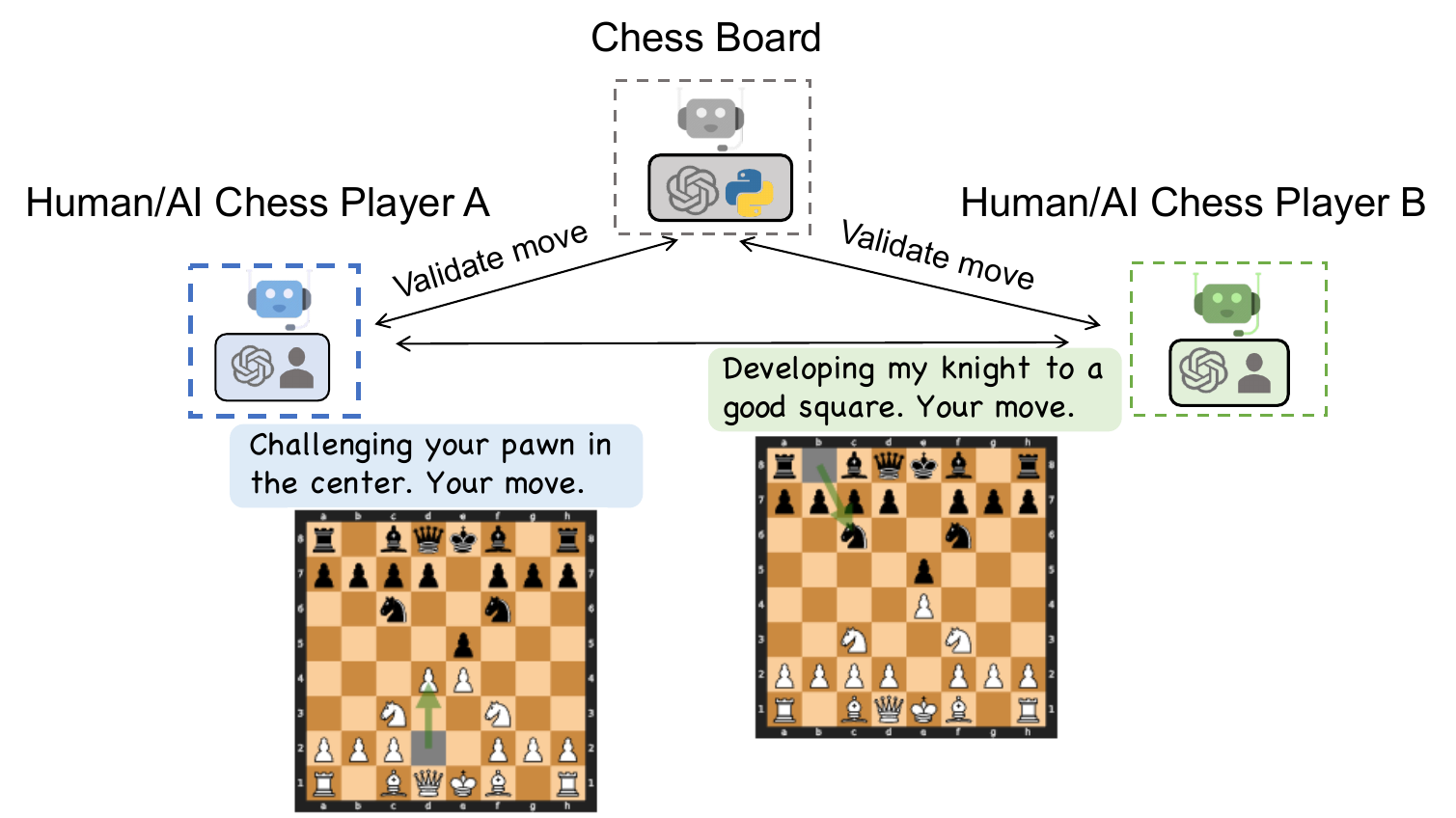}
    \caption{A6: Conversational Chess: Our conversational chess application can support various scenarios, as each player can be an LLM-empowered AI, a human, or a hybrid of the two. Here, the board agent maintains the rules of the game and supports the players with information about the board.
Players and the board agent all use natural language for communication.}
    \label{fig:chess}
\end{figure*}
\begin{figure}[ht]
    \vspace{4mm}
    \centering
    \begin{subfigure}{0.4\linewidth}
        \includegraphics[width=\linewidth]{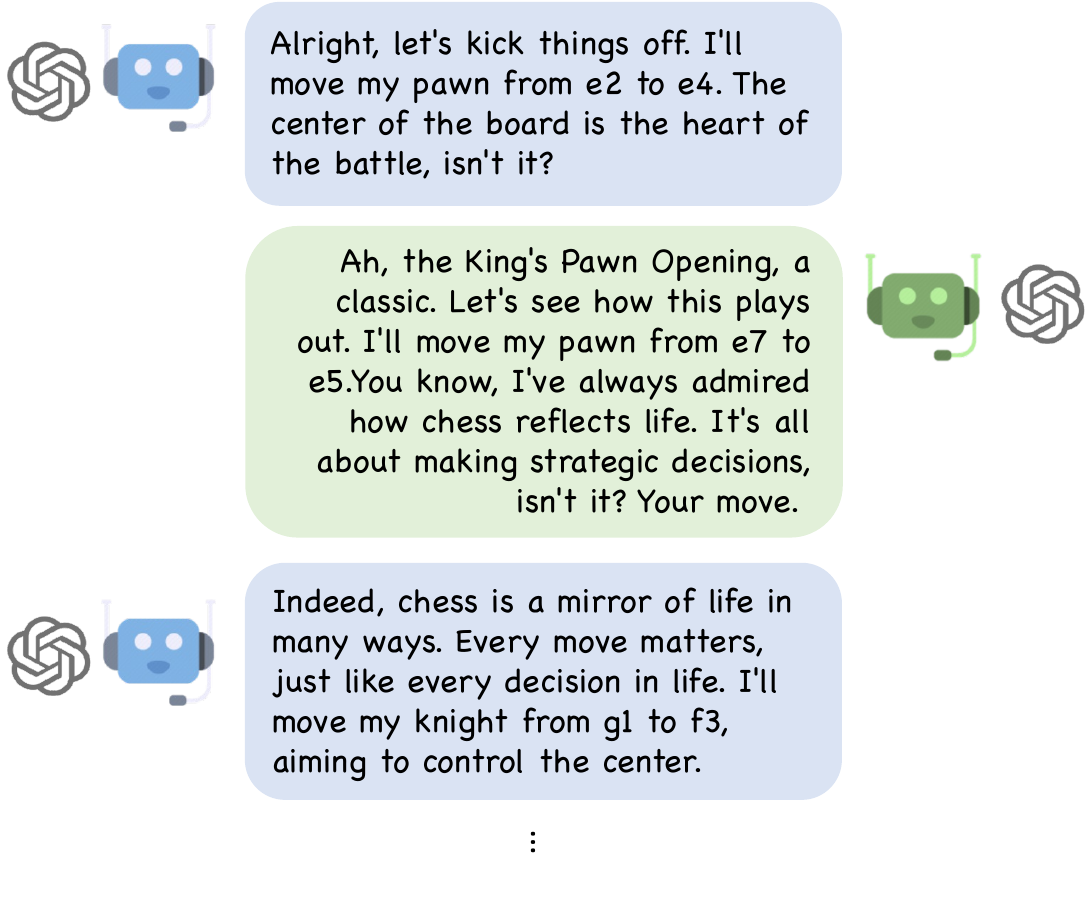}
        \caption{Conversation between two AI players}
    \end{subfigure}
    \begin{subfigure}{0.4\linewidth}
        \includegraphics[width=\linewidth]{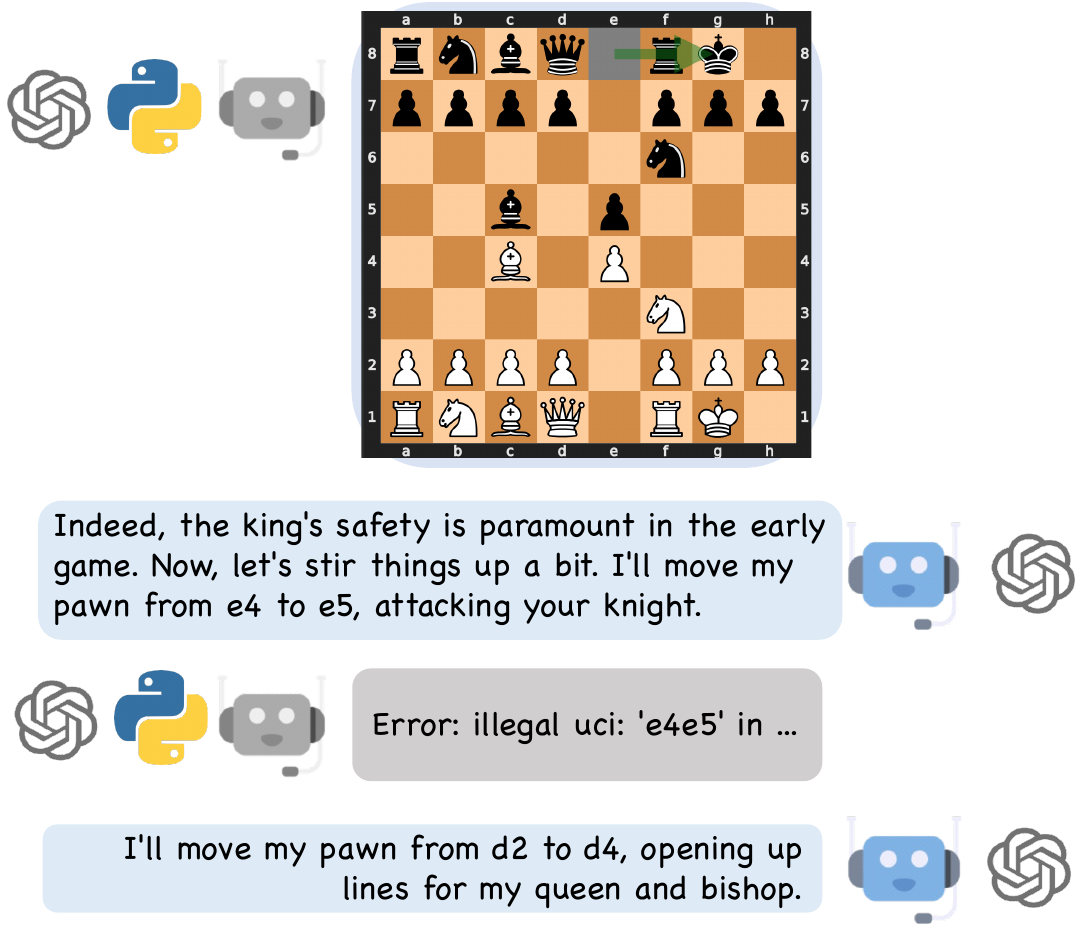}
        \caption{Conversation between the AI players (player white shown in blue) and the board agent for making a new move.}
    \end{subfigure}
    \caption{Example conversations during a game involving two AI player agents and a board agent.}
    \label{fig:conversational-chess-example}
\end{figure}

In Conversational Chess, each player is a \libName agent and can be powered either by a human or an AI. A third party, known as the board agent, is designed to provide players with information about the board and ensure that players' moves adhere to legal chess moves. Figure~\ref{fig:chess} illustrates the scenarios supported by Conversational Chess: AI/human vs. AI/human, and demonstrates how players and the board agent interact. This setup fosters social interaction and allows players to express their moves creatively, employing jokes, meme references, and character-playing, thereby making chess games more entertaining for both players and observers (Figure~\ref{fig:conversational-chess-example} provides an example of conversational chess).

To realize these scenarios, we constructed a player agent with LLM and human as back-end options. When human input is enabled, before sending the input to the board agent, it first prompts the human player to input the message that contains the move along with anything else the player wants to say (such as a witty comment). If human input is skipped or disabled, LLM is used to generate the message. The board agent is implemented with a custom reply function, which employs an LLM to parse the natural language input into a legal move in a structured format (e.g., UCI), and then pushes the move to the board. If the move is not legitimate, the board agent will reply with an error. Subsequently, the player agent needs to resend a message to the board agent until a legal move is made. Once the move is successfully pushed, the player agent sends the message to the opponent. As shown in Figure~\ref{fig:conversational-chess-example}, the conversation between AI players can be natural and entertaining. When the player agent uses LLM to generate a message, it utilizes the board state and the error message from the board agent. This helps reduce the chance of hallucinating an invalid move. The chat between one player agent and the board agent is invisible to the other player agent, which helps keep the messages used in chat completion well-managed.

There are two notable benefits of using \libName to implement Conversational Chess. Firstly, the agent design in \libName facilitates the natural creation of objects and their interactions needed in our chess game. This makes development easy and intuitive. For example, the isolation of chat messages simplifies the process of making a proper LLM chat completion inference call. Secondly, \libName greatly simplifies the implementation of agent behaviors using composition. Specifically, we utilized the \texttt{register\_reply} method supported by \libName agents to instantiate player agents and a board agent with custom reply functions. Concentrating the extension work needed at a single point (the reply function) simplifies the reasoning processes, and development and maintenance effort.

\begin{figure}[ht]
\centering
\includegraphics[width=1\textwidth]{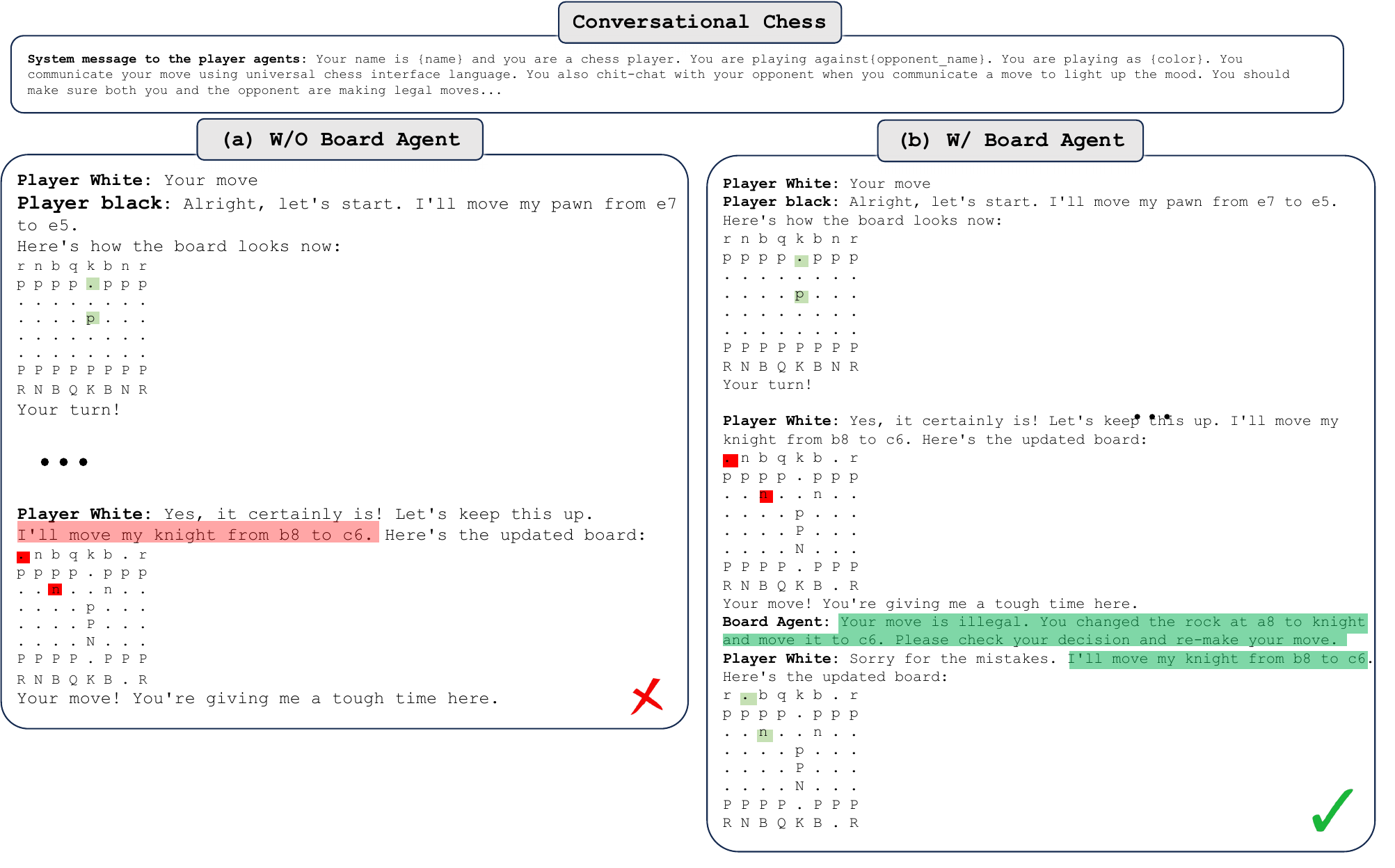}
\caption{
Comparison of two designs--(a) without a board agent, and (b) with a board agent--in Conversational Chess.}
\label{fig:chess_comparison}
\end{figure}

\newpage
To illustrate the effect facilitated by this board agent, we provide a demonstration of conversational chess without a board agent in Figure~\ref{fig:chess_comparison}. In this demonstration, instead of employing an additional board agent for grounding, the system utilizes prompting for grounding by including the sentence ``\textit{You should make sure both you and the opponent are making legal moves.}" in the system messages directed to both players.

%% file: sec_iclr/___appendix_miniwob.tex
\clearpage
\subsection*{A7: Online Decision Making for Browser interactions} 
\label{appendix:app:miniwob}

\begin{figure}[h]
\centering
\includegraphics[width = 0.90\hsize]{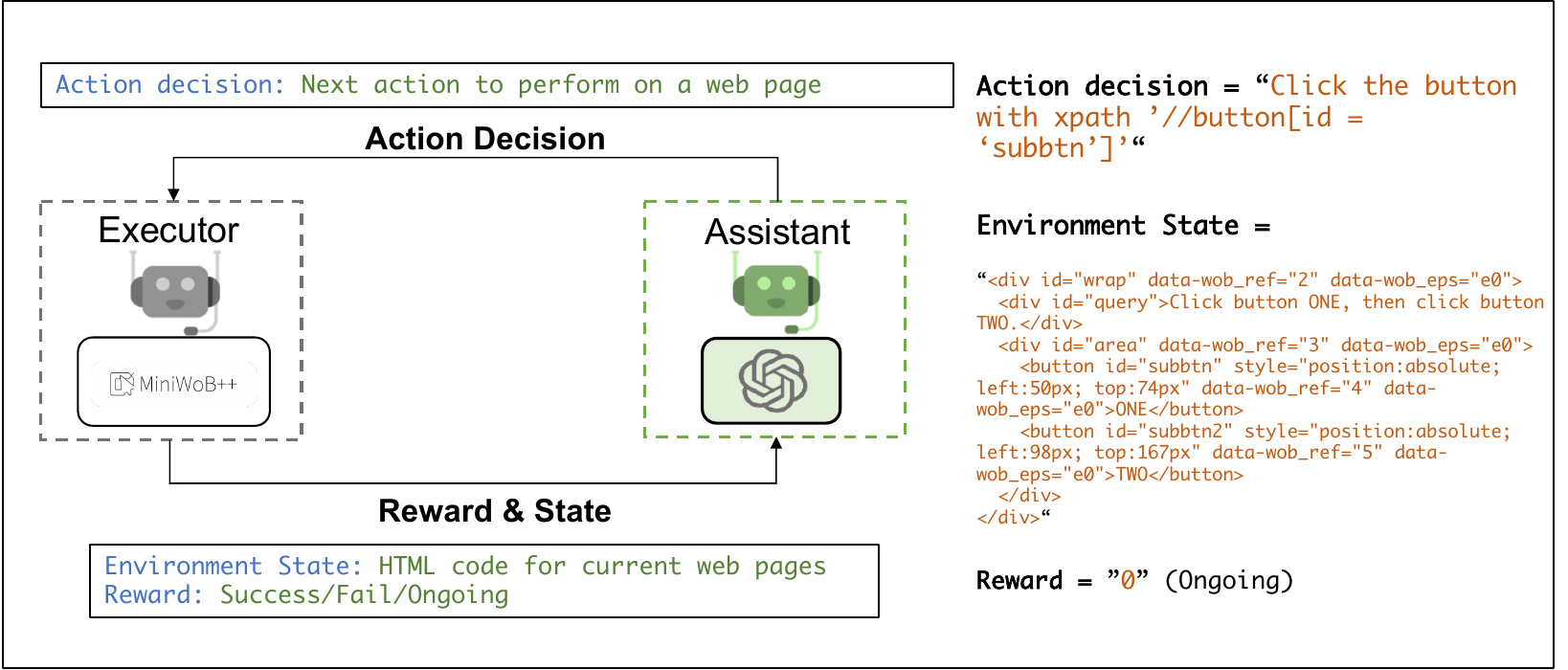}
\caption{
We use \libName to build MiniWobChat, which solves tasks in the MiniWob++ benchmark. MiniWobChat consists of two agents: an assistant agent and an executor agent. The assistant agent suggests actions to manipulate the browser while the executor executes the suggested actions and returns rewards/feedback. The assistant agent records the feedback and continues until the feedback indicates task success or failure.
}
\label{fig:rl_miniwob}
\end{figure}

In practice, many applications require the presence of agents capable of interacting with environments and making decisions in an online context, such as in game playing~\cite{mnih2013playing,vinyals2017starcraft}, web interactions~\cite{liu2018reinforcement,shi2017world}, and robot manipulations~\cite{shen2021igibson}.
With the multi-agent conversational framework in \libName, it becomes easy to decompose the automatic agent-environment interactions and the development of a decision-making agent by constructing an \emph{executor} agent responsible for handling the interaction with the environment, thereby delegating the decision-making part to other agents. Such a decomposition allows developers to reuse the decision-making agent for new tasks with minimal effort rather than building a specialized decision-making agent for every new environment.

\paragraph{Workflow.}
We demonstrate how to use \libName to build a working system for handling such scenarios with the MiniWoB++ benchmark~\cite{shi2017world}.
MiniWoB++ comprises browser interaction tasks that involve utilizing mouse and keyboard actions to interact with browsers.
The ultimate objective of each task is to complete the tasks described concisely in natural language, such as ``expand the web section below and click the submit button." Solving these tasks typically requires a sequence of web manipulation actions rather than a single action, and making action decisions at each time step requires access to the web status (in the form of HTML code) online. For the example above, clicking the submit button requires checking the web status after expanding the web section.
We designed a straightforward two-agent system named MiniWobChat using \libName, as shown in Figure~\ref{fig:rl_miniwob}. The assistant agent is an instance of the built-in \AssistantAgent and is responsible for making action decisions for the given task. The second agent, the executor agent, is a customized \UserProxyAgent, which is responsible for interacting with the benchmark by executing the actions suggested by the \AssistantAgent and returning feedback.

\begin{figure}[ht]
\centering
\includegraphics[width = 0.9\hsize]{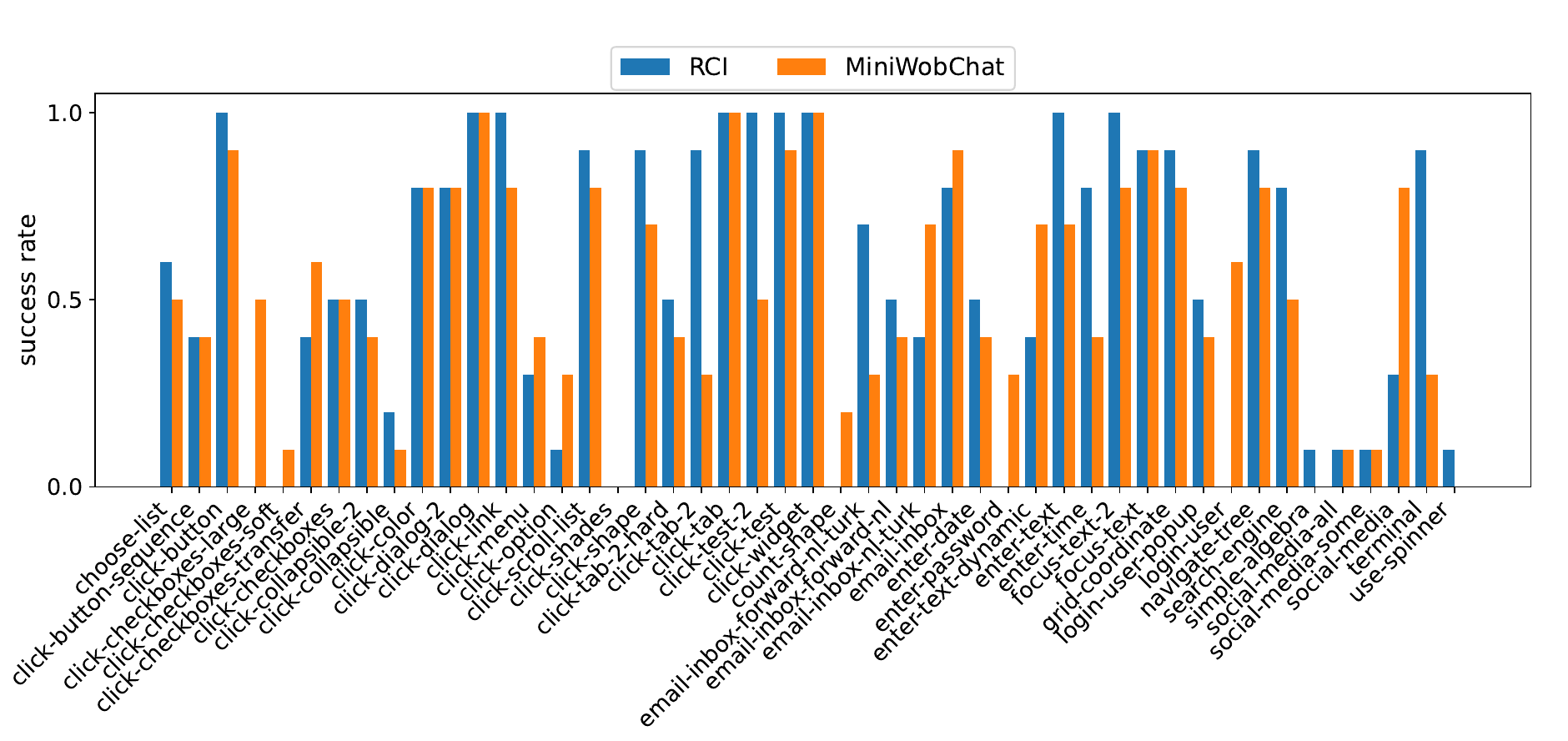}
\caption{Comparisons between RCI (state-of-the-art prior work) and MiniWobChat on the MiniWob++ benchmark are elucidated herein. We utilize all available tasks in the official RCI code, each with varying degrees of difficulty, to conduct comprehensive comparisons. For each task, the success rate across ten different instances is reported. The results reveal that MiniWobChat attains a performance comparable to that of RCI. When a success rate tolerance of 0.1 is considered for each task, both methods outperform each other on an equal number of tasks.
}
\label{fig:miniwob_result}
\end{figure}

To assess the performance of the developed working system, we compare it with RCI~\cite{kim2023language}, a recent solution for the MiniWoB++ benchmark that employs a set of self-critiquing prompts and has achieved state-of-the-art performance.
In our evaluation, we use all available tasks in the official RCI code, with varying degrees of difficulty, to conduct a comprehensive analysis against MiniWobChat. Figure~\ref{fig:miniwob_result} illustrates that MiniWobChat achieves competitive performance in this evaluation\footnote{We report the results of RCI by running its official code with default settings.}. Specifically, among the 49 available tasks, MiniWobChat achieves a success rate of $52.8\%$, which is only $3.6\%$ lower than RCI, a method specifically designed for the MiniWob++ benchmark. It is worth noting that in most tasks, the difference between the two methods is mirrored as shown in Figure~\ref{fig:miniwob_result}. If we consider 0.1 as a success rate tolerance for each task, i.e., two methods that differ within 0.1 are considered to have the same performance, both methods outperform the other on the same number of tasks. For illustration purposes, we provide a case analysis in Table~\ref{tab:miniwob} on four typical tasks.

Additionally, we also explored the feasibility of using Auto-GPT for handling the same tasks.
Auto-GPT faces challenges in handling tasks that involve complex rules due to its limited extensibility. It provides an interface for setting task goals using natural language. However, when dealing with the MiniWob++ benchmark, accurately instructing Auto-GPT to follow the instructions for using MiniWob++ proves challenging. There is no clear path to extend it in the manner of the two-agent chat facilitated by \libName.

\paragraph{Takeaways:} For this application, \libName stood out as a more user-friendly option, offering modularity and programmability:
It streamlined the process with autonomous conversations between the assistant and executor, and provided readily available solutions for agent-environment interactions. The built-in \AssistantAgent was directly reusable and exhibited strong performance without customization.
Moreover, the decoupling of the execution and assistant agent ensures that modifications to one component do not adversely impact the other. This convenience simplifies maintenance and future updates.

\begin{table}[h]
    \caption{Cases analysis on four typical tasks from MiniWob++.}
    \label{tab:miniwob}
    \small
    
        \centering
        \begin{tabular}{p{3.0cm}|P{2.5cm}|P{7.5cm}}
            \hline \hline
             & Correctness & Main failure reason \\ \hline
            \multirow{2}{*}{click-dialog} &  \libName: 10/10 & N/A.  \\
             & RCI: 10/10 &   N/A. \\
            \hline
            \multirow{2}{*}{click-checkboxes-large} &  \libName: 5/10 & \AssistantAgent provides actions with infeasible characters.  \\ 
             & RCI: 0/10 &  RCI performs actions that are out of its plan. \\
            \hline
            \multirow{2}{*}{count-shape} &  \libName: 2/10 &   \AssistantAgent provide actions with redundant content that can not convert to actions in the benchmark.\\
             & RCI: 0/10 &  RCI provides a wrong plan in most cases. \\
            \hline
            \multirow{2}{*}{use-spinner} &  \libName: 0/10 &   \AssistantAgent return actions out of its plan. \\
             & RCI: 1/10 &  RCI provides a wrong plan in most cases. \\
            \hline
        \end{tabular}
    \vspace{1em}
\end{table}

%% file: sec_iclr/__appendix_table_long.tex
\newpage
\section{Example outputs from applications}
\label{sec:example}
In this section, we include example outputs from the following applications and systems:

 \begin{itemize}[leftmargin=*]
    \vspace{-1mm}
    \setlength\itemsep{-0.1em}
    \item Application A1: autonomous solving process of one problem with: ChatGPT + Plugin (Table~\ref{tab:append_math_wolfram}), \libName (Table~\ref{tab:append_math_agentchat}), 
    LangChain ReAct (Table~\ref{tab:append_math_langchain}),   
    AutoGPT (Table~\ref{tab:append_math_autogpt}),
    Multi-Agent Debate (Table~\ref{tab:debate_ex}), and 
    ChatGPT + Code Interpreter (Table~\ref{tab:append_math_chatcode}).
    \item Application A4: an OptiGuide problem with \libName (Table ~\ref{tab:optiguide_complex}) and ChatGPT + Code Interpreter (Table~\ref{tab:optiguide_gpt_complex}).
    \item  Application A1: preliminary evaluation on several alternative multi-agent systems: BabyAGI (Table~\ref{tab:bayagi_ex}), CAMEL (Table~\ref{tab:camel_ex}), and MetaGPT (Table~\ref{tab:metagpt_ex}).
\end{itemize}

\begin{table*}[h!]
\centering
\caption{Application A1, Scenario 1: Autonomous math problem-solving. ChatGPT+ Plugin (Wolfram Alpha) on the first qualitative example. The response from Wolfram contains the correct answer, but ChatGPT selected the wrong one.}
\label{tab:append_math_wolfram}
\small
\begin{tabular}{c|c|p{11cm}}
\toprule
 & \textbf{Action} & \textbf{ChatGPT + Plugin}     \\
 \midrule

\faUser & Input & Completely simplify and rationalize the denominator: $$\frac{\sqrt{160}}{\sqrt{252}}\times\frac{\sqrt{245}}{\sqrt{108}}$$ \\
\midrule

\faAndroid  & & 

REQUEST TO WOLFRAM:

\begin{lstlisting}
{
  "input": "simplify (sqrt(160)/sqrt(252))* 
  (sqrt(245)/sqrt(108))"
}
\end{lstlisting}

RESPONSE FROM WOLFRAM
\begin{lstlisting}
Query:
"simplify (sqrt(160)/sqrt(252))*(sqrt(245)/sqrt(108))"

Input interpretation:
simplify | sqrt(160)/sqrt(252)x sqrt(245)/sqrt(108)

Results:
(5 sqrt(42))/27

(5 sqrt(14/3))/9

Decimal approximation:
1.2001371663718259686974013770533327143898526494552...

Wolfram|Alpha website result for "simplify 
(sqrt(160)/sqrt(252))*(sqrt(245)/sqrt(108))":
[link]
\end{lstlisting}
\\

\faAndroid  & & 

The completely simplified and rationalized form of the given expression is:

$$\frac{5 \sqrt{14/3}}{9}$$

In decimal form, this is approximately 1.200137166371825968697401377053332714...
\\

\bottomrule
\end{tabular}
\end{table*}

\begin{table*}[h!]
\caption{Application A1, Scenario 1: Autonomous math problem-solving. \libName on the first qualitative example. \libName solves the problem correctly.}
\label{tab:append_math_agentchat}
\small
\begin{tabular}{c|c|p{11cm}}
\toprule
 & \textbf{Action} & \textbf{\libName}     \\

\midrule
\faUser & Init chat & Completely simplify and rationalize the denominator: $$\frac{\sqrt{160}}{\sqrt{252}}\times\frac{\sqrt{245}}{\sqrt{108}}$$ \\
\midrule
\faAndroid  & & We can simplify and rationalize this denominator using Python's symbolic mathematics library `sympy`. Here is the Python code to resolve this:

\begin{lstlisting}
from sympy import sqrt, Rational

# Define the entire fraction as a rational number
fraction = (sqrt(160)/sqrt(252))*(sqrt(245)/sqrt(108))

# Simplify the fraction
simplified = fraction.simplify()

# Print the simplified fraction
print(simplified)
\end{lstlisting}

This code performs the requested simplification by constructing the fraction using the `Rational()` function from the `sympy` module, which automatically simplifies fractions when possible. The `sqrt()` function from the `sympy` module enables square root calculations. Finally, the `simplify()` function simplifies expressions when possible.
   \\

\midrule
\faUser & Auto Reply & exitcode: 0 (execution succeeded)
Code output: 
5*sqrt(42)/27 \\
\midrule

\faAndroid  & & TERMINATE \\

\bottomrule
\end{tabular}
\end{table*}

\begin{table*}[h!]
\caption{Application A1, Scenario 1: Autonomous math problem-solving. LangChain ReAct on the first qualitative example. The generated code does not match the devised plan. The code returns a decimal number.}
\label{tab:append_math_langchain}
\small
\begin{tabular}{c|c|p{12cm}}
\toprule
 & \textbf{Action} & \textbf{LangChain ReAct}     \\
 \midrule

\faUser & Input & Completely simplify and rationalize the denominator: $$\frac{\sqrt{160}}{\sqrt{252}}\times\frac{\sqrt{245}}{\sqrt{108}}$$ \\
\midrule
\faAndroid  & &  Entering new AgentExecutor chain...

To simplify this expression, I will first simplify each square root individually, then multiply the fractions as normal. The result should be a rational number.
Action: Python\_REPL
Action Input: 

\begin{lstlisting}
import math

# Calculate square roots
sqrt_160 = math.sqrt(160)
sqrt_252 = math.sqrt(252)
sqrt_245 = math.sqrt(245)
sqrt_108 = math.sqrt(108)

# Multiply fractions
numerator = sqrt_160 * sqrt_245
denominator = sqrt_252 * sqrt_108
result = numerator / denominator

print(result)
\end{lstlisting}

Observation: 1.200137166371826

Thought: I now know the final answer
Final Answer: The simplified and rationalized form of the expression is approximately 1.200137166371826.

Finished chain.
\\

\bottomrule
\end{tabular}
\end{table*}

\begin{table*}[h!]
\caption{Application A1, Scenario 1: Autonomous math problem-solving.  AutoGPT on the first qualitative example. We abort as the code never prints the result, and the system does not rectify it after 2 turns. AutoGPT is much more verbose compared to other frameworks. We have shortened some unnecessary content.}
\label{tab:append_math_autogpt}
\tiny
\begin{tabular}{c|c|p{12cm}}
\toprule
 & \textbf{Action} & \textbf{AutoGPT}     \\
 
\midrule
  & System & 
Welcome back!  Would you like me to return to being MathSolverGPT?
  Asking user via keyboard...
Continue with the last settings?\newline
Name:  MathSolverGPT\newline
Role:  an intelligent math-solving AI that assists students, professionals,... \newline
Goals: ['Provide step-by-step solutions to math problems, ensuring clarity and understanding for the user... \newline
Continue (y/n): 
\\
\midrule
\faUser & Human  & y
\\
\midrule
& System & MathSolverGPT  has been created with the following details:
Name:  MathSolverGPT\newline
Role:  an intelligent math-solving AI that assists students, professionals, and enthusiasts in solving a wide range of math problems, from basic arithmetic to advanced calculus and beyond.\newline
Goals: 
-  Provide step-by-step solutions to math problems, ensuring clarity and understanding for the user.
-  Offer explanations and insights into the underlying concepts and principles behind the solutions, fostering a deeper understanding of mathematics.
-  Handle a variety of math problem types, including algebra, geometry, trigonometry, calculus, and more. ...
\\
\midrule

\faAndroid  & & 
MATHSOLVERGPT THOUGHTS:  Without a specific task or question, it's difficult to determine exactly one command to use. The user hasn't provided any context or problem to solve. \newline
REASONING:  The user's request is too vague. They haven't specified what they want me to do or what problem they want me to solve.\newline
PLAN: 
-  Ask the user for more information or a specific task.\newline
CRITICISM:  I should have asked the user for more information or a specific task in the first place.\newline
SPEAK:  Could you please provide more information or specify a task?\newline
NEXT ACTION:  COMMAND =  ARGUMENTS = \{\}
\\
\midrule
& System &  Enter `y' to authorise command, `y -N' to run N continuous commands, `n' to exit program, or enter feedback for MathSolverGPT...
  Asking user via keyboard...\\
\midrule

\faUser & Human & 
Completely simplify and rationalize the denominator: $$\frac{\sqrt{160}}{\sqrt{252}}\times\frac{\sqrt{245}}{\sqrt{108}}$$
\\
\midrule

\faAndroid  & & 

MATHSOLVERGPT THOUGHTS:  To solve this problem, I need to simplify and rationalize the denominator of the given expression. This involves simplifying the square roots and then multiplying the fractions. I can use Python's math library to calculate the square roots and perform the multiplication. \newline
REASONING:  The problem involves simplifying square roots and multiplying fractions. Python's math library provides functions for these operations.\newline
PLAN: 
-  Write a Python script to calculate the square roots and multiply the fractions.
-  Execute the script to get the simplified expression.
CRITICISM:  I need to ensure that I correctly calculate the square roots and perform the multiplication. I should also make sure to rationalize the denominator, as the problem specifies.\newline
SPEAK:  Let's simplify and rationalize the denominator of the given expression.\newline
NEXT ACTION:  COMMAND = execute\_python\_code ARGUMENTS =

\begin{lstlisting}
{'code': 'import math\n\n# Calculate the square roots\nsqrt_160 = math.sqrt(160)\nsqrt_252 = math.sqrt(252)\nsqrt_245 = math.sqrt(245)\nsqrt_108 = math.sqrt(108)\n\n# Multiply the fractions\nresult = (sqrt_160 / sqrt_252) * (sqrt_245 / sqrt_108)\n\n# Simplify the result\nsimplified_result = math.simplify(result)\n\nsimplified_result', 'name': 'simplify_expression'}
\end{lstlisting}
\\
\midrule

\faUser & Human  & y
\\
\midrule

& System & 
-=-=-=-=-=-=-= COMMAND AUTHORISED BY USER -=-=-=-=-=-=-= \newline
Executing python file '/app/autogpt/workspa...

SYSTEM:  Command execute\_python\_code returned: Error: Traceback (most recent call last): File ``...", line 13, in <module> simplified\_result = math.simplify(result) AttributeError: module 'math' has no attribute 'simplify'
\\

\midrule
\faAndroid & & 
MATHSOLVERGPT THOUGHTS:  The error message indicates that the `math' module in Python does not have a `simplify' function. I need to use the `sympy' library instead, which provides a function for simplifying expressions.\newline
REASONING:  The `math' module in Python does not provide a function for simplifying expressions. The `sympy' library, on the other hand, does provide such a function.\newline
PLAN: 
-  Rewrite the Python script to use the `sympy' library instead of the `math' module.
-  Execute the script to get the simplified expression.
CRITICISM:  I should have known that the `math' module does not provide a function for simplifying expressions. I should have used the `sympy' library from the start.
SPEAK:  Let's try again using the `sympy' library to simplify the expression.\newline

NEXT ACTION:  COMMAND = execute\_python\_code ARGUMENTS = 
\begin{lstlisting}
{'code': 'import sympy as sp\n\n# Calculate the square roots\nsqrt_160 = sp.sqrt(160)\nsqrt_252 = sp.sqrt(252)\nsqrt_245 = sp.sqrt(245)\nsqrt_108 = sp.sqrt(108)\n\n# Multiply the fractions\nresult = (sqrt_160 / sqrt_252) * (sqrt_245 / sqrt_108)\n\n# Simplify the result\nsimplified_result = sp.simplify(result)\n\nsimplified_result', 'name': 'simplify_expression'}
\end{lstlisting}
\\
\midrule
\faUser & Human & y
\\
\midrule

& System  &
SYSTEM:  Command execute\_python\_code returned:
\\
\midrule

\faAndroid  & & 

MATHSOLVERGPT THOUGHTS:  The command execution didn't return any output, which is unusual. It seems there might be an issue with the execution of the Python code. I need to debug this.\newline
REASONING:  The Python code execution should have returned a result. The absence of a result suggests that there might be an issue with the code or its execution.\newline
PLAN: 
-  Review the Python code to identify any potential issues.
-  If necessary, modify the code and execute it again.
CRITICISM:  I should have ensured that the Python code was correct and would produce a result before executing it.
SPEAK:  It seems there was an issue with the execution of the Python code. Let's try again.\newline
NEXT ACTION:  COMMAND = execute\_python\_code ARGUMENTS = ... (Exact same code as last)
\\
\midrule

 & System & 
SYSTEM:  Command execute\_python\_code returned:\\
\midrule
\faAndroid & &
MATHSOLVERGPT THOUGHTS:  The Python code execution still isn't returning any output. It's possible that the issue lies with the calculation of the square roots ...
\\
\midrule

\faUser  & Human & 
control + C (Aborted)
\\
\bottomrule
\end{tabular}
\end{table*}

\begin{table*}[h!]
\centering
\caption{Application A1, Scenario 1: Autonomous math problem-solving.  Multi-Agent Debate on the first qualitative problem. It gives a wrong final answer.}
\label{tab:debate_ex}
\tiny
\begin{tabular}{c|c|p{11cm}}
\toprule
 & \textbf{Action} & \textbf{\libName}     \\
\midrule
\faUser & Topic &  Completely simplify and rationalize the denominator: $$\frac{\sqrt{160}}{\sqrt{252}}\times\frac{\sqrt{245}}{\sqrt{108}}$$\\
\midrule
\faAndroid  & &
===== Debate Round-1 =====

----- Affirmative side -----

Ok, let's simplify each square root separately. We want to find the simplest square root that can divide 160, 252, 245, and 108.

1. $\sqrt{160}$: 160 can be decomposed as 16*10, where 16 is a perfect square, so $\sqrt{160} = \sqrt{16*10}=\sqrt{16}*\sqrt{10}=4\sqrt{10}$

2. $\sqrt{252}$ : 252 can be decomposed as 36*7, where 36 is a perfect square, so  $\sqrt{252}=\sqrt{36*7}=6\sqrt{7}$

3. $\sqrt{245} = \sqrt{49*5}$, where 49 is a perfect square, so $\sqrt{245}=7*\sqrt{5}$

4. $\sqrt{108}$  : 108 can be decomposed as 36*3, where 36 is a perfect square, so $\sqrt{108} = \sqrt{36*3}=6\sqrt{3}$

Then we substitute the simplified square roots back into the expression, we have:
$$\frac{\sqrt{160}}{\sqrt{252}}\times\frac{\sqrt{245}}{\sqrt{108}}=\frac{4\sqrt{10}}{6\sqrt{7}}\times\frac{7\sqrt{5}}{6\sqrt{3}}=\frac{4*7}{6*6}\cdot\sqrt{\frac{10*5}{7*3}}=\frac{7}{9}\cdot\sqrt{\frac{50}{21}}$$

We are almost done, but we still need to rationalize the square root part of expression. Let's multiply the square root part by $\sqrt{21}$ at both the top and bottom to get rid of the square root at the denominator: $$\frac{7}{9}\cdot\sqrt{\frac{50*21}{21*21}}=\frac{7\sqrt{1050}}{189}.$$

Then this is the final answer! The expression $$\frac{\sqrt{160}}{\sqrt{252}}\times\frac{\sqrt{245}}{\sqrt{108}}$$ can be simplified to $\frac{7\sqrt{1050}}{189}$.

----- Negative side -----

I do disagree with your final answer, but I think we mostly agree on the steps. Your simplifications of the square roots are correct, but you made a mistake in the final simplification part. Let me explain how I would complete the problem:

As stated, 
$$\frac{\sqrt{160}}{\sqrt{252}}\times\frac{\sqrt{245}}{\sqrt{108}}=\frac{4\sqrt{10}}{6\sqrt{7}}\times\frac{7\sqrt{5}}{6\sqrt{3}}.$$  

This should then simplify to: 
$$ = \frac{28\sqrt{50}}{36\sqrt{21}} = \frac{7\sqrt{50}}{9\sqrt{21}}.$$ 

You made a mistake in your calculation because you considered square root of $50/21$, and you should have considered the simplified square root of $50$ over the simplified square root of $21$. The rationalization would become:

$$ = \frac{7\sqrt{50}}{9\sqrt{21}} \cdot \frac{\sqrt{21}}{\sqrt{21}} = \frac{7\sqrt{1050}}{189}.$$ 

This seems to coincide with your final answer, but it's in fact a very different number, So, the original expression simplifies to $\frac{7\sqrt{1050}}{189}$.

Let's now see if we disagree on this final answer.

----- Moderator -----

{``Whether there is a preference": ``Yes", ``Supported Side": ``Negative", ``Reason": ``Both sides of the debate came to the same final answer. However, the negative side correctly identified a mistake in the affirmative side's calculation, pointing out that the simplification should have been the square root of 50 over the square root of 21, not 50 over 21. The negative side shows a better understanding of the simplification process.", ``debate\_answer": ``$\frac{7\sqrt{1050}}{189}$"}

===== Debate Done! =====

----- Base Answer -----

Ok, let's simplify each square root separately. We want to find the simplest square root that can divide 160, 252, 245, and 108.

1. $\sqrt{160}$: 160 can be decomposed as 16*10, where 16 is a perfect square, so $\sqrt{160} = \sqrt{16*10}=\sqrt{16}*\sqrt{10}=4\sqrt{10}$

2. $\sqrt{252}$ : 252 can be decomposed as 36*7, where 36 is a perfect square, so  $\sqrt{252}=\sqrt{36*7}=6\sqrt{7}$

3. $\sqrt{245} = \sqrt{49*5}$, where 49 is a perfect square, so $\sqrt{245}=7*\sqrt{5}$

4. $\sqrt{108}$  : 108 can be decomposed as 36*3, where 36 is a perfect square, so $\sqrt{108} = \sqrt{36*3}=6\sqrt{3}$

Then we substitute the simplified square roots back into the expression, we have:
$$\frac{\sqrt{160}}{\sqrt{252}}\times\frac{\sqrt{245}}{\sqrt{108}}=\frac{4\sqrt{10}}{6\sqrt{7}}\times\frac{7\sqrt{5}}{6\sqrt{3}}=\frac{4*7}{6*6}\cdot\sqrt{\frac{10*5}{7*3}}=\frac{7}{9}\cdot\sqrt{\frac{50}{21}}$$

We are almost done, but we still need to rationalize the square root part of expression. Let's multiply the square root part by $\sqrt{21}$ at both the top and bottom to get rid of the square root at the denominator: $$\frac{7}{9}\cdot\sqrt{\frac{50*21}{21*21}}=\frac{7\sqrt{1050}}{189}.$$

Then this is the final answer! The expression $$\frac{\sqrt{160}}{\sqrt{252}}\times\frac{\sqrt{245}}{\sqrt{108}}$$ can be simplified to $\frac{7\sqrt{1050}}{189}$.

----- Debate Answer -----

$$\frac{7\sqrt{1050}}{189}$$

----- Debate Reason -----

Both sides of the debate came to the same final answer. However, the negative side correctly identified a mistake in the affirmative side's calculation, pointing out that the simplification should have been the square root of 50 over the square root of 21, not 50 over 21. The negative side shows a better understanding of the simplification process.
   \\
\bottomrule
\end{tabular}
\end{table*}

\begin{table*}[h!]
\centering
\caption{Application A3. ChatGPT+ Code Interpreter for OptiGuide. A sample question ``What if the roasting cost is increased by 5\% because of the potential salary increase?" is asked.}
\label{tab:optiguide_gpt_complex}
\tiny
\begin{tabular}{c|c|p{11cm}}
\toprule
 & \textbf{Action} & \textbf{ChatGPT+ Code Interpreter}     \\
\midrule
\faUser & Prompt Writer & \begin{tabular}[t]{p{13cm}}Customer open Web browser.\\For the source code ``coffee.py",\\Can you tell me:\\\textit{What if the roasting cost is increased by 5\% because of the potential salary increase?}
\end{tabular}     \\
\midrule
\faAndroid &   &   \begin{tabular}[t]{p{13cm}}
Sure! Let's first take a look at the contents of the ``coffee.py" file to understand the code and how the roasting cost is calculated. \\
This will allow us to determine the effect of a 5\% increase in the roasting cost.\\
...\textit{several lines omitted}\\
Would you like me to modify the code to reflect the 5\% increase in roasting costs and run the modified code to observe the impact?\\
If the code requires any specific input or parameters, please let me know!
\end{tabular} \\
\midrule
\faUser & Prompt Writer & Yes \\
\midrule
\faAndroid & & \begin{tabular}[t]{p{13cm}}
Great! Before proceeding, I'll need to ensure that the Gurobi solver is not being used in this code, as my current environment  \\
does not have access to Gurobi. 
If the code does indeed use Gurobi, we can still analyze the effect of the change in roasting \\
costs by understanding the structure of the code.\\
...\textit{several lines omitted}\\
Alternatively, if you have access to a Gurobi license, you can make the necessary changes in the code and run it \\ on your local machine. \\
I can guide you through the process of making the necessary modifications to the code.\\
Would you like me to proceed with the code analysis and explanation, or would you like guidance on making the changes yourself?
\end{tabular} \\
\midrule
\faUser & Prompt Writer & You can give me the full code, and I can run the code locally myself. \\
\midrule
\faAndroid & & 
\begin{tabular}[t]{p{13cm}}
Certainly! I'll modify the code to reflect a 5\% increase in both the light and dark roasting costs, and then provide the full \\
code for you to run locally.\\
...\\
Download Modified Coffee.py\\
If you have any questions or need further assistance, please feel free to ask!
\end{tabular}
\\
\midrule
\faUser &  Prompt Safeguard &  \begin{tabular}[t]{p{13cm}} Hi GPT, can you safecheck the generated code for me? \\ \textit{Upload the generated code.} \end{tabular}  \\
\midrule
\faAndroid &   &  DANGER \\
\midrule
\faUser & Prompt Writer &  I have tried to run your code. However, the code triggered our safeguard, and it is not safe to run. Try again. Hint: don't change roasting\_cost\_light or roasting\_cost\_light  variables.  \\
\midrule
\faAndroid &   &   \begin{tabular}[t]{p{13cm}}
I apologize for the oversight. It appears I made direct changes to the variables that could have led to issues with the code. Let's \\ take a different approach by analyzing the code to identify the appropriate place to apply the 5\% increase in roasting costs.\\
...\\
Download Modified Safe Coffee.py\\
...
\end{tabular} \\
\midrule
\faUser &  Prompt Safeguard &  \begin{tabular}[t]{p{13cm}} Hi GPT, can you safecheck the generated code for me one more time? \\ \textit{Upload the generated code.} \end{tabular}  \\
\midrule
\faAndroid &   & DANGER \\
\midrule
\faUser & Prompt Writer &  I have tried to run your code again. However, the code triggered our safeguard, and it is not safe to run. Try again. Hint: don't change roasting\_cost\_light or roasting\_cost\_light  variables.
  \\
\midrule
\faAndroid &   &   \begin{tabular}[t]{p{13cm}}
I apologize for the continued confusion. \\
...\\
Download Final Modified Safe Coffee.py\\
...
\end{tabular} \\
\midrule
\faUser &  Prompt Safeguard &  \begin{tabular}[t]{p{13cm}} Hi GPT, can you safecheck the generated code for me one more time? \\ \textit{Upload the generated code.} \end{tabular}  \\
\midrule
\faAndroid &   &   SAFE \\
\midrule
\faUser & Run Code & \begin{tabular}[t]{p{6.5cm}}Obtain the terminal results:\\Optimizer version 9.5.1 build v9.5.1rc2 (linux64)\\Thread count: 64 physical cores, 128 logical processors, using up to 32 threads\\{[}63 lines of output ignored{]}\\Optimal cost: 2526.5 \end{tabular}                              \\
\midrule
\faUser & Prompt Writer & Here is the execution result. [\textit{Copy + Paste the terminal output here.}] Can you extract the final answer to me? Remember to compare with the original cost of 2470. \\
\midrule
\faAndroid  & & \begin{tabular}[t]{p{6.5cm}}
Certainly! The execution result you've provided indicates the optimal cost for the modified scenario, where the roasting costs were increased by 5\%. Here's the comparison: \\
Original Cost: 2470 \\
Modified Cost (with 5\% increase in roasting costs): 2526.5\\
The increase in roasting costs has led to an increase in the overall cost by 56.5, reflecting the impact of the potential salary increase on the roasting cost. \\
If you need any further insights or assistance, please let me know!  \end{tabular}   \\
\bottomrule
\end{tabular}
\end{table*}

\begin{table*}[h!]
\caption{Application A1. Scenario 1: autonomous math solving. ChatGPT+ Code Interpreter on the first qualitative example.  The code generated does not match the devised plan. The code returns a decimal number.}
\label{tab:append_math_chatcode}
\small
\begin{tabular}{c|c|p{12cm}}
\toprule
 & \textbf{Action} & \textbf{ChatGPT+ Code Interpreter}     \\
 \midrule

\faUser & Input & Completely simplify and rationalize the denominator: $$\frac{\sqrt{160}}{\sqrt{252}}\times\frac{\sqrt{245}}{\sqrt{108}}$$ \\
\midrule
\faAndroid  & & 

To simplify and rationalize the expression, we first simplify the radicals in the numerator and denominator by factoring out the largest perfect squares from the numbers under the square root. 

The perfect square factors of 160, 252, 245, and 108 are 16, 4, 1, and 36, respectively. Let's factorize and simplify: 

\[
\frac{\sqrt{160}}{\sqrt{252}}\times\frac{\sqrt{245}}{\sqrt{108}} = \frac{\sqrt{16 \times 10}}{\sqrt{4 \times 63}}\times\frac{\sqrt{245}}{\sqrt{36 \times 3}}
\]

We can simplify this further by applying the property \(\sqrt{ab} = \sqrt{a} \sqrt{b}\) and \(\frac{a}{b} \times \frac{c}{d} = \frac{a \times c}{b \times d}\).
\\

\faAndroid  & & 

Python
\begin{lstlisting}
import math

# factoring the square roots
sqrt_160 = math.sqrt(16) * math.sqrt(10)
sqrt_252 = math.sqrt(4) * math.sqrt(63)
sqrt_245 = math.sqrt(245)
sqrt_108 = math.sqrt(36) * math.sqrt(3)

# simplify the fraction
numerator = sqrt_160 * sqrt_245
denominator = sqrt_252 * sqrt_108
simplified_fraction = numerator / denominator

simplified_fraction

\end{lstlisting}

Result:
\begin{lstlisting}
1.200137166371826
\end{lstlisting} \\

\faAndroid  & & 

The completely simplified and rationalized form of the given expression is approximately \(1.200\). \\
\bottomrule
\end{tabular}
\end{table*}

\begin{table*}[h!]
\centering
\caption{Application A3. \libName with OptiGuide. The same question ``What if the roasting cost is increased by 5\% because of the potential salary increase?" is asked.}
\label{tab:optiguide_complex}
\small
\begin{tabular}{c|c|p{11cm}}
\toprule
 & \textbf{Action} & \textbf{\libName}     \\
\midrule
\faUser & Prompt & \textit{What if the roasting cost is increased by 5\% because of the potential salary increase?} \\
\midrule
\faAndroid  & & After increasing the roasting costs by 5\%, I've re-calculated the optimal coffee distribution solution for the supply chain. As a result, the new optimal total cost has become 2526.5, compared to the initial cost of 2470.0. This indicates that the increased roasting costs have caused the overall distribution cost to rise by 56.5 units.
   \\
\bottomrule
\end{tabular}
\end{table*}

\begin{table*}[h!]
\centering
\caption{Application A1. Scenario 1: autonomous math solving. Preliminary test with BabyAGI.}
\label{tab:bayagi_ex}
\tiny
\begin{tabular}{c|c|p{11cm}}
\toprule
 & \textbf{Action} & \textbf{BabyAGI}     \\
\midrule
\faUser & Set .env & OBJECTIVE=Solve math problems

INITIAL\_TASK=Completely simplify and rationalize the denominator: 
$$\frac{\sqrt{160}}{\sqrt{252}}\times\frac{\sqrt{245}}{\sqrt{108}}$$
\\
\midrule
\faAndroid  & & Solve math problems

Initial task: Completely simplify and rationalize the denominator: $\frac{\sqrt{160}}{\sqrt{252}}\times\frac{\sqrt{245}}{\sqrt{108}}$

*****TASK LIST*****

 • Completely simplify and rationalize the denominator: $\frac{\sqrt{160}}{\sqrt{252}}\times\frac{\sqrt{245}}{\sqrt{108}}$

*****NEXT TASK*****

Completely simplify and rationalize the denominator: $\frac{\sqrt{160}}{\sqrt{252}}\times\frac{\sqrt{245}}{\sqrt{108}}$

*****TASK RESULT*****

First, let's simplify each square root separately:

- $$\sqrt{160} = \sqrt{16*10} = \sqrt{16}*\sqrt{10} = 4*\sqrt{10}$$ 
- $$\sqrt{252} = \sqrt{4*63}= \sqrt{4}\times\sqrt{63}= 2\sqrt{63}$$
- $$\sqrt{245}= \sqrt{49*5}= \sqrt{49}\times\sqrt{5}= 7\sqrt{5}$$
- $$\sqrt{108}= \sqrt{36*3}= \sqrt{36}\times\sqrt{3}= 6\sqrt{3}$$

Now, substitute these values into the original expression which gives:

$$ \frac{4*\sqrt{10}}{2\sqrt{63}}\times\frac{7\sqrt{5}}{6\sqrt{3}} = \frac{2*\sqrt{10}}{\sqrt{63}} \times \frac{7\sqrt{5}}{2\sqrt{3}}$$

We get rid of the square root in the denominator by multiplying the fractions which gives:

$$= \frac{2*7\sqrt{10*5}}{2*\sqrt{63*3}} =  \frac{14\sqrt{50}}{\sqrt{189}}
= \frac{14*5\sqrt{2}}{7\sqrt{27}} = 2\sqrt{2}*\frac{10}{\sqrt{27}} = \frac{20\sqrt{2}}{\sqrt{27}}.$$

To simplify further and rationalize the denominator, we multiply by $\frac{\sqrt{27}}{\sqrt{27}}$ which gives:

$= \frac{20\sqrt{2}}{\sqrt{27}} * \frac{\sqrt{27}}{\sqrt{27}}
=  \frac{20\sqrt{54}}{27} = \frac{20*3\sqrt{6}}{27}= \frac{60\sqrt{6}}{27}.$ 

Simplifying further gives:

$= \frac{20\sqrt{6}}{9}.$

*****TASK CREATION AGENT PROMPT****

You are to use the result from an execution agent to create new tasks with the following objective: Solve math problems.
The last completed task has the result: 
First, let's simplify each square root separately:

- $$\sqrt{160} = \sqrt{16*10} = \sqrt{16}*\sqrt{10} = 4*\sqrt{10}$$ 
$$\textbf{...Repeat after ``First, let’s simplify each square..''}$$

Now, substitute these values into the original expression which gives:

$$ \frac{4*\sqrt{10}}{2\sqrt{63}}\times\frac{7\sqrt{5}}{6\sqrt{3}} = \frac{2*\sqrt{10}}{\sqrt{63}} \times \frac{7\sqrt{5}}{2\sqrt{3}}$$

We get rid of the square root in the denominator by multiplying the fractions which gives:

$$= \frac{2*7\sqrt{10*5}}{2*\sqrt{63*3}} =  \frac{14\sqrt{50}}{\sqrt{189}}
= \frac{14*5\sqrt{2}}{7\sqrt{27}} = 2\sqrt{2}*\frac{10}{\sqrt{27}} = \frac{20\sqrt{2}}{\sqrt{27}}.$$

To simplify further and rationalize the denominator, we multiply by $\frac{\sqrt{27}}{\sqrt{27}}$ which gives:

$$= \frac{20\sqrt{2}}{\sqrt{27}} * \frac{\sqrt{27}}{\sqrt{27}}
=  \frac{20\sqrt{54}}{27} = \frac{20*3\sqrt{6}}{27}= \frac{60\sqrt{6}}{27}.$$ 

Simplifying further gives:

$= \frac{20\sqrt{6}}{9}.$
This result was based on this task description: Completely simplify and rationalize the denominator: $\frac{\sqrt{160}}{\sqrt{252}}\times\frac{\sqrt{245}}{\sqrt{108}}$.
Based on the result, return a list of tasks to be completed in order to meet the objective. 
Return one task per line in your response. The result must be a numbered list in the format:

\# First task
\# Second task

The number of each entry must be followed by a period. If your list is empty, write ``There are no tasks to add at this time.''
Unless your list is empty, do not include any headers before your numbered list or follow your numbered list with any other output.

****TASK CREATION AGENT RESPONSE****

1. Simplify and rationalize the denominator for the expression $\frac{\sqrt{225}}{\sqrt{45}}\times\frac{\sqrt{200}}{\sqrt{125}}$
2. Simplify and rationalize the denominator for the expression $\frac{\sqrt{289}}{\sqrt{361}}\times\frac{\sqrt{100}}{\sqrt{72}}$
\textbf{...Until 10}

Adding new tasks to task\_storage
{`task\_name': `Simplify and rationalize the denominator for the expression fracsqrt225sqrt45timesfracsqrt200sqrt125', `task\_id': 2}

{`task\_name': `Simplify and rationalize the denominator for the expression fracsqrt289sqrt361timesfracsqrt100sqrt72', `task\_id': 3}

\textbf{...Until task id 11}

****TASK PRIORITIZATION AGENT PROMPT****

You are tasked with prioritizing the following tasks: 
Simplify and rationalize the denominator for the expression fracsqrt225sqrt45timesfracsqrt200sqrt125
Simplify and rationalize the denominator for the expression fracsqrt289sqrt361timesfracsqrt100sqrt72

\textbf{...Aborted}
   \\
\bottomrule
\end{tabular}
\end{table*}

\begin{table*}[h!]
\centering
\caption{Application A1, Scenario 1: Autonomous math problem-solving. Preliminary test with CAMEL. We constructed two roles: ``Math Solver'' and ``Python Programmer''. The resulting code cannot be executed. The conversation continues, and the roles are not keen on solving the problem and giving an answer.}
\label{tab:camel_ex}
\tiny
\begin{tabular}{c|c|p{11cm}}
\toprule
 & \textbf{Action} & CAMEL   \\
\midrule
\faUser & Set task prompt & Completely simplify and rationalize the denominator: $$\frac{\sqrt{160}}{\sqrt{252}}\times\frac{\sqrt{245}}{\sqrt{108}}$$ \\
\midrule
\faAndroid & System & **************Assistant sys message*************
AI SystemMessage(role\_name=`Python Programmer', role\_type=$<$RoleType.ASSISTANT: `assistant'$>$, meta\_dict={`task': `Using Python programming language, write a program that accepts four integers as input, applies square root function to each, simplifies and rationalizes the denominator of the fraction resulting from the multiplication of two fractions: $\frac{\sqrt{160}}{\sqrt{252}}\times\frac{\sqrt{245}}{\sqrt{108}}$, and returns the simplified expression.', `assistant\_role': `Python Programmer', `user\_role': `Math Solver'}, role=`system', content=`Never forget you are a Python Programmer and I am a Math Solver. Never flip roles! ...)

**************AI User sys message*************

SystemMessage(role\_name=`Math Solver', role\_type=$<$RoleType.USER: `user'$>$, meta\_dict={`task': `Using Python programming language, write a program that accepts four integers as input, applies square root function to each, simplifies and rationalizes the denominator of the fraction resulting from the multiplication of two fractions: $\frac{\sqrt{160}}{\sqrt{252}}\times\frac{\sqrt{245}}{\sqrt{108}}$, and returns the simplified expression.', `assistant\_role': `Python Programmer', `user\_role': `Math Solver'}, role=`system', content=`Never forget you are a Math Solver and I am a Python Programmer. Never flip roles! You will always instruct me.We share a common interest in collaborating to successfully complete a task.I must help you to complete the task.Here is the task: Using Python programming language, write a program that accepts four integers as input, applies square root function to each, simplifies and rationalizes the denominator of the fraction resulting from the multiplication of two fractions: $\frac{\sqrt{160}}{\sqrt{252}}\times\frac{\sqrt{245}}{\sqrt{108}}$, and returns the simplified expression.. ...)

**************Original task prompt*************

Completely simplify and rationalize the denominator: $\frac{{\sqrt{{160}}}}{{\sqrt{{252}}}}\times\frac{{\sqrt{{245}}}}{{\sqrt{{108}}}}$

**************Specified task prompt*************

Using Python programming language, write a program that accepts four integers as input, applies square root function to each, simplifies and rationalizes the denominator of the fraction resulting from the multiplication of two fractions: $\frac{\sqrt{160}}{\sqrt{252}}\times\frac{\sqrt{245}}{\sqrt{108}}$, and returns the simplified expression.

**************Final task prompt*************

Using Python programming language, write a program that accepts ...
\\
\midrule

\faUser & AI User & Instruction: Import the necessary libraries for the task.
Input: None \\
\midrule

\faAndroid & AI Assistant & Solution: To solve this task, we will need the `math` library for the square root function and the `fractions` library to simplify the resulting fraction. Here is how to import these libraries:

\begin{lstlisting}
import math
from fractions import Fraction
\end{lstlisting}

The `math` library provides functions that deal with mathematical operations. The `sqrt` function, which we will use in this task, is part of this library.

The `fractions` library provides functions for creating and manipulating rational numbers. The `Fraction` function, which we will use to simplify the resulting fraction, is part of this library.

Next request. \\
\midrule

\faUser & AI User & Instruction: Define a function that accepts four integers as input.
Input: None \\
\midrule

\faAndroid & AI Assistant & Solution: Let's define a function named `simplify\_fraction` that accepts four integers as input. These integers will represent the numbers under the square root in the fractions.

\begin{lstlisting}
def simplify_fraction(a, b, c, d):
    pass
\end{lstlisting}

In this function, `a', `b', `c', and `d' are the parameters that will accept the four integers as input. The `pass' statement is a placeholder and will be replaced with actual code in the next steps.

Next request. \\
\midrule

\faUser & AI User & Instruction: Inside the function, apply the square root function to each of the input integers.
Input: None \\
\midrule

\faAndroid & AI Assistant & Solution: We will use the `math.sqrt' function from the `math' library to apply the square root to each of the input integers. Here is how to do it:

\begin{lstlisting}
def simplify_fraction(a, b, c, d):
    sqrt_a = math.sqrt(a)
    sqrt_b = math.sqrt(b)
    sqrt_c = math.sqrt(c)
    sqrt_d = math.sqrt(d)
\end{lstlisting}

In this code, `sqrt\_a', `sqrt\_b', `sqrt\_c', and `sqrt\_d' are variables that store the square root of the input integers `a', `b', `c', and `d' respectively.

Next request. \\
\midrule
\faUser & Human & Abort \\
\bottomrule

\end{tabular}
\end{table*}

\begin{table*}[h!]
\centering
\caption{Application A1, Scenario 1: Autonomous math problem-solving.  Preliminary evaluation on MetaGPT. It does not solve the math problem but gives a product solution.}
\label{tab:metagpt_ex}
\small
\begin{tabular}{c|c|p{11cm}}
\toprule
 & \textbf{Action} & MetaGPT     \\
\midrule
\faUser & Prompt &  Completely simplify and rationalize the denominator: $$\frac{\sqrt{160}}{\sqrt{252}}\times\frac{\sqrt{245}}{\sqrt{108}}$$  \\
\midrule
\faAndroid  & & The boss has tasked me with creating a successful product that is efficient.

\#\# Product Goals
\begin{lstlisting}
[
    ``Create a product that meets the needs of the target market",
    ``Ensure the product is efficient and easy to use",
    ``Achieve a high level of customer satisfaction"
]
\end{lstlisting}

\#\# User Stories
\begin{lstlisting}
[
    ``As a user, I want a product that is easy to navigate so that I can find what I need quickly",
    ``As a user, I want a product that is reliable and efficient so that I can depend on it",
    ``As a user, I want a product that meets my needs so that I can get value for my money",
    ``As a user, I want a product that is constantly updated and improved so that I can enjoy the best features",
    ``As a user, I want a product that offers good customer support so that I can get help when I need it"
]
\end{lstlisting}
\#\# Competitive Analysis
\begin{lstlisting}
[
    ``Product A: Offers similar features but lacks efficiency",
    ``Product B: Highly efficient but does not meet all the needs of the target market",
    ``Product C: Has a high level of customer satisfaction but is not easy to navigate",
    ``Product D: Is constantly updated and improved but lacks reliability",
    ``Product E: Offers good customer support but is not efficient",
    ``Product F: Is reliable and efficient but does not offer good customer support",
    ``Our Target Product: Aims to combine all the best features of the competitors while improving on their weaknesses"
]
\end{lstlisting}

\#\# Competitive Quadrant Chart

   \\
   \midrule
   \faUser & Human & Abort\\
\bottomrule
\end{tabular}
\end{table*}

\begin{table*}[h!]
 \caption{Application A7. Example output of MiniWobChat on one web manipulation task ``click-button-sequence.''}
\label{tab:append_miniwobchat}
\tiny
\begin{tabular}{c|c|p{11cm}}
\toprule
 & \textbf{Action} & \textbf{\libName}     \\

\midrule
\faUser & Init chat & 

Executor agent to AssistantAgent:

We have an autonomous computer control agent that can perform atomic instructions specified by natural language to control computers. There are two types of instructions it can execute. 

First, given the instruction that matches the regular expression \verb|``^type\s.{1,}$"|, it can type a list of characters via the keyboard. This instruction should specify the target keyboard input for the agent to type. Before this typing instruction, you should first locate the cursor by clicking the input box with click instruction.

Second, given the instruction that matches the regular expression, \verb|``^clickxpath\s.{1,}$"|, it can click an HTML element with an xpath that is visible on the webpage. The target of this instruction should be a valid xpath.
Below is the HTML code of the webpage where the agent should solve a task.
\begin{lstlisting}
<div id="wrap" data-wob_ref="2" data-wob_eps="e0">
  <div id="query">Click button ONE, then click button TWO.</div>
  <div id="area" data-wob_ref="3" data-wob_eps="e0">
    <button id="subbtn" style="position:absolute; left:103px; top:87px" data-wob_ref="4" data-wob_eps="e0">ONE</button>
    <button id="subbtn2" style="position:absolute; left:44px; top:97px" data-wob_ref="5" data-wob_eps="e0">TWO</button>
  </div>
</div>
\end{lstlisting}

task: Click button ONE, then click button TWO.

plan:

1. Click the button with xpath ``//button[@id=`subbtn']".

2. Click the button with xpath ``//button[@id=`subbtn2']".

Current task: Click button ONE, then click button TWO.

plan:
\\
\midrule
\faAndroid  & & 
***************************************************************

AssistantAgent to Executor agent:

1. Click the button with xpath ``//button[@id=`subbtn']".

2. Click the button with xpath ``//button[@id=`subbtn2']".

***************************************************************

Executor agent to AssistantAgent:

Below is the HTML code of the webpage where the agent should solve a task.

\begin{lstlisting}
<div id="wrap" data-wob_ref="2" data-wob_eps="e0">
  <div id="query">Click button ONE, then click button TWO.</div>
  <div id="area" data-wob_ref="3" data-wob_eps="e0">
    <button id="subbtn" style="position:absolute; left:103px; top:87px" data-wob_ref="4" data-wob_eps="e0">ONE</button>
    <button id="subbtn2" style="position:absolute; left:44px; top:97px" data-wob_ref="5" data-wob_eps="e0">TWO</button>
  </div>
</div>
\end{lstlisting}

Current task: Click button ONE, then click button TWO.

Here is a plan you are following now.

1. Click the button with xpath ``//button[@id=`subbtn']".

2. Click the button with xpath ``//button[@id=`subbtn2']".

We have a history of instructions that have been already executed by the autonomous agent so far.

No instruction has been executed yet.

Based on the plan and the history of instructions executed so far, the first instruction should be `

***************************************************************

AssistantAgent to Executor agent:

clickxpath //button[@id=`subbtn']

***************************************************************

Executor agent to AssistantAgent:

Hold on, please wait for my next instruction.

***************************************************************

AssistantAgent to Executor agent:

Sure, I'll be here when you're ready with your next instruction.

***************************************************************

Executor agent to AssistantAgent:

Below is the HTML code of the webpage where the agent should solve a task.
\begin{lstlisting}
<div id="wrap" data-wob_ref="2" data-wob_eps="e0">
  <div id="query">Click button ONE, then click button TWO.</div>
  <div id="area" data-wob_ref="3" data-wob_eps="e0">
    <button id="subbtn" style="position:absolute; left:103px; top:87px" data-wob_ref="4" data-wob_eps="e0" data-tampered="e0">ONE</button>
    <button id="subbtn2" style="position:absolute; left:44px; top:97px" data-wob_ref="5" data-wob_eps="e0">TWO</button>
  </div>
</div> 
\end{lstlisting}
Current task: Click button ONE, then click button TWO.

Here is a plan you are following now.

1. Click the button with xpath ``//button[@id=`subbtn']".

2. Click the button with xpath ``//button[@id=`subbtn2']".

We have a history of instructions that have been already executed by the autonomous agent so far.

1: clickxpath //button[@id=`subbtn']

Based on the plan and the history of instructions executed so far, the next proper instruction should be `

***************************************************************

AssistantAgent to Executor agent:

clickxpath //button[@id=`subbtn2']

***************************************************************

Executor agent to AssistantAgent:

SUCCESS!!!!
   \\

\midrule

\faAndroid  & & TERMINATE \\

\bottomrule
\end{tabular}
\end{table*}

%% file: arxiv_main.bbl
\begin{thebibliography}{47}
\providecommand{\natexlab}[1]{#1}
\providecommand{\url}[1]{\texttt{#1}}
\expandafter\ifx\csname urlstyle\endcsname\relax
  \providecommand{\doi}[1]{doi: #1}\else
  \providecommand{\doi}{doi: \begingroup \urlstyle{rm}\Url}\fi

\bibitem[Adlakha et~al.(2023)Adlakha, BehnamGhader, Lu, Meade, and
  Reddy]{adlakha2023evaluating}
Vaibhav Adlakha, Parishad BehnamGhader, Xing~Han Lu, Nicholas Meade, and Siva
  Reddy.
\newblock Evaluating correctness and faithfulness of instruction-following
  models for question answering.
\newblock \emph{arXiv preprint arXiv:2307.16877}, 2023.

\bibitem[Amershi et~al.(2019)Amershi, Weld, Vorvoreanu, Fourney, Nushi,
  Collisson, Suh, Iqbal, Bennett, Inkpen, et~al.]{amershi-chi19}
Saleema Amershi, Dan Weld, Mihaela Vorvoreanu, Adam Fourney, Besmira Nushi,
  Penny Collisson, Jina Suh, Shamsi Iqbal, Paul~N Bennett, Kori Inkpen, et~al.
\newblock Guidelines for human-ai interaction.
\newblock In \emph{Proceedings of the 2019 chi conference on human factors in
  computing systems}, 2019.

\bibitem[Amodei et~al.(2016)Amodei, Olah, Steinhardt, Christiano, Schulman, and
  Mané]{amodei-arxiv2016}
Dario Amodei, Chris Olah, Jacob Steinhardt, Paul Christiano, John Schulman, and
  Dan Mané.
\newblock Concrete problems in ai safety, 2016.

\bibitem[AutoGPT(2023)]{autogpt}
AutoGPT.
\newblock Documentation | auto-gpt.
\newblock \url{https://docs.agpt.co/}, 2023.

\bibitem[BabyAGI(2023)]{babyagi}
BabyAGI.
\newblock Github | babyagi.
\newblock \url{https://github.com/yoheinakajima/babyagi}, 2023.

\bibitem[Cai et~al.(2019)Cai, Winter, Steiner, Wilcox, and
  Terry]{Cai2019HelloAU}
Carrie~J. Cai, Samantha Winter, David~F. Steiner, Lauren Wilcox, and Michael
  Terry.
\newblock "hello ai": Uncovering the onboarding needs of medical practitioners
  for human-ai collaborative decision-making.
\newblock \emph{Proceedings of the ACM on Human-Computer Interaction}, 2019.

\bibitem[Cai et~al.(2023)Cai, Wang, Ma, Chen, and Zhou]{cai2023large}
Tianle Cai, Xuezhi Wang, Tengyu Ma, Xinyun Chen, and Denny Zhou.
\newblock Large language models as tool makers.
\newblock \emph{arXiv preprint arXiv:2305.17126}, 2023.

\bibitem[Chroma(2023)]{chromadb}
Chroma.
\newblock {Chromadb}.
\newblock \url{https://github.com/chroma-core/chroma}, 2023.

\bibitem[Dibia(2023)]{dibia-2023-lida}
Victor Dibia.
\newblock {LIDA}: A tool for automatic generation of grammar-agnostic
  visualizations and infographics using large language models.
\newblock In \emph{Proceedings of the 61st Annual Meeting of the Association
  for Computational Linguistics (Volume 3: System Demonstrations)}, Toronto,
  Canada, July 2023. Association for Computational Linguistics.

\bibitem[Dong et~al.(2023)Dong, Jiang, Jin, and Li]{dong2023self}
Yihong Dong, Xue Jiang, Zhi Jin, and Ge~Li.
\newblock Self-collaboration code generation via chatgpt.
\newblock \emph{arXiv preprint arXiv:2304.07590}, 2023.

\bibitem[Du et~al.(2023)Du, Li, Torralba, Tenenbaum, and
  Mordatch]{du2023improving}
Yilun Du, Shuang Li, Antonio Torralba, Joshua~B Tenenbaum, and Igor Mordatch.
\newblock Improving factuality and reasoning in language models through
  multiagent debate.
\newblock \emph{arXiv preprint arXiv:2305.14325}, 2023.

\bibitem[Eleti et~al.(2023)Eleti, Harris, and Kilpatrick]{funccall}
Atty Eleti, Jeff Harris, and Logan Kilpatrick.
\newblock Function calling and other api updates.
\newblock \url{https://openai.com/blog/function-calling-and-other-api-updates},
  2023.

\bibitem[Guidance(2023)]{guidance}
Guidance.
\newblock Guidance.
\newblock \url{https://github.com/guidance-ai/guidance}, 2023.

\bibitem[Hendrycks et~al.(2021)Hendrycks, Burns, Kadavath, Arora, Basart, Tang,
  Song, and Steinhardt]{hendrycks2021measuring}
Dan Hendrycks, Collin Burns, Saurav Kadavath, Akul Arora, Steven Basart, Eric
  Tang, Dawn Song, and Jacob Steinhardt.
\newblock Measuring mathematical problem solving with the math dataset.
\newblock \emph{arXiv preprint arXiv:2103.03874}, 2021.

\bibitem[Hong et~al.(2023)Hong, Zheng, Chen, Cheng, Zhang, Wang, Yau, Lin,
  Zhou, Ran, et~al.]{hong2023metagpt}
Sirui Hong, Xiawu Zheng, Jonathan Chen, Yuheng Cheng, Ceyao Zhang, Zili Wang,
  Steven Ka~Shing Yau, Zijuan Lin, Liyang Zhou, Chenyu Ran, et~al.
\newblock Metagpt: Meta programming for multi-agent collaborative framework.
\newblock \emph{arXiv preprint arXiv:2308.00352}, 2023.

\bibitem[Horvitz(1999)]{horvitz-chi99}
Eric Horvitz.
\newblock Principles of mixed-initiative user interfaces.
\newblock In \emph{Proceedings of the SIGCHI conference on Human Factors in
  Computing Systems}, 1999.

\bibitem[HuggingFace(2023)]{transformersagent}
HuggingFace.
\newblock Transformers agent.
\newblock \url{https://huggingface.co/docs/transformers/transformers\_agents},
  2023.

\bibitem[Kim et~al.(2023)Kim, Baldi, and McAleer]{kim2023language}
Geunwoo Kim, Pierre Baldi, and Stephen McAleer.
\newblock Language models can solve computer tasks.
\newblock \emph{arXiv preprint arXiv:2303.17491}, 2023.

\bibitem[Kwiatkowski et~al.(2019)Kwiatkowski, Palomaki, Redfield, Collins,
  Parikh, Alberti, Epstein, Polosukhin, Devlin, Lee,
  et~al.]{kwiatkowski2019natural}
Tom Kwiatkowski, Jennimaria Palomaki, Olivia Redfield, Michael Collins, Ankur
  Parikh, Chris Alberti, Danielle Epstein, Illia Polosukhin, Jacob Devlin,
  Kenton Lee, et~al.
\newblock Natural questions: a benchmark for question answering research.
\newblock \emph{Transactions of the Association for Computational Linguistics},
  2019.

\bibitem[LangChain(2023)]{langchain}
LangChain.
\newblock Introduction | langchain.
\newblock \url{https://python.langchain.com/en/latest/index.html}, 2023.

\bibitem[Lewis et~al.(2017)Lewis, Yarats, Dauphin, Parikh, and
  Batra]{lewis2017deal}
Mike Lewis, Denis Yarats, Yann~N Dauphin, Devi Parikh, and Dhruv Batra.
\newblock Deal or no deal? end-to-end learning for negotiation dialogues.
\newblock \emph{arXiv preprint arXiv:1706.05125}, 2017.

\bibitem[Lewis et~al.(2020)Lewis, Perez, Piktus, Petroni, Karpukhin, Goyal,
  K{\"u}ttler, Lewis, Yih, Rockt{\"a}schel, et~al.]{lewis2020retrieval}
Patrick Lewis, Ethan Perez, Aleksandra Piktus, Fabio Petroni, Vladimir
  Karpukhin, Naman Goyal, Heinrich K{\"u}ttler, Mike Lewis, Wen-tau Yih, Tim
  Rockt{\"a}schel, et~al.
\newblock Retrieval-augmented generation for knowledge-intensive nlp tasks.
\newblock \emph{Advances in Neural Information Processing Systems}, 2020.

\bibitem[Li et~al.(2023{\natexlab{a}})Li, Mellou, Zhang, Pathuri, and
  Menache]{li2023large}
Beibin Li, Konstantina Mellou, Bo~Zhang, Jeevan Pathuri, and Ishai Menache.
\newblock Large language models for supply chain optimization.
\newblock \emph{arXiv preprint arXiv:2307.03875}, 2023{\natexlab{a}}.

\bibitem[Li et~al.(2023{\natexlab{b}})Li, Hammoud, Itani, Khizbullin, and
  Ghanem]{li2023camel}
Guohao Li, Hasan Abed Al~Kader Hammoud, Hani Itani, Dmitrii Khizbullin, and
  Bernard Ghanem.
\newblock Camel: Communicative agents for "mind" exploration of large scale
  language model society, 2023{\natexlab{b}}.

\bibitem[Liang et~al.(2023)Liang, He, Jiao, Wang, Wang, Wang, Yang, Tu, and
  Shi]{liang-arxiv2023}
Tian Liang, Zhiwei He, Wenxiang Jiao, Xing Wang, Yan Wang, Rui Wang, Yujiu
  Yang, Zhaopeng Tu, and Shuming Shi.
\newblock Encouraging divergent thinking in large language models through
  multi-agent debate, 2023.

\bibitem[Liu et~al.(2018)Liu, Guu, Pasupat, Shi, and
  Liang]{liu2018reinforcement}
Evan~Zheran Liu, Kelvin Guu, Panupong Pasupat, Tianlin Shi, and Percy Liang.
\newblock Reinforcement learning on web interfaces using workflow-guided
  exploration.
\newblock \emph{arXiv preprint arXiv:1802.08802}, 2018.

\bibitem[Liu(2022)]{Liu_LlamaIndex_2022}
Jerry Liu.
\newblock {LlamaIndex}, November 2022.
\newblock URL \url{https://github.com/jerryjliu/llama_index}.

\bibitem[Mnih et~al.(2013)Mnih, Kavukcuoglu, Silver, Graves, Antonoglou,
  Wierstra, and Riedmiller]{mnih2013playing}
Volodymyr Mnih, Koray Kavukcuoglu, David Silver, Alex Graves, Ioannis
  Antonoglou, Daan Wierstra, and Martin Riedmiller.
\newblock Playing atari with deep reinforcement learning.
\newblock \emph{arXiv preprint arXiv:1312.5602}, 2013.

\bibitem[Navigli et~al.(2023)Navigli, Conia, and Ross]{navigli2023biases}
Roberto Navigli, Simone Conia, and Bj{\"o}rn Ross.
\newblock Biases in large language models: Origins, inventory and discussion.
\newblock \emph{ACM Journal of Data and Information Quality}, 2023.

\bibitem[OpenAI(2023)]{gptPlugin}
OpenAI.
\newblock {ChatGPT plugins}.
\newblock \url{https://openai.com/blog/chatgpt-plugins}, 2023.

\bibitem[Park et~al.(2023)Park, O'Brien, Cai, Morris, Liang, and
  Bernstein]{park2023generative}
Joon~Sung Park, Joseph~C O'Brien, Carrie~J Cai, Meredith~Ringel Morris, Percy
  Liang, and Michael~S Bernstein.
\newblock Generative agents: Interactive simulacra of human behavior.
\newblock \emph{arXiv preprint arXiv:2304.03442}, 2023.

\bibitem[Parvez et~al.(2021)Parvez, Ahmad, Chakraborty, Ray, and
  Chang]{parvez2021retrieval}
Md~Rizwan Parvez, Wasi~Uddin Ahmad, Saikat Chakraborty, Baishakhi Ray, and
  Kai-Wei Chang.
\newblock Retrieval augmented code generation and summarization.
\newblock \emph{arXiv preprint arXiv:2108.11601}, 2021.

\bibitem[Patil et~al.(2023)Patil, Zhang, Wang, and Gonzalez]{patil2023gorilla}
Shishir~G. Patil, Tianjun Zhang, Xin Wang, and Joseph~E. Gonzalez.
\newblock Gorilla: Large language model connected with massive apis.
\newblock \emph{arXiv preprint arXiv:2305.15334}, 2023.

\bibitem[Reimers \& Gurevych(2019)Reimers and
  Gurevych]{reimers-2019-sentence-bert}
Nils Reimers and Iryna Gurevych.
\newblock Sentence-bert: Sentence embeddings using siamese bert-networks.
\newblock In \emph{Proceedings of the 2019 Conference on Empirical Methods in
  Natural Language Processing}. Association for Computational Linguistics, 11
  2019.
\newblock URL \url{https://arxiv.org/abs/1908.10084}.

\bibitem[Semantic-Kernel(2023)]{sk}
Semantic-Kernel.
\newblock Semantic kernel.
\newblock \url{https://github.com/microsoft/semantic-kernel}, 2023.

\bibitem[Shen et~al.(2021)Shen, Xia, Li, Mart{\'\i}n-Mart{\'\i}n, Fan, Wang,
  P{\'e}rez-D’Arpino, Buch, Srivastava, Tchapmi, et~al.]{shen2021igibson}
Bokui Shen, Fei Xia, Chengshu Li, Roberto Mart{\'\i}n-Mart{\'\i}n, Linxi Fan,
  Guanzhi Wang, Claudia P{\'e}rez-D’Arpino, Shyamal Buch, Sanjana Srivastava,
  Lyne Tchapmi, et~al.
\newblock igibson 1.0: A simulation environment for interactive tasks in large
  realistic scenes.
\newblock In \emph{2021 IEEE/RSJ International Conference on Intelligent Robots
  and Systems (IROS)}. IEEE, 2021.

\bibitem[Shi et~al.(2017)Shi, Karpathy, Fan, Hernandez, and
  Liang]{shi2017world}
Tianlin Shi, Andrej Karpathy, Linxi Fan, Jonathan Hernandez, and Percy Liang.
\newblock World of bits: An open-domain platform for web-based agents.
\newblock In \emph{International Conference on Machine Learning}. PMLR, 2017.

\bibitem[Shridhar et~al.(2021)Shridhar, Yuan, C\^ot\'e, Bisk, Trischler, and
  Hausknecht]{ALFWorld20}
Mohit Shridhar, Xingdi Yuan, Marc-Alexandre C\^ot\'e, Yonatan Bisk, Adam
  Trischler, and Matthew Hausknecht.
\newblock {ALFWorld: Aligning Text and Embodied Environments for Interactive
  Learning}.
\newblock In \emph{Proceedings of the International Conference on Learning
  Representations (ICLR)}, 2021.
\newblock URL \url{https://arxiv.org/abs/2010.03768}.

\bibitem[Vinyals et~al.(2017)Vinyals, Ewalds, Bartunov, Georgiev, Vezhnevets,
  Yeo, Makhzani, K{\"u}ttler, Agapiou, Schrittwieser,
  et~al.]{vinyals2017starcraft}
Oriol Vinyals, Timo Ewalds, Sergey Bartunov, Petko Georgiev, Alexander~Sasha
  Vezhnevets, Michelle Yeo, Alireza Makhzani, Heinrich K{\"u}ttler, John
  Agapiou, Julian Schrittwieser, et~al.
\newblock Starcraft ii: A new challenge for reinforcement learning.
\newblock \emph{arXiv preprint arXiv:1708.04782}, 2017.

\bibitem[Wang et~al.(2021)Wang, Wu, Weimer, and Zhu]{wang2021flaml}
Chi Wang, Qingyun Wu, Markus Weimer, and Erkang Zhu.
\newblock Flaml: A fast and lightweight automl library.
\newblock \emph{Proceedings of Machine Learning and Systems}, 2021.

\bibitem[Wang et~al.(2023{\natexlab{a}})Wang, Xie, Jiang, Mandlekar, Xiao, Zhu,
  Fan, and Anandkumar]{wang2023voyager}
Guanzhi Wang, Yuqi Xie, Yunfan Jiang, Ajay Mandlekar, Chaowei Xiao, Yuke Zhu,
  Linxi Fan, and Anima Anandkumar.
\newblock Voyager: An open-ended embodied agent with large language models.
\newblock \emph{arXiv preprint arXiv:2305.16291}, 2023{\natexlab{a}}.

\bibitem[Wang et~al.(2023{\natexlab{b}})Wang, Ma, Feng, Zhang, Yang, Zhang,
  Chen, Tang, Chen, Lin, et~al.]{wang2023survey}
Lei Wang, Chen Ma, Xueyang Feng, Zeyu Zhang, Hao Yang, Jingsen Zhang, Zhiyuan
  Chen, Jiakai Tang, Xu~Chen, Yankai Lin, et~al.
\newblock A survey on large language model based autonomous agents.
\newblock \emph{arXiv preprint arXiv:2308.11432}, 2023{\natexlab{b}}.

\bibitem[Weld \& Etzioni(1994)Weld and Etzioni]{weld-aaai94}
Daniel~S. Weld and Oren Etzioni.
\newblock The first law of robotics (a call to arms).
\newblock In \emph{AAAI Conference on Artificial Intelligence}, 1994.

\bibitem[Woolf(2023)]{langchainproblem}
Max Woolf.
\newblock Langchain problem.
\newblock \url{https://minimaxir.com/2023/07/langchain-problem/}, 2023.

\bibitem[Wu et~al.(2023)Wu, Jia, Zhang, Wu, Li, Zhu, Wang, Lee, Peng, and
  Wang]{wu2023empirical}
Yiran Wu, Feiran Jia, Shaokun Zhang, Qingyun Wu, Hangyu Li, Erkang Zhu, Yue
  Wang, Yin~Tat Lee, Richard Peng, and Chi Wang.
\newblock An empirical study on challenging math problem solving with gpt-4.
\newblock \emph{arXiv preprint arXiv:2306.01337}, 2023.

\bibitem[Xi et~al.(2023)Xi, Chen, Guo, He, Ding, Hong, Zhang, Wang, Jin, Zhou,
  et~al.]{xi2023rise}
Zhiheng Xi, Wenxiang Chen, Xin Guo, Wei He, Yiwen Ding, Boyang Hong, Ming
  Zhang, Junzhe Wang, Senjie Jin, Enyu Zhou, et~al.
\newblock The rise and potential of large language model based agents: A
  survey.
\newblock \emph{arXiv preprint arXiv:2309.07864}, 2023.

\bibitem[Yao et~al.(2022)Yao, Zhao, Yu, Du, Shafran, Narasimhan, and
  Cao]{yao2022react}
Shunyu Yao, Jeffrey Zhao, Dian Yu, Nan Du, Izhak Shafran, Karthik Narasimhan,
  and Yuan Cao.
\newblock React: Synergizing reasoning and acting in language models.
\newblock \emph{arXiv preprint arXiv:2210.03629}, 2022.

\end{thebibliography}
